\documentclass[fleqn,10pt]{wlscirep}

\usepackage[utf8]{inputenc}
\usepackage[T1]{fontenc}
\usepackage{lineno}
\nolinenumbers

\usepackage{algorithm}
\usepackage{algpseudocode}
\usepackage{xcolor}
\usepackage{colortbl}
\usepackage{subcaption}
\usepackage{arydshln}
\usepackage{mathrsfs}
\usepackage{amsmath}
\usepackage{CJKutf8}
\usepackage{tikz}
\newcommand\circled[1]{\tikz[baseline=(char.base)]{
            \node[shape=circle,draw,inner sep=2pt] (char) {#1};}}

\newcommand{\eg}{\textit{e.g., }}
\newcommand{\ie}{\textit{i.e., }}
\newcommand{\etal}{\textit{et al.}}

\newcommand{\cmark}{\textcolor{cyan}{\checkmark}}
\newcommand{\xmark}{\textcolor{magenta}{$\times$}}
\newcommand{\norm}[1]{\left\lVert#1\right\rVert}

\title{The Health Gym:\\
Synthetic Health-Related Datasets for the Development of Reinforcement Learning Algorithms}

\author[1,*]{Nicholas I-Hsien Kuo}
\author[2]{Mark N. Polizzotto}
\author[3, 4, 5]{Simon Finfer}
\author[6]{\\Federico Garcia}
\author[7]{Anders S\"onnerborg}
\author[8]{Maurizio Zazzi}
\author[9]{Michael B\"ohm}
\author[1]{\\Louisa Jorm}
\author[1]{Sebastiano Barbieri}
\affil[1]{Centre for Big Data Research in Health, University of New South Wales, Sydney, Australia}
\affil[2]{Australian National University, Canberra, Australia}
\affil[3]{The George Institute for Global Health, Sydney, Australia}
\affil[4]{University of New South Wales, Sydney, Australia}
\affil[5]{Imperial College London, London, United Kingdom}
\affil[6]{Hospital Universitario San Cecilio, Granada, Spain}
\affil[7]{Karolinska Institutet, Stockholm, Sweden}
\affil[8]{Universit{\`a} degli Studi di Siena, Siena, Italy}
\affil[9]{Universit{\"a}tsklinikum K{\"o}ln, K{\"o}ln, Germany}

\affil[*]{Corresponding author: Nicholas I-Hsien Kuo (n.kuo@unsw.edu.au)}

\begin{abstract}
In recent years, the machine learning research community has benefited tremendously from the availability of openly accessible benchmark datasets. Clinical data are usually not openly available due to their highly confidential nature. This has hampered the development of reproducible and generalisable machine learning applications in health care. Here we introduce the Health Gym -– a growing collection of highly realistic synthetic medical datasets that can be freely accessed to prototype, evaluate, and compare machine learning algorithms, with a specific focus on reinforcement learning. The three synthetic datasets described in this paper present patient cohorts with acute hypotension and sepsis in the intensive care unit, and people with human immunodeficiency virus (HIV) receiving antiretroviral therapy in ambulatory care. The datasets were created using a novel generative adversarial network (GAN). The distributions of variables, and correlations between variables and trends over time in the synthetic datasets mirror those in the real datasets. Furthermore, the risk of sensitive information disclosure associated with the public distribution of the synthetic datasets is estimated to be very low.
\end{abstract}
\begin{document}

\flushbottom
\maketitle

\thispagestyle{empty}

%%%===%%%
\section*{Background \& Summary}
\textit{Reinforcement learning}\cite{sutton2018reinforcement} (RL) is an area of artificial intelligence (AI) which learns a behavioural \textit{policy} -- a mapping from states to actions -- which maximises a cumulative reward in an evolving environment. Recent studies that combine RL with neural networks have achieved super-human performances in tasks from video games\cite{mnih2013playing} to complex board games\cite{silver2016mastering}. The success of RL was greatly facilitated by the availability of \textit{standard benchmark problems}: tasks with publicly available datasets which allowed the community to develop, test, and compare RL algorithms (\eg OpenAI Gym\cite{brockman2016openai}, DeepMind Lab\cite{beattie2016deepmind}, and D4RL\cite{fu2020d4rl}). Health-related data is, however, not as easily accessible due to privacy concerns around the disclosure of private information. To address this challenge, this paper introduces the \textbf{Health Gym} project -- a collection of highly realistic synthetic medical datasets that can be freely accessed to facilitate the development of machine learning (ML) algorithms, with a specific focus on RL.

\subsection*{Reinforcement Learning for Health Care: Promises and Challenges}
Clinicians treating individuals with chronic disorders (\eg human immunodeficiency virus (HIV) infection) or with potentially life-threatening conditions (\eg sepsis) often prescribe a series of treatments to maximise the chances of favourable outcomes. This generally requires modifying the duration, dosage, or type of treatment over time; and is challenging due to patient heterogeneity in responses to treatments, potential relapses, and side-effects. Clinicians often rely, at least in part, on clinical judgment to assign sequences of treatment, because the clinical evidence base is incomplete and available evidence may not represent the diversity of real-life clinical states. There is thus vast potential for RL algorithms to optimise personalised treatment regimens, as shown by early research on antiretroviral therapy in HIV\cite{yu2019incorporating}, radiotherapy planning in lung cancer\cite{tseng2017deep}, and management of sepsis\cite{komorowski2018artificial}. Nonetheless, some authors have highlighted the lack of reproducibility and potential for patient harm inherent in these methods\cite{challen2019artificial}. In particular, recommendations made by RL algorithms may not be safe if the training data omit variables that influence clinical decision making, or if the effective sample size is small\cite{gottesman2019guidelines}.

One of the main difficulties in developing robust RL algorithms for healthcare is the highly confidential nature of clinical data. Researchers are often required to establish formal collaborations and execute extensive data use agreements before sharing data. One approach to overcome these barriers is to generate synthetic data that closely resembles the original dataset but does not allow re-identification of individual patients and can therefore be freely distributed. Synthetic data generation has previously been applied to computed tomography images\cite{kim2015implementation} and electronic health records\cite{walonoski2018synthea}; and early studies found that both linear\cite{fienberg1998disclosure} and non-linear\cite{caiola2010random} models could generate continuous and categorical variables. More recently, deep learning techniques such as \textit{Generative Adversarial Networks}\cite{goodfellow2014generative} (GANs) have also been used to generate realistic medical time series\cite{esteban2017real}.

\subsection*{The Health Gym Project}
The Health Gym project is a growing collection of synthetic but realistic datasets for developing RL algorithms. Here we introduce the first three datasets related to the management of \textit{acute hypotension}\cite{gottesman2020interpretable}, \textit{sepsis}\cite{komorowski2018artificial}, and \textit{HIV}\cite{parbhoo2017combining}. All datasets were generated using GANs and the MIMIC-III\cite{johnson2016mimic} and EuResist\cite{zazzi2012predicting} databases. MIMIC-III comprises health-related data for patients who stayed in intensive care units (ICUs) of the Beth Israel Deaconess Medical Centre (Boston, USA) between 2001 and 2012. Within MIMIC-III, we identified two cohorts of patients: $3,910$ patients with acute hypotension and $2,164$ patients with sepsis. Similarly, the EuResist Integrated Database was used to extract longitudinal information related to $8,916$ people with HIV. For severe hypotension and sepsis, we extracted the related timeseries of vital signs, laboratory test results, medications (\eg administered intravenous fluids and vasopressors), and demographics. For people with HIV, we included demographics and time series of antiretroviral medications, cluster of differentiation 4+ T-lymphocytes (CD4) count and viral load measurements.

Both MIMIC-III and EuResist contain only de-identified data (\textit{i.e.,} personal identifiers have been removed and other treatments such as date shifting applied to minimise disclosure risk); however, there is a small remaining risk that personal information may be disclosed if an ``attacker'' or ``adversary'' (a person or process seeking to learn sensitive information about an individual) is able to link our published data back to personal identifiers. To minimise this risk, we evaluated the synthetic data using current best practices\cite{goncalves2020generation} in terms of both membership disclosure (\ie that no record in the synthetic data can be mapped directly to a record in the real data) and attribute disclosure (\ie that even if part of the data are known to an attacker, the remaining attributes cannot be recovered exactly). In Appendix \ref{Sec:BImpact}, we provide a broader impact statement to discuss the implementations and applications of our work.

%%%===%%%
\section*{Methods}
In this work, we applied GANs to longitudinal data extracted from the MIMIC-III\cite{johnson2016mimic} and EuResist\cite{zazzi2012predicting} databases to generate three synthetic datasets. The inclusion and exclusion criteria used to define the patient cohorts were adapted from previous studies: Gottesman \etal\cite{gottesman2020interpretable} for defining the patient cohort with acute hypotension, Komorowski \etal\cite{komorowski2018artificial} to define the sepsis cohort, and Parbhoo \etal\cite{parbhoo2017combining} to define the HIV cohort. Our synthetic datasets thus include variables that can be used to define the observations, actions, and rewards associated with RL
problems for the management of these clinical conditions. 

In order to describe our data generation procedure, we start by describing the real datasets and then provide details on our neural network design for generating the synthetic datasets. Our synthetic datasets include all variables in their real counterparts, in the identical formats, and are described in the \textbf{Data Records} section.

%%%===%%%
% Remember to use past tense to describe the real datasets
\subsection*{The Real Datasets}

\hypertarget{Sec:Methods-GroundTruths}{The} set of variables contained in each dataset is reported below. We refer interested readers to previous studies for the descriptive statistics (\ie quantiles and mean values) of the real datasets\footnote{It is out of the scope of this paper to address the descriptive statistics of the real datasets. Instead, we suggest interested readers to refer to Feng \etal\cite{feng2018transthoracic} for such details on the MIMIC-III dataset; and likewise, refer to Oette \etal\cite{oette2012efficacy} for those in EuResist.\label{FN:StatisticsForRealDatasets}}. 

\subsubsection*{Acute Hypotension}
The real dataset for the management of acute hypotension was originally proposed in the work of Gottesman \etal\cite{gottesman2020interpretable}. It was derived from MIMIC-III and contains the following clinical variables, measured over a 48-hour time period in 3,910 patients:
\begin{itemize}[noitemsep,topsep=0pt]
    \item mean arterial pressure (MAP), diastolic and systolic blood pressures (BPs);
    \item laboratory results of Alanine and Aspartate Aminotransferase (ALT and AST), lactate, and serum creatinine;
    \item mechanical ventilation parameters such as partial pressure of oxygen (PaO2) and fraction of inspired oxygen (FiO2);
    \item Glasgow Coma Scale (GCS) score;
    \item administered fluid boluses and vasopressors; and
    \item urine outputs.
\end{itemize}

Further details are reported in Table \ref{Tab:Hypotension}. Data were aggregated for every hour in the time series; there are hence 48 data points per variable for each patient. Data missingness in clinical time series is usually highly informative, indicating \eg the need for specific laboratory tests. Hence the real dataset includes variables with suffix \textit{(M)} to indicate whether a variable was measured at a specific point in time. 

In their work, Gottesman \etal\textcolor{white}{ } used this dataset to develop an RL agent which suggested the optimal amounts of fluid boluses and vasopressors for the management of acute hypotension. Notably, they binned both fluid boluses and vasopressors into multiple categories for the RL agent to make decisions in a discrete action space. Appendix \ref{App:HypotensionDetail} contains technical details for deriving this real dataset.

\subsubsection*{Sepsis}
The real sepsis dataset constructed by Komorowski \etal\cite{komorowski2018artificial} was also derived from MIMIC-III. It is more complex than the real hypotension dataset and comprises 44 variables, including vital signs, laboratory results, mechanical ventilation information, and various patient measurements. The complete list of variables is reported in Tables \ref{Tab:Sepsis_Numeric} and \ref{Tab:Sepsis.NonNumeric}.

The real sepsis dataset contains time series data for 2,164 patients. However, the duration of hospital stay varies for each patient. The shortest record is 8 hours long and the longest record lasts 80 hours. Furthermore, the data are reported in 4-hour windows; hence, the shortest patient record contains 2 data points, whereas the longest contains 20 data points. Appendix \ref{App:Sepsis} describes the technical details for deriving the real sepsis dataset.

In their paper, Komorowski \etal\textcolor{white}{ } employed an RL agent to prescribe different doses of intravenous fluids and vasopressors based on a patient's clinical variables. Their RL agent was trained by assigning rewards depending on whether the patients transitioned to a more favourable health state following the actions taken.

We purposely left out some variables from the work of Komorowski \etal\text{ } in our real sepsis dataset. Namely, we did not include the four items of FiO2 ratio (P/F ratio), shock index, sequential organ failure assessment score (SOFA score), and systemic inflammatory response syndrome criteria (SIRS criteria). These items were excluded because they can easily be derived from the other variables that are included. In Appendix \ref{App:Sepsis_DP}, we provide further information on deriving these auxiliary variables.

\subsubsection*{HIV}
Our real HIV dataset is based on the study of Parbhoo \etal\cite{parbhoo2017combining}. In their paper, Parbhoo \etal\textcolor{white}{ } extracted a cohort of people with HIV from the EuResist\cite{zazzi2012predicting} database; and proposed a mixture-of-experts approach for the therapy selection for people with HIV. They first used kernel-based methods to identify clusters of similar people, and then they employed an RL agent to optimise the treatment strategy. 

Although our real HIV dataset was based on their work, we made additional changes to the real HIV dataset in order to reflect a recent guideline published by the \textit{World Health Organisation} (WHO)\cite{world2016consolidated}\textcolor{white}{ } on the standardisation of antiretroviral therapy for HIV. We included 8,916 people from the EuResist database who started therapy after 2015 and were treated with the top 50 medication combinations spanning 21 medications. Appendix \ref{Sec:A.3.1} provides a discussion on the WHO guideline. 

The variables in our real HIV dataset are reported in Table \ref{Tab:HIV}. They include demographics, viral load (VL), CD4 counts, and regimen information. VL reflects how much HIV virus is in a person's body; and this variable allows medical experts to surmise the state of infection, select appropriate medications, and infer the effectiveness of past treatments. CD4 counts measure how many T-cells (a type of white blood cell) are in the body; and can be used to infer the well-being of the immune system of a person. People with very low CD4 counts are at risk of negative health outcomes. Following the aforementioned WHO guideline, we deconstructed each person's medication regimen into a collection of categorical variables representing the most commonly used base medication combinations, as well as auxiliary medications from different medication classes\footnote{The medication classes listed in Table \ref{Tab:HIV} are the \textit{integrase inhibitors} (INIs), \textit{non-nucleotide reverse transcriptase inhibitors} (NNRTIs), \textit{protease inhibitors} (PIs), and \textit{pharmacokinetic enhancers} (pk-En). The medication classes that are not explicitly mentioned included the \textit{nucleoside reverse transcriptase inhibitors} (NRTIs) and \textit{nucleotide reverse transcriptase inhibitors} (NtRTIs). NRTIs and NtRTIs are not listed separately in the table because they already formed the backbone of the most commonly found medication combinations. See further discussion in Appendix \ref{App:A.3.3}. \label{FN:DClasses}}.

Similar to the real sepsis dataset, the length of therapy in the real HIV dataset varies across people. Thus, we truncated the records and modified their lengths to the closest multiples of 10-month periods. Hence, the real HIV dataset consists of people with 10, 20, 30, etc \ldots month-long data. The shortest patient record is 10 months long whereas the longest patient record is 100 months long. Since each entry recorded 1 month of patient data, the shortest record is of length 10, and the longest record is of length 100. Similar to the hypotension dataset (Table \ref{Tab:Hypotension}), the real HIV dataset is very sparse and thus we included binary variables with suffix (M) to indicate whether a variable was measured at a specific time. Appendix \ref{App:A.3.3} reports further details on the derivation process of the real HIV dataset.

\subsection*{The Heath Gym GAN}
\hypertarget{Sec:Methods-GAN}{The} overarching pipeline of the \textit{Generative Adversarial Network\cite{goodfellow2014generative}} (GAN) is shown in Figure \ref{Fig:GANs}(a). The setup iteratively and concurrently fine-tunes two networks -- the \textit{generator} and the \textit{discriminator}\footnote{In this paper, we adopted the terminology given in the original GAN paper\cite{goodfellow2014generative}. The original paper referred to the network that classified the real and fake data as the \textit{discriminator}. However in the WGAN paper\cite{arjovsky2017wasserstein}, it was referred as the \textit{critic}; but it was again referred as the discriminator in the WGAN-GP paper\cite{gulrajani2017improved}.} -- to create highly realistic synthetic data.

%%%===%%%
% preferrably to use present tense here to contrast the tone
\subsubsection*{The GAN Setup}
The process of training a GAN model can be thought as a two-player-game with two complementary training dynamics. At first, the generator produces synthetic data samples. Then, these synthetic data are compared with samples of the real clinical data by the discriminator. The job of the discriminator is to distinguish between real and synthetic data. A mathematical description of the training procedure for the GAN is reported below. Training is concluded when the discriminator can no longer tell the real and synthetic data apart. That is, a generator is considered to be able to create highly realistic synthetic data when the discriminator is guessing randomly. When the training ends, we use the generator to create our Health Gym synthetic datasets.

For our descriptions below, we denote the generator as $G$ and the discriminator as $D$. Furthermore, we use $X_{\text{real}}$ and $X_{\text{syn}}$ to represent the real and synthetic datasets; and likewise, we designate $x_{\text{real}}$ and $x_{\text{syn}}$ as real and synthetic data batches respectively.

\subsubsection*{The Models}
As shown in Figure \ref{Fig:GANs}(a), the generator $G$ creates the synthetic data based on pseudo-random inputs $z$. The elements of $z$ are sampled from a multivariate Gaussian distribution, and they can be considered latent variables that describe intrinsic aspects of the clinical dataset. The task of network $G$ is to transform a time series of latent descriptions into a set of synthetic but realistic time series of clinical variables $G: z\rightarrow x_{\text{syn}}$.

The intermediate steps of the generator transformation are illustrated in Figure \ref{Fig:GANs}(b). Since the input is a set of latent time series, we employ a bidirectional \textit{Long Short-Term Memory}~\cite{hochreiter1997long, graves2005bidirectional} (biLSTM) \textit{recurrent neural network} (RNN) module to interpolate the relations among the latent features along the time dimension. The RNN is then followed by three fully connected dense layers\cite{rumelhart1986learning} -- high dimensional non-linear transformations responsible for feature extraction and synthetic data construction.

In order to evaluate the realisticness of the synthetic data $x_{\text{syn}}$, we forward the synthetic data along with a batch of real data $x_{\text{real}}$ to the discriminator $D$. As shown in Figure \ref{Fig:GANs}(c), the discriminator network is also a mixture of recurrent and feedforward modules. To facilitate training, the generator and the discriminator were designed to have a similar number of parameters. Since the input to $D$ (both the synthetic data and the real data) contains binary and categorical variables, we use \textit{soft embeddings}~\cite{landauer1998introduction, mottini2018airline} to represent them as numeric vectors in a machine-readable format. The discriminator $D$ employs two fully connected dense layers to interconnect all features among the variables of the data. Then, it employs a biLSTM RNN to interpolate the extrated features along the time dimension; before using a third fully connected dense layer to combine all features to output a realisticness score. Appendix \ref{App:GAN_B_1} reports the technical details on the network's dimensionality and the variable embedding.

%%%===%%%
% Here, again, we are not talking about the novelty of our own work, so we use past tense
\subsubsection*{Training the GAN Model}
We adopted the training objective of \textit{Wasserstein GAN with Gradient Penalty}~\cite{arjovsky2017wasserstein, gulrajani2017improved} (WGAN-GP) to train our GAN model. The networks were updated using
\begin{align}
    &\text{the discriminator loss:}
    \hspace{10mm} L_D =  \underbrace{\mathbb{E}\left[D\big(G(z)\big)\right] - \mathbb{E}\left[D(x_{\text{real}})\right]}_{\text{Wasserstein value function}} + \underbrace{\lambda_{\text{GP}} \mathbb{E}\left[\left(\norm{\nabla_{x_{\text{syn}}}D(x_{\text{syn}})}_2 - 1\right)^2\right]}_{\text{Gradient penalty loss}} \hspace{5mm}\text{and}
    \label{Eq:WGANGP.D}\\
    &\text{the generator loss:}\hspace{15.5mm} L_G =-\mathbb{E}\left[ D(G(z))\right].\label{Eq:WGANGP.G} 
\end{align}
The discriminator network was trained by minimising $L_D$; and likewise the generator network was trained by minimising $L_G$.

The first two terms of Equation (\ref{Eq:WGANGP.D}) form the Wassertein value function~\cite{mallows1972note, levina2001earth} which was constructed through the \textit{Kantorovich-Rubinstein duality} theorem~\cite{villani2008optimal}. This required the theoretical guarantees on the smoothness of network $D$; in practical terms, this was enforced by the gradient penalty loss term to satisfy the Lipschitz continuity with the gradient normality of $1$. Furthermore, the constant $\lambda_{\text{GP}}$ served as a regularisation term that controlled the strength of the gradient penalty loss.

An intuitive interpretation of Equations (\ref{Eq:WGANGP.D}) and (\ref{Eq:WGANGP.G}) can be obtained by noting that for both losses, the component $D\big(G(z)\big)$ is identical to $D(x_{\text{syn}})$. Component $D\big(G(z)\big)$ can hence be conceptualised as a score of the realisticness of the synthetic data. Thus, the generated data is considered more realistic if Equation (\ref{Eq:WGANGP.G}) is minimised. In the discriminator loss of Equation (\ref{Eq:WGANGP.D}), a two-player-game takes place to make it possible to iteratively fine-tune both subnetworks. The Wasserstein value function leverages the discriminator network $D$ as a critic function to compare the realisticness of the synthetic data $D\big(G(z)\big)$ against the ground truth $D(x_{\text{real}})$. While the generator $G$ is trained to fool the discriminator $D$ by maximising the realisticness $\mathbb{E}\left[D\big(G(z)\big)\right]$\footnote{Maximising $\mathbb{E}\left[D\big(G(z)\big)\right]$ is equivalent to minimising the unrealisticness $-\mathbb{E}\left[D\big(G(z)\big)\right]$, of which is Equation (\ref{Eq:WGANGP.G}).}, the discriminator $D$ is fine-tuned to maximise the difference in the realisticness between the real data and the synthetic data $\mathbb{E}\left[D(x_{\text{real}})\right] - \mathbb{E}\left[D\big(G(z)\big)\right]$\footnote{Maximising $\mathbb{E}\left[D(x_{\text{real}})\right] - \mathbb{E}\left[D\big(G(z)\big)\right]$ is equivalent to minimising $\mathbb{E}\left[D\big(G(z)\big)\right] - \mathbb{E}\left[D(x_{\text{real}})\right]$, of which is the Wassertein value function in Equation (\ref{Eq:WGANGP.D}). Maximising the difference in the realisticness between the real and synthetic data means that the discriminator $D$ considers the real data to be more realistic than the synthetic data.}. This allowed the discriminator to become better at differentiating between real and synthetic data, and in turn yielded a higher loss in Equation (\ref{Eq:WGANGP.G}) to further fine-tune the generator $G$. 

%%%===%%%
% changing back to present tense between of novel work
Prior studies in GANs are mostly focused on generating static images for computer vision tasks. However, our aim for the Health Gym is to generate contiguous time series data. That is, we are concerned with both the realistic distributions of individual variables and the correlation among variables over time. To ensure that correlations among variables are captured correctly by the GAN model, we found it useful to make a slight modification to the generator loss function of Equation (\ref{Eq:WGANGP.G}). We augmented the vanilla generator loss function as
\begin{align}
    L_G =-\mathbb{E}\left[ D(G(z))\right] + \underbrace{\lambda_{\text{corr}}\sum_{i = 1}^n\sum_{j = 1}^{i - 1}\left\lVert r^{(i,j)}_\text{syn} - r^{(i,j)}_\text{real}\right\lVert_{L_{1}}}_{\text{Alignment loss}}
    \label{Eq:OurGLoss}
\end{align}
where the additional term is denoted as the alignment loss. We first calculate the \textit{Pearson's r correlation}~\cite{mukaka2012guide} $r^{(i, j)}$ for every unique pair of variables $X^{(i)}$ and $X^{(j)}$; then the alignment loss is calculated as the $L_1$ loss between the differences in correlations between the synthetic data $r_{\text{syn}}$ and their real counterparts $r_{\text{real}}$. Furthermore, $\lambda_{\text{corr}}$ is a positive constant which serves as a weight to control the strength of the alignment loss. Appendix \ref{App:GAN_B_2} reports more details on the training procedure and on the selection of hyper-parameters. 

%%%===%%%
\section*{Data Records}
All of our synthetic datasets are stored as \textit{comma separated value} (CSV) files and are accessible through the Health Gym website (see \url{healthgym.ai}). The synthetic hypotension and sepsis datasets are currently hosted on PhysioNet\cite{kuo2021synthetic} -- a research resource for complex physiologic signals which also hosts the official MIMIC-III\cite{johnson2016mimic} database. 

All synthetic datasets follow the formats of their real counterparts which we described in \textbf{The Real Datasets} in \textbf{Methods}. This section describes specific properties of the synthetic datasets. Quality assurance tests are reported later in the \textbf{Technical Validation} section.

\subsection*{The Synthetic Hypotension Dataset}
The synthetic hypotension dataset is $21.7$ MB and follows the format of the real hypotension dataset of Gottesman \etal\cite{gottesman2020interpretable} containing $3,910$ synthetic patients. Like its real counterpart, there are $48$ data points per patient representing time series of $48$ hours. There are hence $187,680$ (=$3,910\times48$) records (rows) in total.

The synthetic hypotension data comprises 22 variables (columns). The first 20 variables are listed in Table \ref{Tab:Hypotension} -- there are 9 numeric variables, 4 categorical, and 7 binary variables. The 21$^\text{st}$ variable contains the synthetic patient IDs and the 22$^\text{nd}$ variable indicates the hour in the time series. The units and descriptive statistics of the clinical variables are shown in Table \ref{Tab:Hypotension}. The descriptive statistics column shows the first, second, and third quartiles (\ie the 25$^\text{th}$ percentile, median, and 75$^\text{th}$ percentile) for the numeric variables; and the share, in percentage, of each unique class for the binary and categorical variables.

The information presented in this table corresponds to the distributions of the synthetic variables in Figure \ref{Fig:Hypotension.Validation001}. Several numeric variables (\eg urine, serum creatinine) are right-skewed, whereas binary and categorical variables are heavily class imbalanced. This will likely require variable transformation for downstream machine learning applications. Interested readers may consider our proposed pre-processing scheme in Appendix \ref{App:B_1_2}.

\subsection*{The Synthetic Sepsis Dataset}

The synthetic sepsis dataset is $16$ MB and follows the format of the real sepsis dataset of Komorowski \etal\cite{komorowski2018artificial} containing $2,164$ synthetic patients. The synthetic dataset is designed with 20 data points per patient representing times series of 80 hours of data reported in 4-hour windows ($80=20\times4$). There are hence $43,280$ ($=2,164\times20$) records in total.

The synthetic sepsis dataset contains 46 variables -- the first 44 variables are listed in Tables \ref{Tab:Sepsis_Numeric} and \ref{Tab:Sepsis.NonNumeric}. Similar to the synthetic hypotension dataset, the 45$^{\text{th}}$ variable contains the synthetic patient IDs and the 46$^{\text{th}}$ variable indicates the time steps in the time series. Table \ref{Tab:Sepsis_Numeric} presents the 35 numeric variables along with their units and descriptive statistics (\ie the first, second, and third quartiles). Table \ref{Tab:Sepsis.NonNumeric} lists the 3 binary variables and 6 categorical variables; together with the share, in percentage, of each unique class. Unlike the synthetic hypotension dataset, the sepsis dataset contains two \textit{quasi-identifiers}\cite{el2020evaluating}, age and gender, that may be used to disclose personal information. A disclosure risk assessment is reported in the \textbf{Technical Validation} section.

The distributions of the variables in the dataset are shown in Figures \ref{Fig:Sepsis_Validation001_p1} and \ref{Fig:Sepsis_Validation001_p2}. We also observe several numeric variables in the sepsis dataset are right-skewed and will likely need to be transformed before being used for downstream machine learning applications. Interested readers may consider our proposed pre-processing scheme in Appendix \ref{App:B_1_2}.

There are two types of categorical variables in the sepsis dataset. GCS, for example, is a categorical variable by design -- it is a clinical point-based system to measure a person's level of consciousness. The 5 variables of SpO2, Temp, PTT, PT, and INR were instead originally stored as numeric variables in the MIMIC-III database\cite{johnson2016mimic}. These 5 variables were converted into categorical variables because their original distributions were extremely skewed and it was difficult to apply appropriate power-transformations. We decided to categorise these 5 numeric variables into deciles; the 10 classes of each variable\footnote{The 10 categories correspond to the deciles of the original numeric distributions of the variables in the real sepsis dataset. For instance, category C1 corresponds to values that lie within the 0$^\text{th}$ to the 0.1$^\text{st}$ quantile; and that category C5 corresponds to values that lie within the 0.4$^\text{th}$ to the 0.5$^\text{th}$ quantile.} are reported in Table \ref{Tab:Sepsis.NonNumeric}. 

\subsection*{The Synthetic HIV Dataset}

The synthetic HIV dataset is $44.7$ MB and is similar to the real HIV dataset employed by Parbhoo \etal\cite{parbhoo2017combining}. It contains $8,916$ synthetic patients associated with time series of 60 months. The HIV data are reported in 1-month intervals; and hence there are 60 data points per patient and $534,960$ ($=8,916 \times 60$) records in total.

The synthetic HIV dataset contains 15 variables -- the first 13 variables are listed in Table \ref{Tab:HIV} together with descriptive statistics, and the two remaining variables contain the synthetic patient IDs and the month in the time. There are 3 numeric, 5 binary, and 5 categorical variables. As for the other synthetic datasets, the descriptive statistics include the first, second, and third quartiles for the numeric variables; and the share, in percentage, of each unique class for the binary and categorical variables. This dataset also contains the two quasi identifiers, gender and ethnicity, and a disclosure risk assessment is reported in the \textbf{Technical Validation} section.

The distributions of the variables in the dataset are shown in Figure \ref{Fig:HIV_Validation001}. The numeric variables are all right-skewed and require appropriate transformation before the dataset can be used for further analysis. Interested readers may consider our proposed pre-processing scheme in Appendix \ref{App:B_1_2}. Furthermore, the variables of complementary INI, complementary NNRTI, and extra PI, all have the option of \textit{Not Applied}. This was because the medications in these categories can be substituted by medications from the other classes. A general discussion on medications for ART can be found in Appendix \ref{App:A.3.3}. 

%%%===%%%
\section*{Technical Validation}
This section includes a \textbf{Realisticness Validation Procedure} and a \textbf{Disclosure Risk Assessment}. The first part demonstrates the quality of the generated synthetic datasets; and the second part discusses the potential risk of an adversary learning sensitive information about a real person from the synthetic records. 

Based on previous work on the validation of synthetic medical data \cite{goncalves2020generation}, the \textbf{Realisticness Validation Procedure} serves to confirm that our synthetic datasets fulfil the \textit{fidelity of individual data points} and the \textit{fidelity of the population}. That is, we first ensure that the distributions of individual variables are sufficiently similar between the real and the synthetic datasets. We then check that all correlations between variables and trends over time in the real datasets are mirrored in the synthetic datasets.

In the \textbf{Disclosure Risk Assessment}, we show that while our synthetic datasets are realistic, it remains very unlikely for an adversary to learn any sensitive information about a real person using our synthetic datasets. Based on risk metrics from the \textit{disclosure control}
literature\cite{el2020evaluating}, we will show that our synthetic datasets have a low \textit{identity disclosure} risk and a low \textit{attribute disclosure} risk. Identity disclosure refers to the scenario where an adversary is able to match a synthetic record to a real person; and attribute disclosure occurs when an adversary is able to learn new information about a real person, despite the datasets being anonymised.

\subsection*{Realisticness Validation Procedure}
\begin{figure}[ht]
\centering
\includegraphics[width=\linewidth]{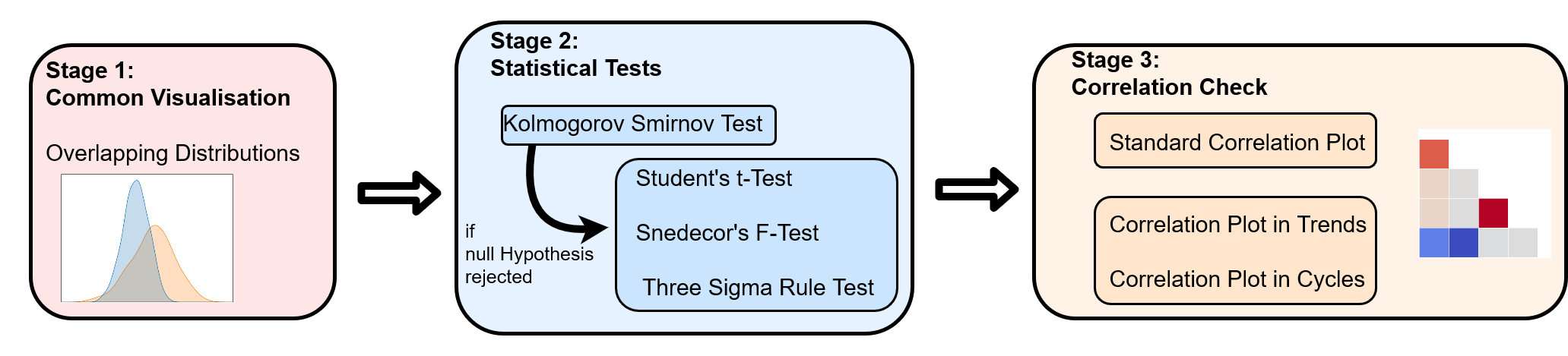}
\caption{A Summary of the Realisticness Validation Procedure\\
The validation includes three stages. First, we perform a qualitative analysis which compares the distributions of real and synthetic variables. Next, we perform a series of statistical tests to assess whether the generated data captured the real data distribution. As a final step, we validate whether the synthetic data captured the correlations between variables over time.}
\label{fig:ValidationSetup}
\end{figure}

Our validation procedure goes beyond prior work\cite{mirza2014conditional, reed2016generative, choi2017generating, zhang2017adversarial} that leveraged GANs to create synthetic data and evaluated the generated data only qualitatively. We summarised the elements of our three-stage validation procedure in Figure \ref{fig:ValidationSetup}. The first two stages analyse the \textit{static} properties of the synthetic data and assess whether the distributions and statistical moments (mean, variance) of the real and synthetic variables are sufficiently similar. Since our generated data are time series, the third stage conducts an additional set of visual comparisons to test the properties of the synthetic variables \textit{over time}.

\subsubsection*{Stage One: Qualitative Analysis}
In the first stage, we superimposed the probability density function of a synthetic numeric variable $X_\text{syn}$ on top of the probability density function of its corresponding real variable $X_\text{real}$. These plots were generated using \textit{kernel density estimations} (KDE)~\cite{davis2011remarks}. Binary and categorical variables were compared by means of side-by-side histogram plots.

\subsubsection*{Stage Two: Statistical Tests}
\hypertarget{Sec:TV-DRC}{The} statistical tests in stage two include the \textit{two-sample Kolmogorov-Smirnov test}\cite{hodges1958significance} (KS test), the \textit{two independent Student’s t-test}\cite{yuen1974two} (t-test), the \textit{Snedecor's F-test}\cite{snedecor1989statistical} (F-test), and the \textit{three sigma rule test}\cite{pukelsheim1994three}. The KS test compares the overall similarity between the distributions of real and synthetic variables. The t-test determines whether there are significant differences between the mean values of the real and synthetic variables; and the F-test compares their variances. Furthermore, the three sigma rule test uses the standard deviations of the real data to check whether the majority of the synthetic data was comprised within a probable range of the real variable values. Definitions and implementations of each test are reported in Appendices \ref{Sec:A.1} - \ref{Sec:A.4}.

We organised the statistical tests in a hierarchical manner. Each synthetic variable (both numeric and categorical) was first assessed using the KS test. The KS test is the most difficult test; and when it was passed, we concluded that a synthetic variable faithfully represents its real counterpart. If a synthetic numeric variable failed the KS test, we applied the t-test, the F-test, and the three sigma rule test. If a synthetic categorical variable failed the KS test, we assessed it further using the \textit{analysis of variance} (ANOVA) F-test and the three sigma rule test. The categorical ANOVA F-test checks the similarity in variances but over different classes. An overview of this procedure can be found in Algorithm \ref{Alg:Stage2} of Appendix \ref{Sec:A.5}.

\subsubsection*{Stage Three: Correlations}
\hypertarget{Sec:TV-CV}{Our} third stage of validation considers correlations between variables and between trends over time, computed using \textit{Kendall's rank correlation coefficients}~\cite{kendall1945treatment}. A brief description is provided below, technical details and a discussion on alternative correlation measures can be found in Appendix \ref{Sec:AppB}.

First, we calculated the \textit{static correlation coefficients} for each pair of variables in the synthetic dataset $X_\text{syn}$ and the real dataset $X_\textit{real}$ (see Appendix \ref{App:B2}).  Next, the correlation coefficients for the two datasets were displayed side-by-side for visual comparison. Ideally, the synthetic dataset should mirror both the \textit{directions} (positive or negative) and \textit{magnitudes} of correlations between variables in the real dataset.

Though informative, the correlation between variables does not provide any information about whether trends over time are captured by the synthetic dataset. Hence, we linearly decomposed each variable as a trend with cycle\cite{hyndman2018forecasting}. The trend indicates the general upward or downward slope of variable over time, and the cycle refers to local periodic patterns. Then, we computed and compared the \textit{average correlation in trends} and \textit{average correlation in cycles} (see Appendix \ref{App:B3}).

\subsubsection*{Validation Outcomes}
\underline{Acute Hypotension}\\
The plots for the first stage of the validation procedure for the hypotension dataset are shown in Figure \ref{Fig:Hypotension.Validation001}. There were no major visual misalignments between the distributions of the real and synthetic datasets, and we proceeded to stage two for statistical confirmations. 

The results of stage two are shown in the hierarchically structured Table \ref{Tab:HypotensionStage2}. The initial KS test was passed by $17$ out of $20$ synthetic variables. The $3$ remaining variables ALT, AST, and PaO2 passed both the t-test and the F-test. This means that these synthetic variables do not perfectly capture the real variable distributions; however, their means and variances are still representative of their real counterparts. These observations are supported by the subplots in Figure \ref{Fig:Hypotension.Validation001}: despite some differences between the real and synthetic data, the overall behaviours are appropriately captured. Furthermore, all of these $3$ variables pass the three sigma rule test. Hence we conclude that all synthetic variables capture the features of the real variable distributions. Appendix \ref{App:HypotensionStatistics} contains the complete statistical results.   

\begin{figure}[ht]
   \centering
   \includegraphics[width=0.8\linewidth]{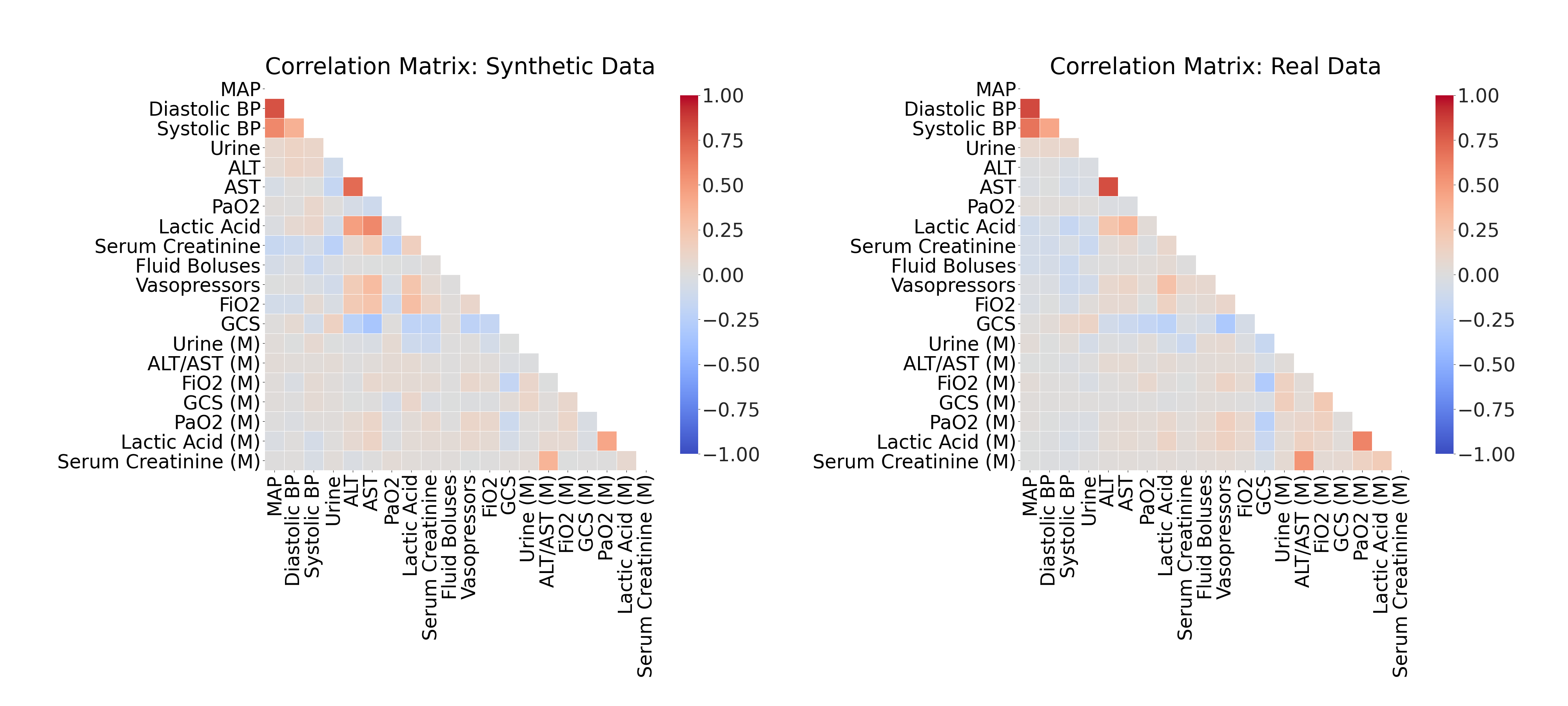}

\caption{The Static Correlations for Acute Hypotension\\
This is a side-by-side comparison of the static correlations in the synthetic dataset and the real dataset. It illustrates the correlation between all pairs of variables, across all patients and timepoints. Positive correlations are coloured in red and negative correlations are in blue. The magnitudes of the correlations are indicated by colour saturation. 
}
\label{Fig:HypotensionCorrelation01} 
\end{figure}

\begin{figure}[ht]
\centering
\begin{subfigure}[b]{0.8\textwidth}
   \includegraphics[width=\linewidth]{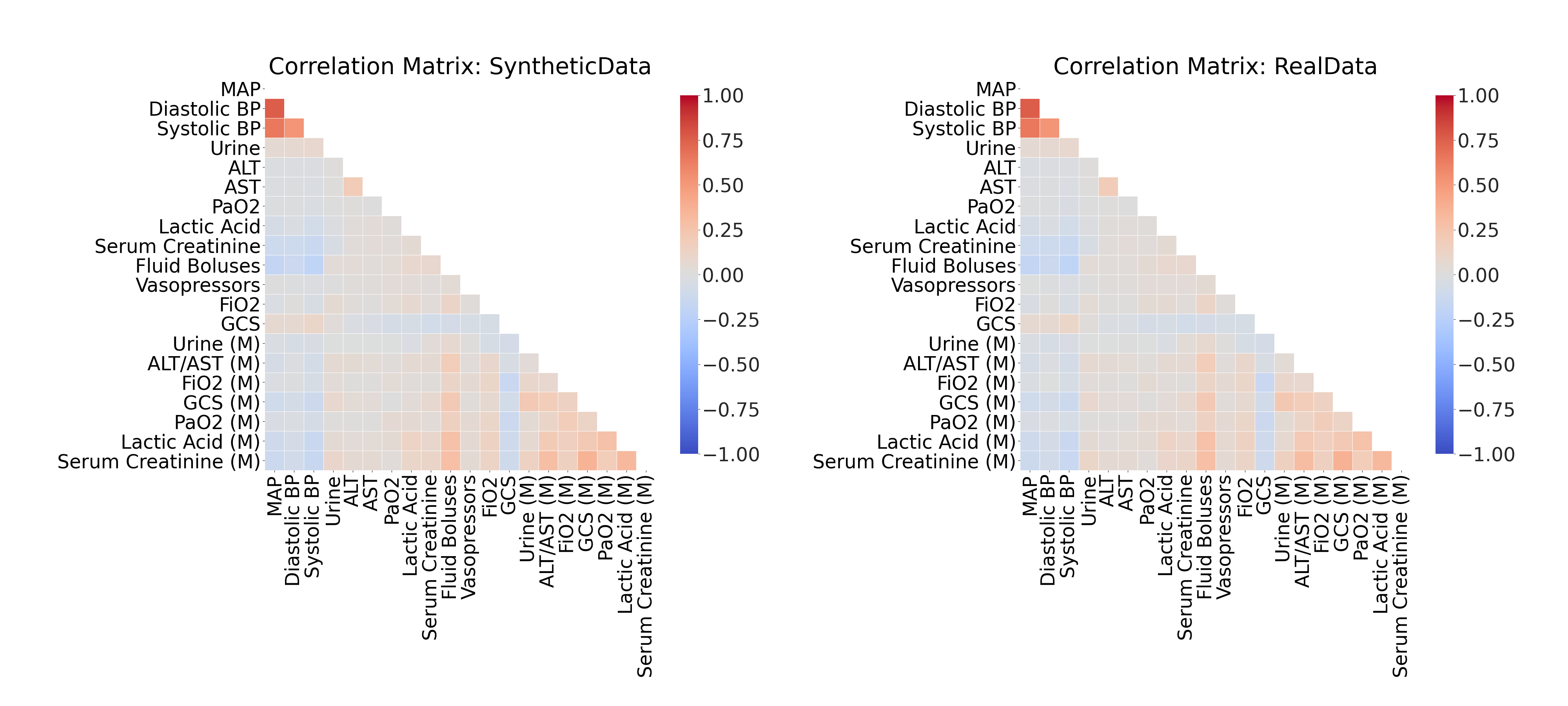}
   \caption{The Average Correlations in Trends}
   \label{Fig:HypotensionCorrelation02}
   \includegraphics[width=\linewidth]{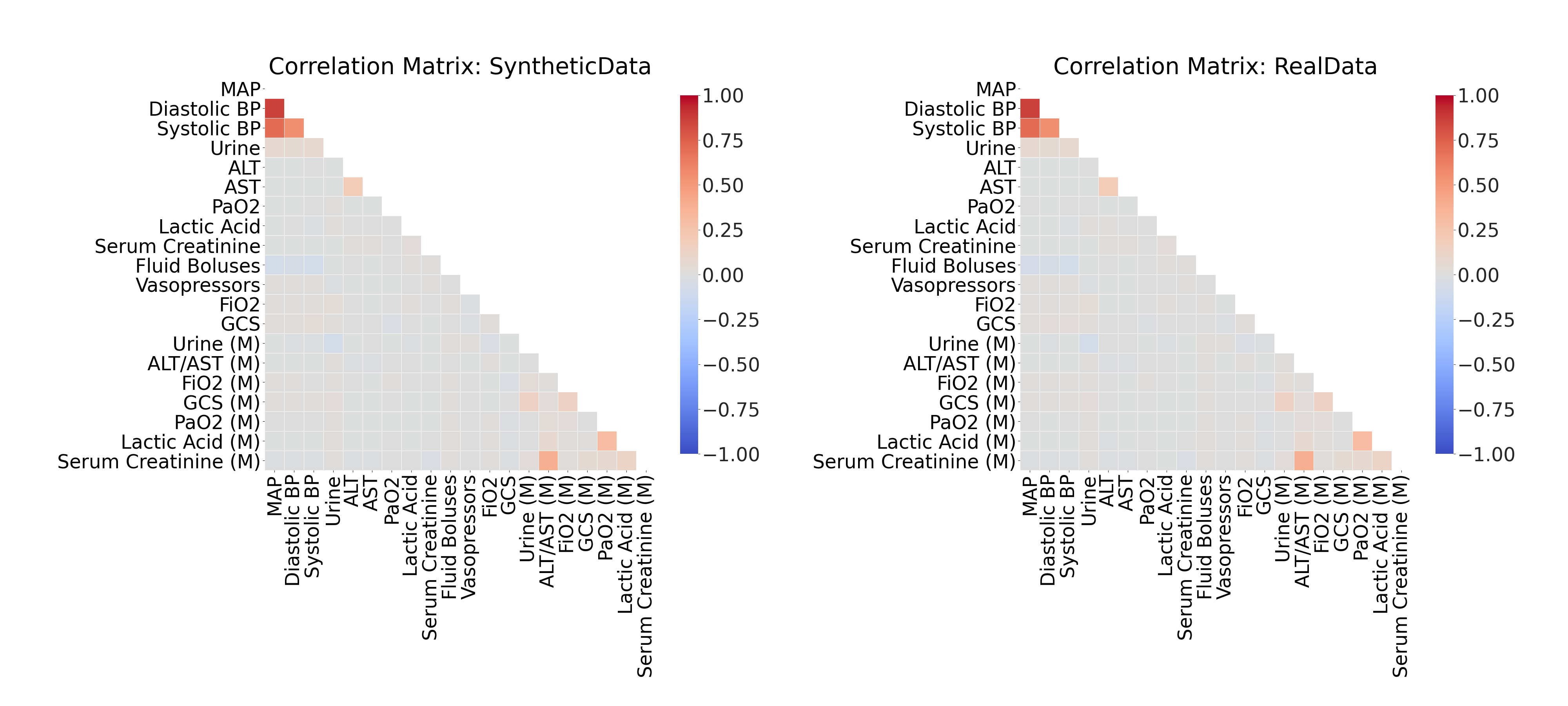}
   \caption{The Average Correlations in Cycles}
   \label{Fig:HypotensionCorrelation03}
\end{subfigure}

\caption{The Dynamic Correlations for Acute Hypotension\\
This is a side-by-side comparison of the dynamic correlations in the synthetic dataset and the real dataset. Unlike the static correlations of Figure \ref{Fig:HypotensionCorrelation01}, all variables are treated as time series and are linearly decomposed into trends and cycles. They illustrate the average correlation between all pairs of variables for each individual patient. Refer to Figure \ref{Fig:HypotensionCorrelation01} for details on the colour scheme.
}
\label{Fig:HypotensionCorrelation_Dynamic} 
\end{figure}

After confirming the realisticness of the individual synthetic variables, we assessed the relations between variables and their longitudinal properties in the third validation stage. We illustrate the static correlations in Figure \ref{Fig:HypotensionCorrelation01} and the dynamic correlations in Figure \ref{Fig:HypotensionCorrelation_Dynamic}. There is no major misalignment between the static correlations of the real and synthetic datasets. However, the synthetic dataset slightly increases the magnitudes of some correlations. For instance, there is a stronger positive correlation between lactic acid and AST in the synthetic dataset than in the real counterpart. Likewise, there is a stronger negative correlation between the synthetic variables of serum creatinine and urine than in the real pair of variables. Nonetheless, the generated data is still highly reliable. In Figure \ref{Fig:HypotensionCorrelation_Dynamic}, the dynamic correlations (both in trends and in cycles) of the decomposed synthetic time series strongly resemble their real counterparts. This indicates that the characteristics of the generated time series variables are realistic. All three stages of our validation confirmed that the synthetic hypotension dataset adequately characterises the properties of the real dataset.\\  

\begin{figure}[ht]
   \centering
   \includegraphics[width=0.8\linewidth]{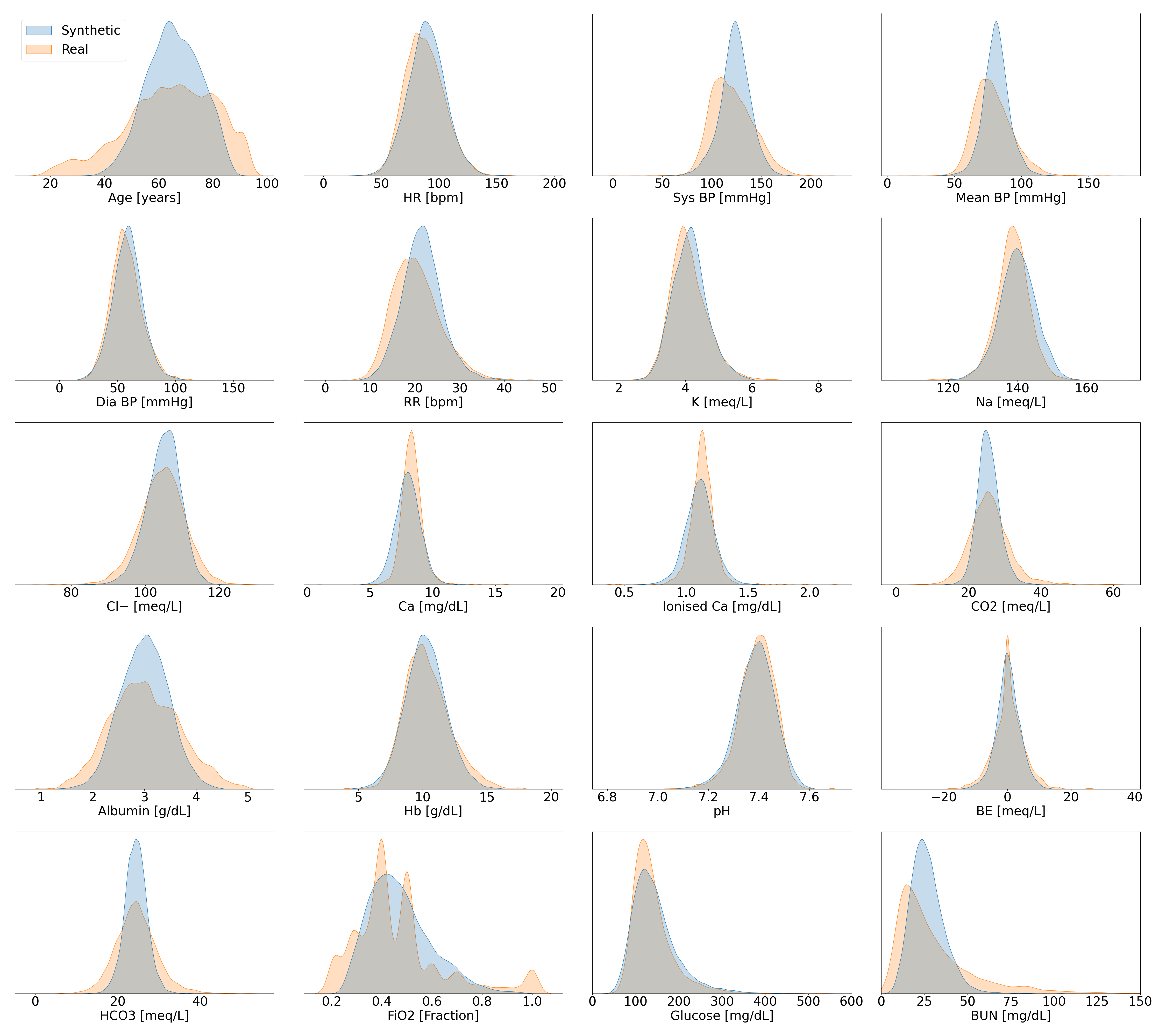}

\caption{Distribution Plots for Sepsis\\
This figure presents visual comparisons between the distributions of variables in the real and synthetic datasets for the management of sepsis. It is continued in Figure \ref{Fig:Sepsis_Validation001_p2}. 
}
\label{Fig:Sepsis_Validation001_p1} 
\end{figure}

\begin{figure}[ht]
   \centering
   \includegraphics[width=0.8\linewidth]{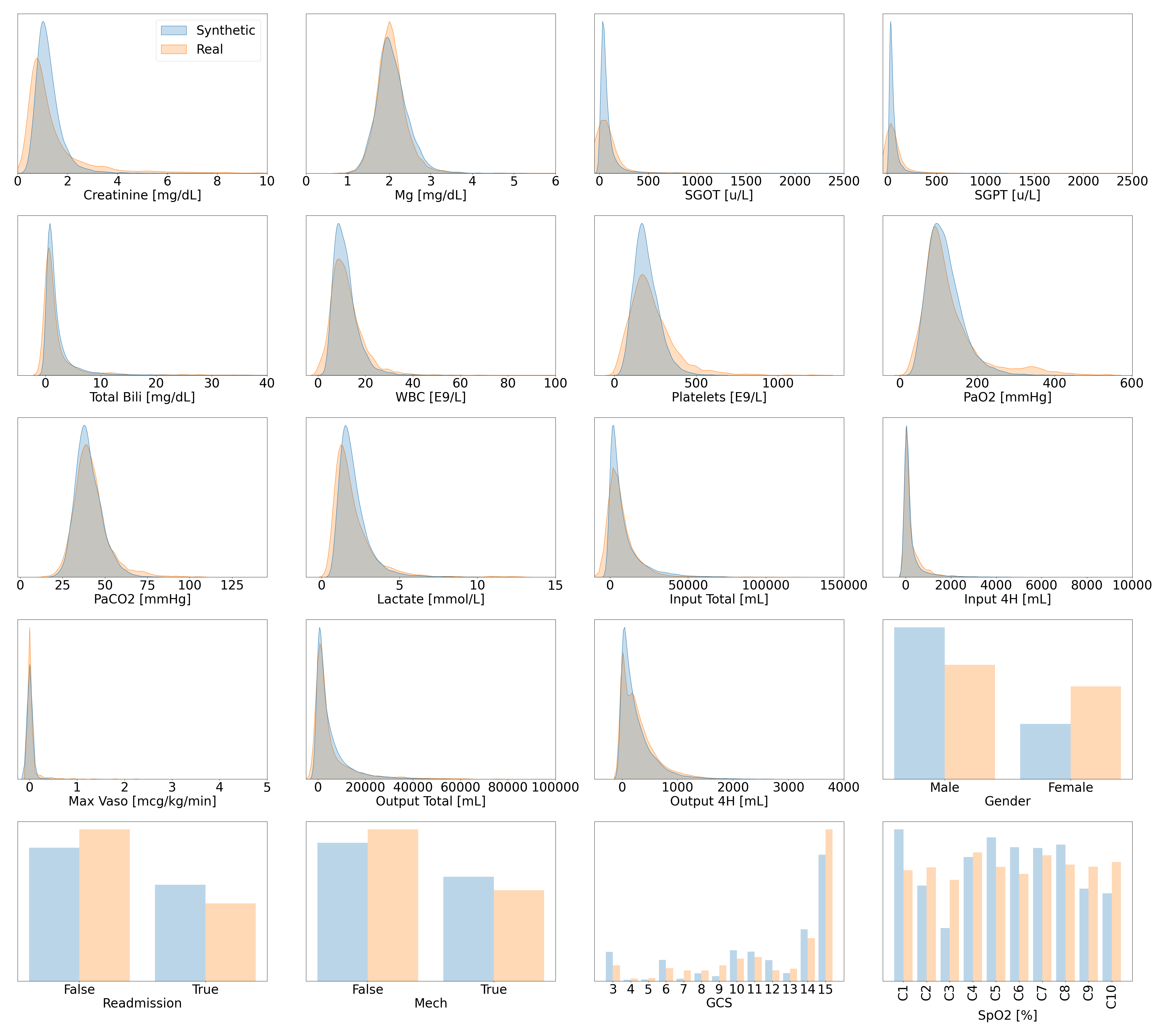}
   \includegraphics[width=0.8\linewidth]{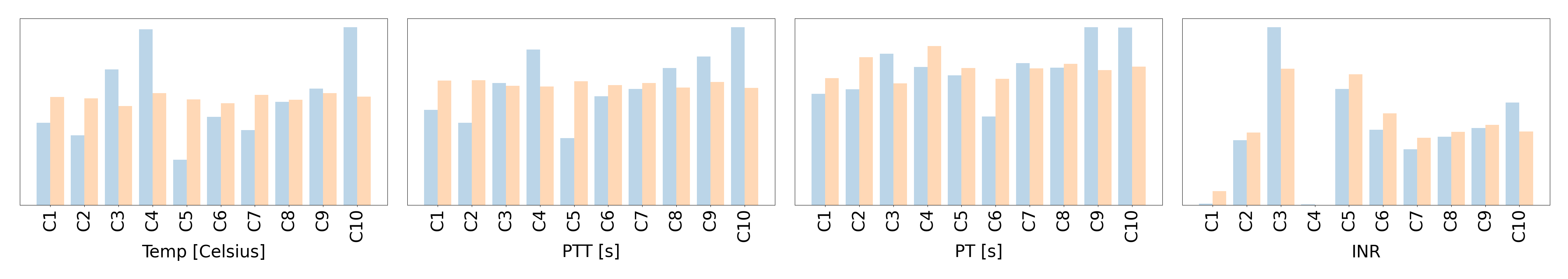}

\caption{Distribution Plots for Sepsis (continued)\\
This figure serves as a continuation to Figure \ref{Fig:Sepsis_Validation001_p1}. All variables are strictly positive but may appear t include negative values as an artefact of using kernel density estimation for plotting the distributions\protect\footnotemark.
\label{FN:PlotArtefect}
}
\label{Fig:Sepsis_Validation001_p2} 
\end{figure}

\hspace{-5mm}\underline{Sepsis}\\
For the synthetic sepsis dataset, we observe in Figures \ref{Fig:Sepsis_Validation001_p1} and \ref{Fig:Sepsis_Validation001_p2} that all synthetic variable distributions were very similar to their real counterparts. In stage two, we found that $43$ out of $44$ variables passed the KS test and therefore almost all synthetic variables mirrored the distributions of their real counterparts. The only variable that failed the KS test was Max Vaso. Since the variable also failed the following F-test, this was because of the differences in the variance (see Table \ref{Tab:SepsisStage2}). As shown in Figure \ref{Fig:Sepsis_Validation001_p2}, Max Vaso is highly skewed. As discussed in the \textbf{Data Records} section, we could have transformed Max Vaso into a categorical variable but decided to keep it as numeric because the closely related variables Input Total, Input 4H, Output Total, and Output 4H are all numeric. These 5 variables collectively describe the input/output measurements of the patients and should therefore share one common data type. Nonetheless, Max Vaso did pass the three sigma rule test. This indicated that while there was a difference in variance for the synthetic Max Vaso variable, the generated data were within the plausible range of the real data. The complete results of all statistical tests are reported in Appendix \ref{App:SepsisStatistics}.   

\begin{figure}[ht]
   \centering
   \includegraphics[width=0.625\linewidth]{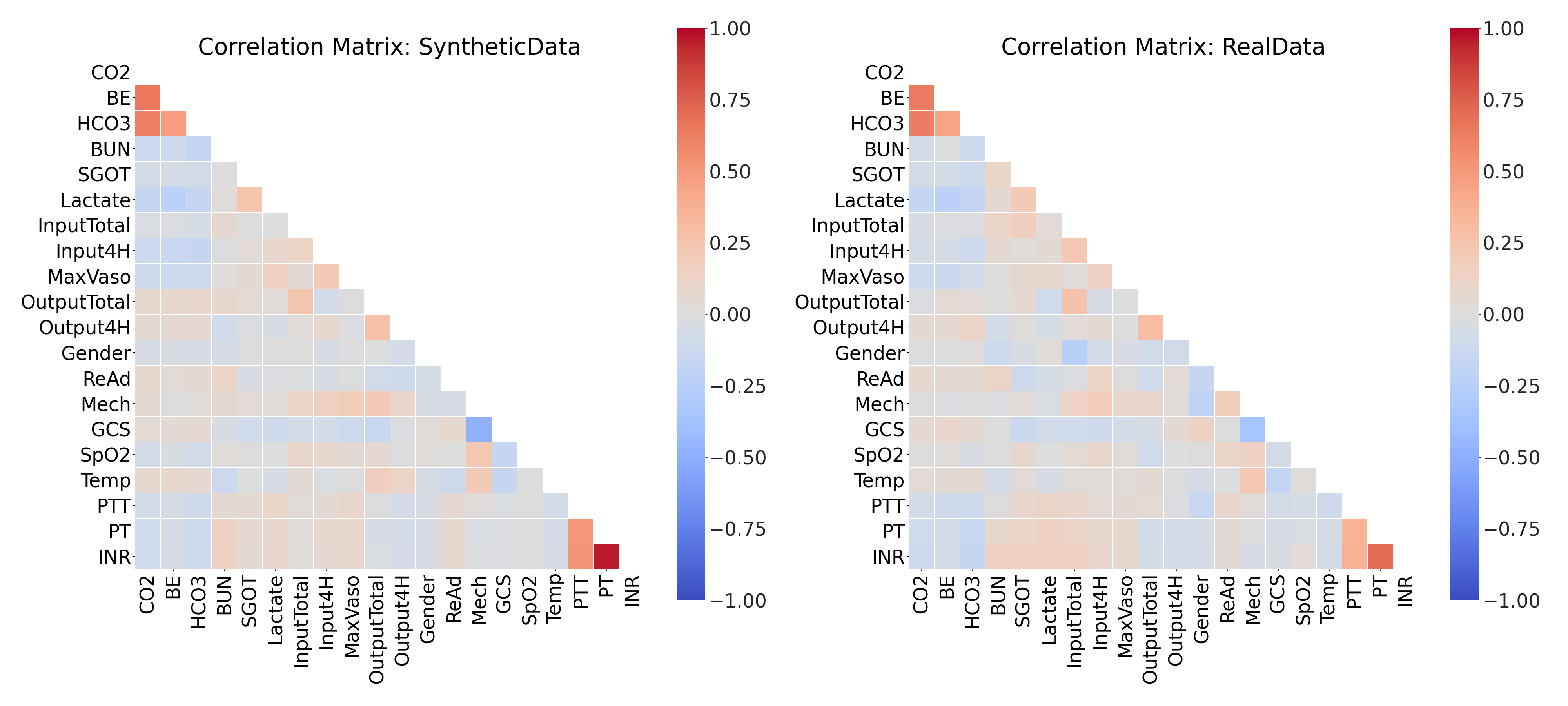}

\caption{The Top 20 Static Correlations for Sepsis\\
This figure presents the static correlations between a subset of the variables in the sepsis dataset. It follows the format of Figure \ref{Fig:HypotensionCorrelation01}; the full correlation plots for all variables are shown in Figure \ref{Fig:Sepsis_Complete_Static}.
}
\label{Fig:Sepsis_Static} 
\end{figure}

\begin{figure}[ht]
\centering
\begin{subfigure}[b]{0.625\textwidth}
   \includegraphics[width=\linewidth]{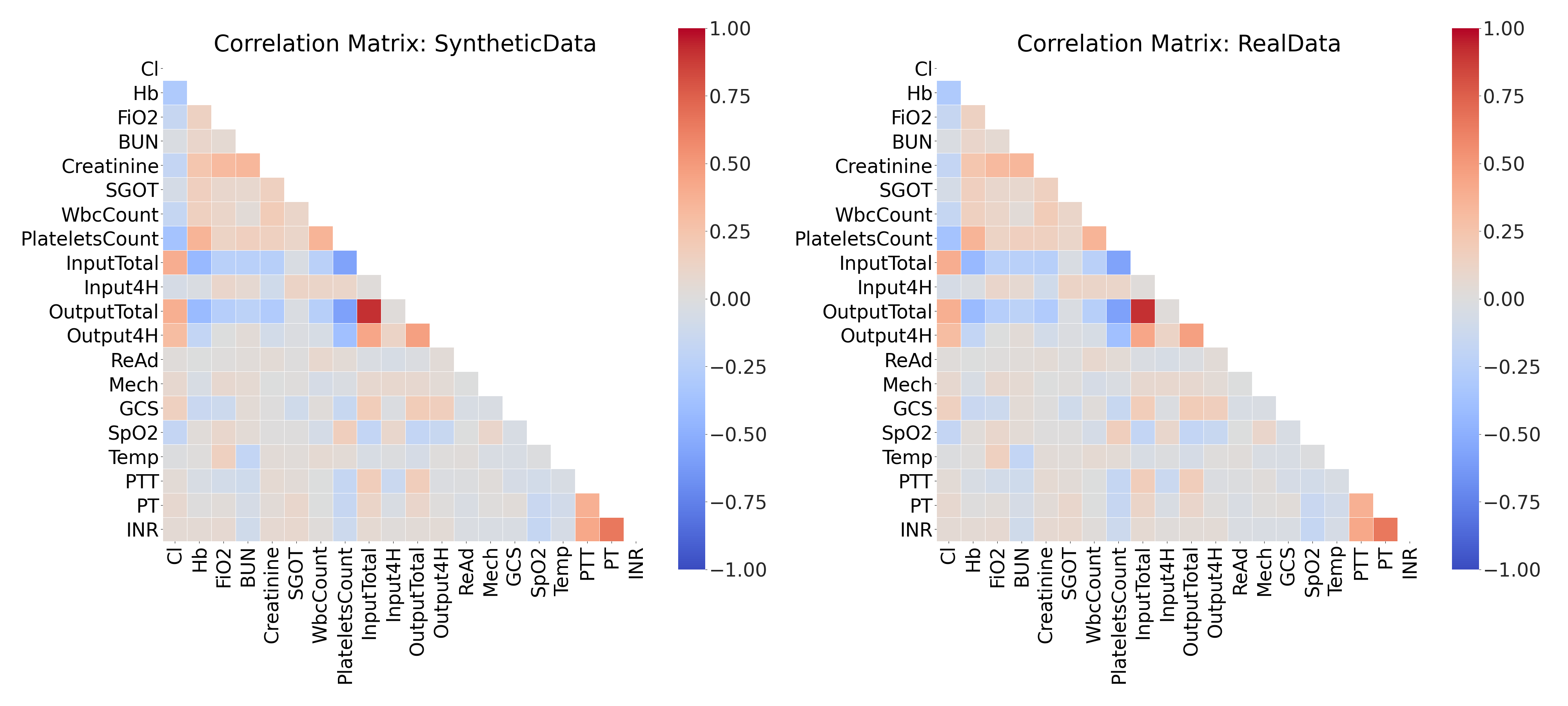}
   \caption{The Top 20 Average Correlations in Trends}
   \label{Fig:SepsisCorrelation02}
   \includegraphics[width=\linewidth]{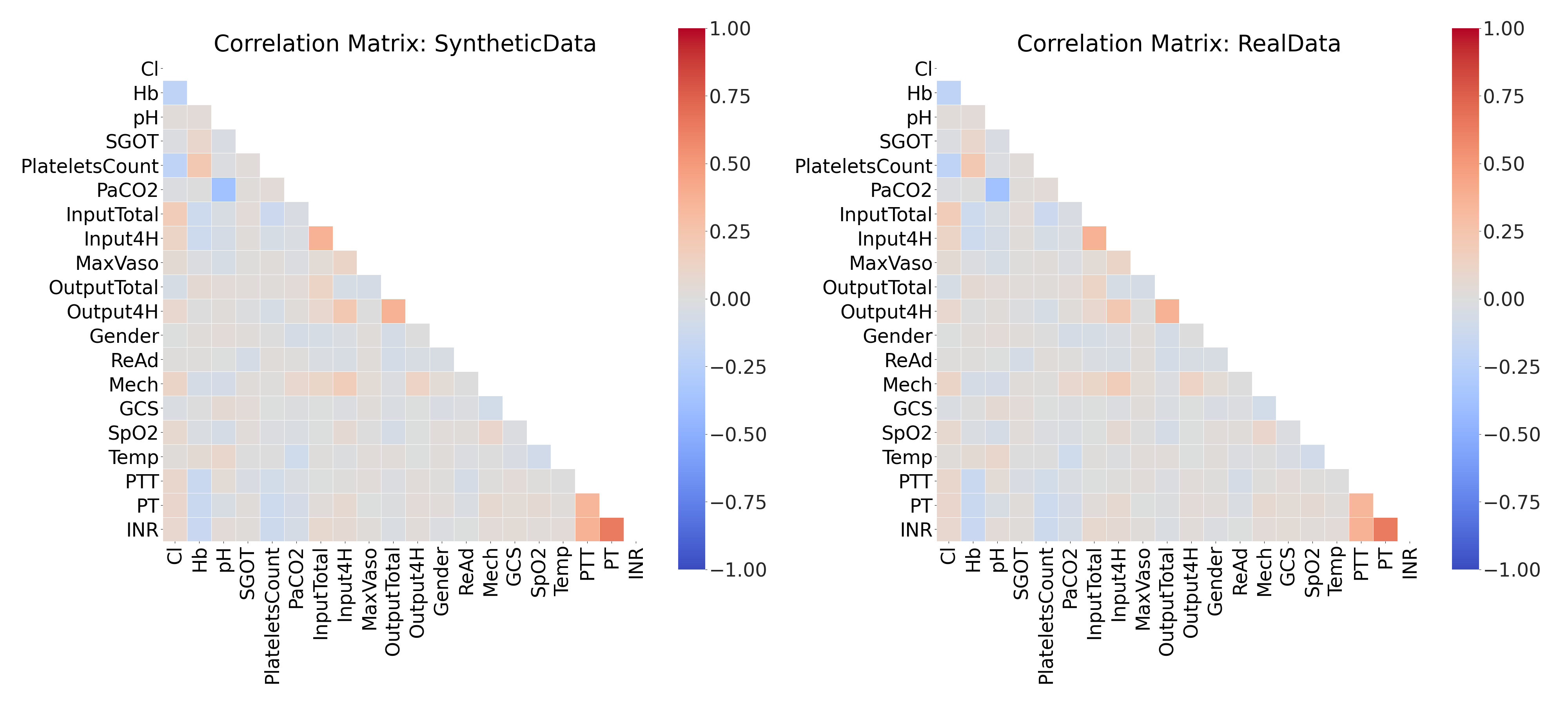}
   \caption{The Top 20 Average Correlations in Cycles}
   \label{Fig:SepsisCorrelation03}
\end{subfigure}

\caption{The Top 20 Dynamic Correlations for Sepsis\\
This figure presents the dynamic correlations between a subset of the variables in the sepsis dataset. It follows the format of Figure \ref{Fig:HypotensionCorrelation_Dynamic}; the full correlation plots for all variables are shown in Figures \ref{Fig:Sepsis_Complete_Trends} and \ref{Fig:Sepsis_Complete_Cycles}.
}
\label{Fig:Sepsis_Dynamic} 
\end{figure}

The correlations computed in stage three of the validation procedure are visualised in Figures \ref{Fig:Sepsis_Static} and \ref{Fig:Sepsis_Dynamic} for a subset of the 20 variables that were associated with the strongest correlations. Both the static and the dynamic correlations were very similar between the real and synthetic dataset. Figures \ref{Fig:Sepsis_Complete_Static} -- \ref{Fig:Sepsis_Complete_Cycles} in Appendix \ref{App:Z004} show the full correlation matrices for all variable pairs.\\  

\begin{figure}[ht]
   \centering
   \includegraphics[width=0.8\linewidth]{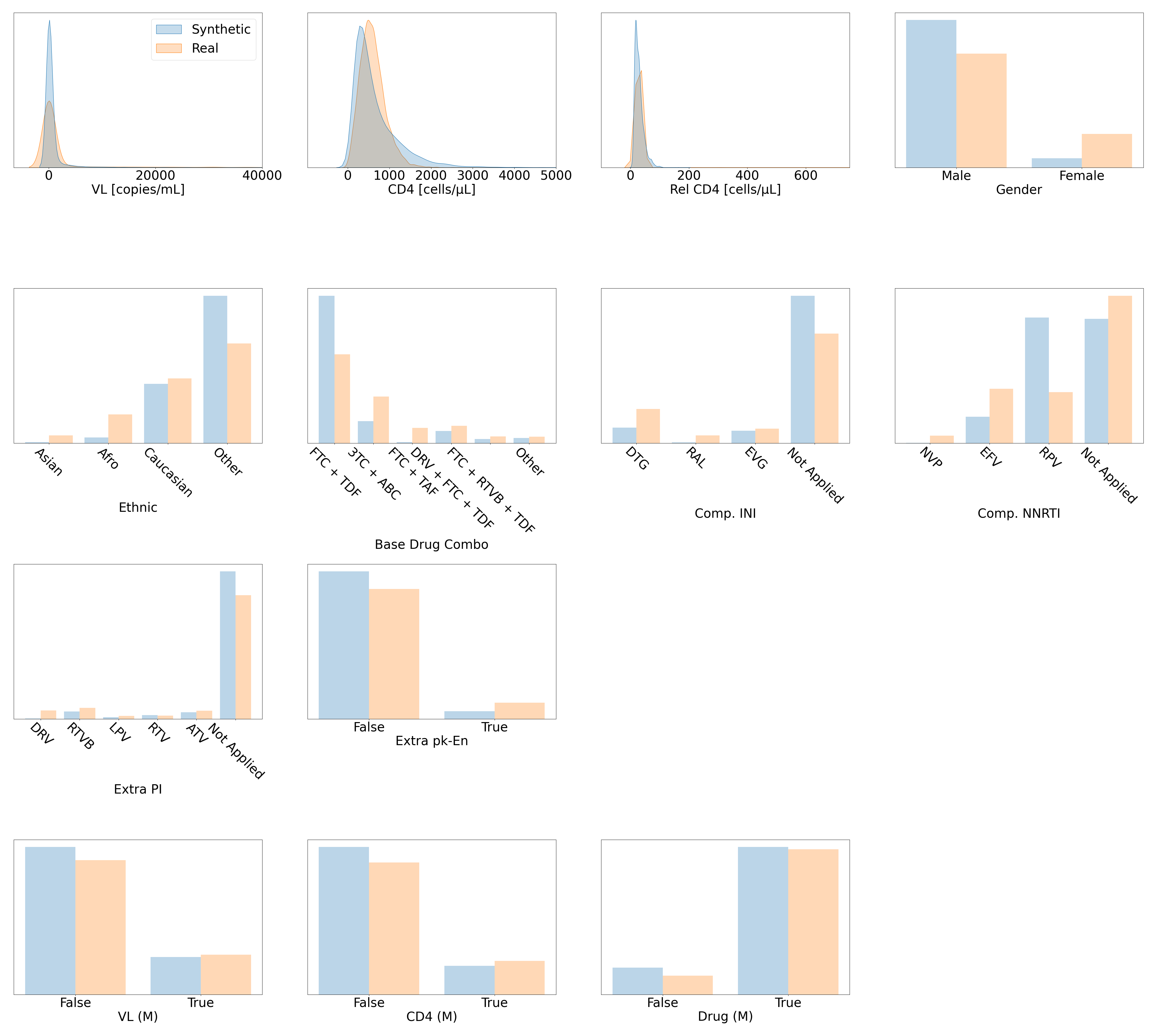}

\caption{Distribution Plots for HIV\\
This figure presents visual comparisons between the distributions of variables in the real and synthetic datasets for the optimisation of antiretroviral therapy for HIV. 
}
\label{Fig:HIV_Validation001} 
\end{figure}

\hspace{-5mm}\underline{HIV}\\
Qualitative comparisons between the distributions of the real and synthetic HIV datasets are shown in Figure \ref{Fig:HIV_Validation001}, indicating high similarity. As presented in Table \ref{Tab:HIVStage2}, $12$ out of $13$ variables passed the KS test, suggesting that the distributions of most synthetic variables matched their real counterparts. The only variable that failed the KS test is VL. VL also failed the F-test, similarly to Max Vaso in the synthetic sepsis dataset. However, VL still passed the three sigma rule test and therefore we can conclude that all variables in the synthetic dataset are highly realistic. Appendix \ref{App:HIVStatistics} contains the complete statistical results.  

\begin{figure}[ht]
   \centering
   \includegraphics[width=0.625\linewidth]{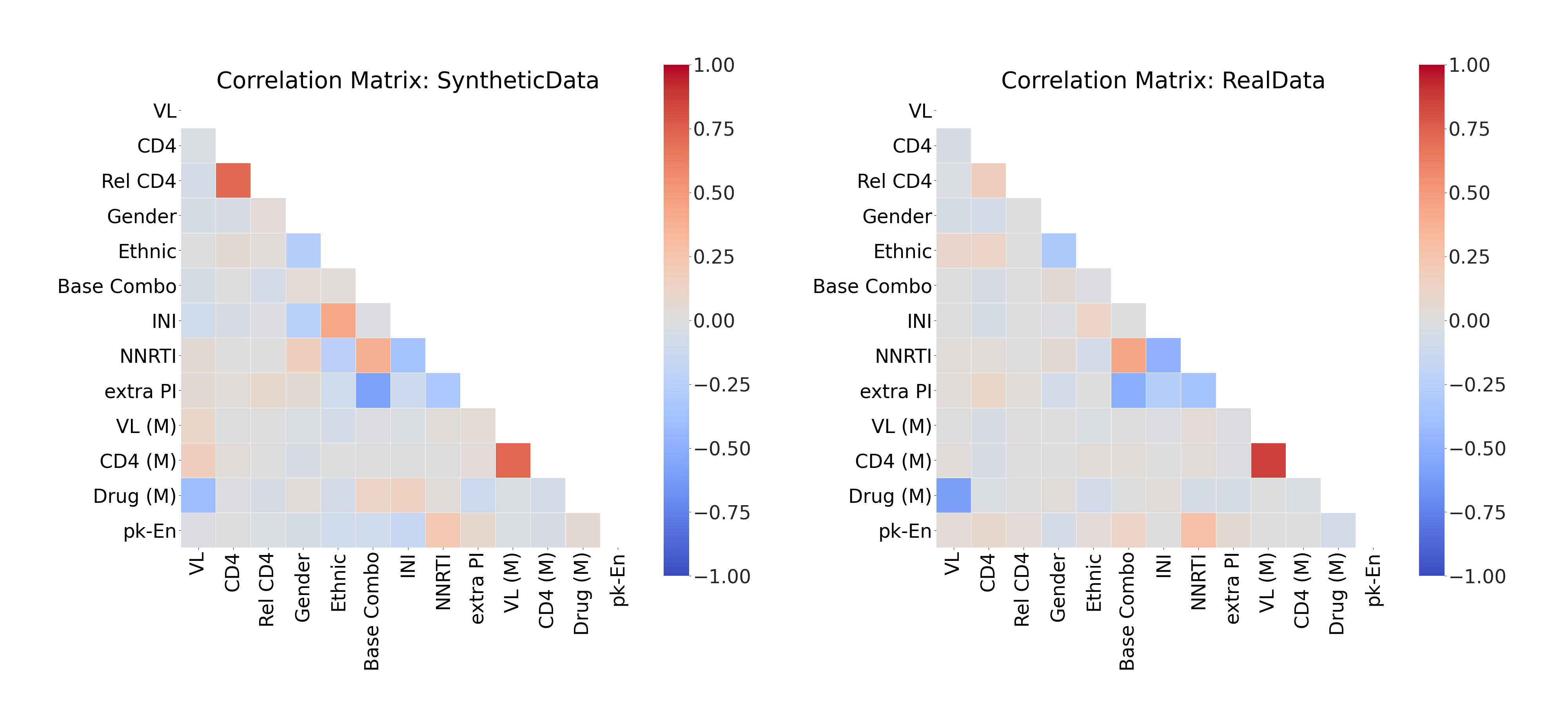}

\caption{The Static Correlations for HIV\\
This figure presents the static correlations between the variables in the HIV dataset. It follows the format of Figure \ref{Fig:HypotensionCorrelation01}.
}
\label{Fig:HIV_Static} 
\end{figure}

\begin{figure}[ht]
\centering
\begin{subfigure}[b]{0.625\textwidth}
   \includegraphics[width=\linewidth]{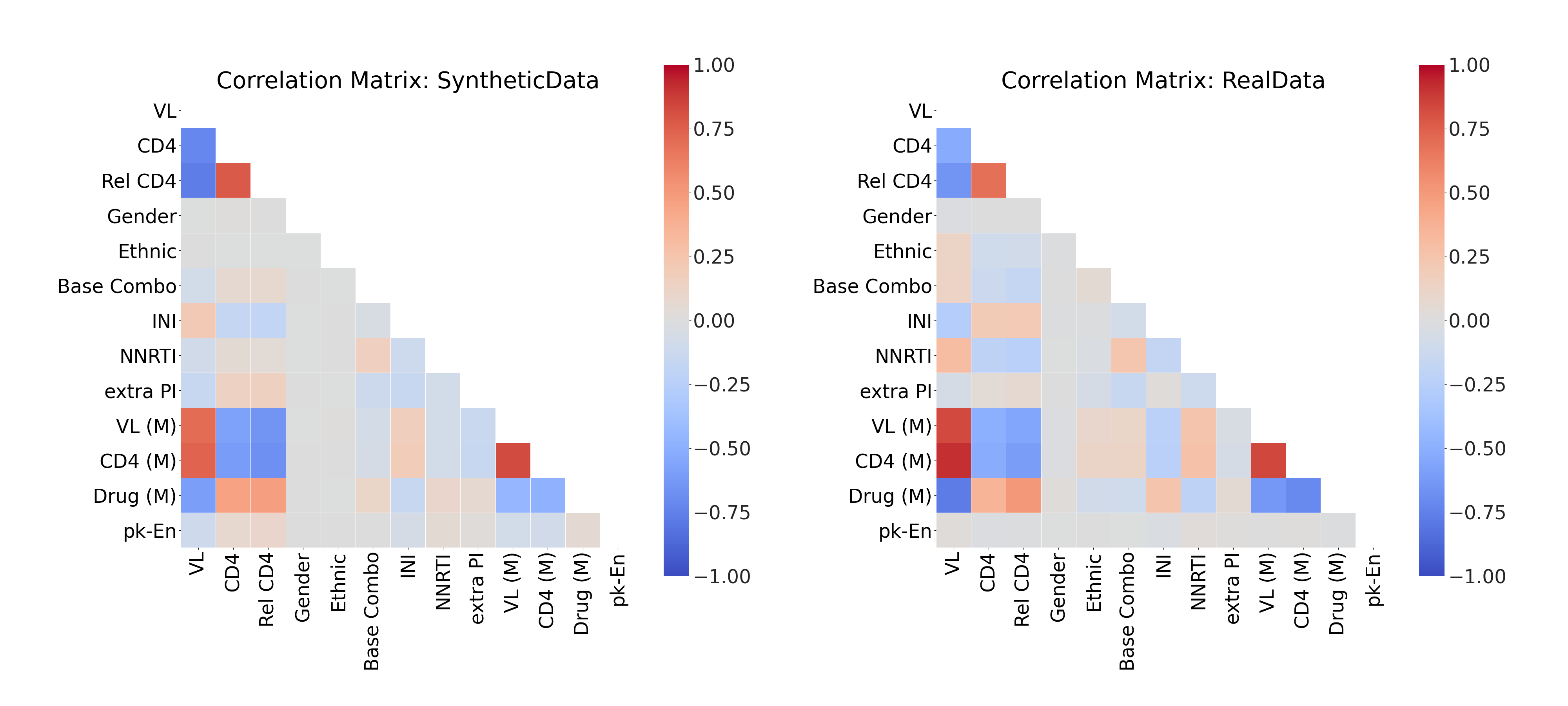}
   \caption{The Average Correlations in Trends}
   \label{Fig:HIVCorrelation02}
   \includegraphics[width=\linewidth]{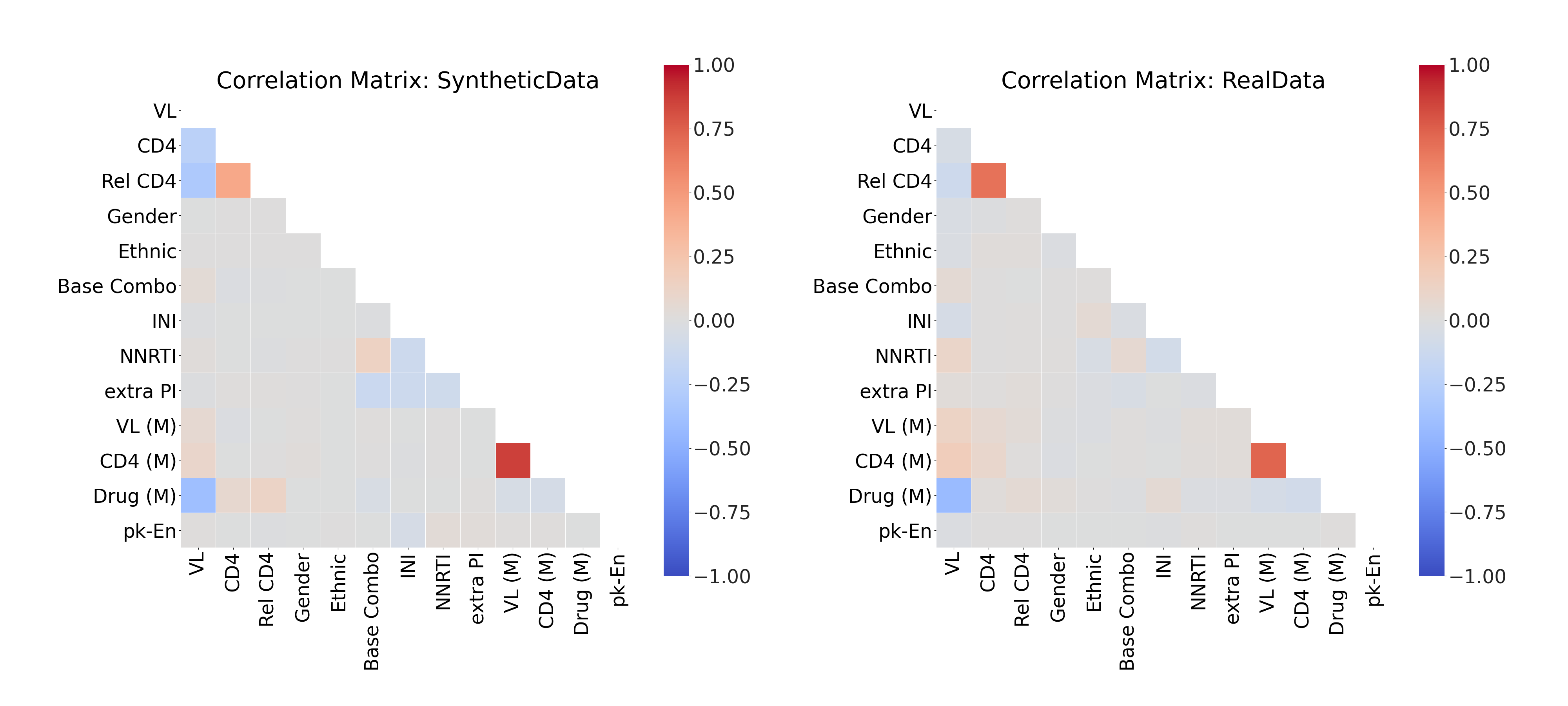}
   \caption{The Average Correlations in Cycles}
   \label{Fig:HIVCorrelation03}
\end{subfigure}

\caption{The Dynamic Correlations for HIV\\
This figure presents the dynamic correlations between the variables in the HIV dataset. It follows the format of Figure \ref{Fig:HypotensionCorrelation_Dynamic}.
}
\label{Fig:HIV_Dynamic} 
\end{figure}

For stage three, we present the correlations in Figures \ref{Fig:HIV_Static} and \ref{Fig:HIV_Dynamic}. Both the static and dynamic correlations reflect that the synthetic dataset captures the relations among the variables in the real dataset.

\subsection*{Disclosure Risk Assessment}
\hypertarget{Sec:TV-SCR}{We} performed two tests to evaluate the likelihood of an attacker learning sensitive information about an individual from the generated synthetic datasets. 

\subsubsection*{Euclidean Distances}
The first test was to ensure that no records in the real datasets were simply copied by the GAN to the synthetic datasets. We computed the Euclidean distances ($L_2$ norms) between records in the real dataset $X_{\text{real}}$ and records in the synthetic dataset $X_{\text{syn}}$. We verified that all distances were greater than zero, \ie that no records in the synthetic datasets perfectly matched any records in the real datasets.

\subsubsection*{Disclosure Risks}
The second test concerned the \textit{disclosure risks} associated with the public distribution of the synthetic datasets. Despite being anonymised, the data may contain sets of variables (\eg age and gender) which, in combination, may be used by an adversary to uniquely identify a person (\eg via linking the data with voter registration lists\cite{benitez2010evaluating}). Variables which in combination constitute personally identifying information are known as \textit{quasi-identifiers}. Individuals with the same combination of quasi-identifiers (\eg all 21-year-old males) form an \textit{equivalent class}.

El Emam \etal\cite{el2020evaluating} introduced two types of disclosure risks based on the concepts of quasi-identifiers and equivalent classes. Depending on the \text{direction of attack}\cite{elliot1999scenarios}, an adversary may attempt to learn new information about a person either by finding out whether an individual in the population (or database) is also included in the real or synthetic dataset (\textit{population-to-sample attack}) or by linking an individual in the real or synthetic dataset back to the original database (\textit{sample-to-population attack}). 

Whereas El Emam \etal\textcolor{white}{.} assume that the real dataset is sampled randomly from the database, in this study the real datasets were constructed using publicly accessible inclusion and exclusion criteria. Therefore, we assumed that the adversary had access not only to the database (\eg MIMIC-III or EuResist) but also to the real dataset. One of the likely reasons to conduct a population-to-sample attack is to determine whether an individual has a specific condition or illness that led to their inclusion in the dataset. When the inclusion criteria are known, population-to-sample attacks become less relevant than sample-to-population attacks, which may be used to learn additional sensitive information about an individual in the synthetic dataset.

The risk of a successful synthetic-to-real attack (\ie the chance of matching a random individual in the synthetic dataset to an individual in the real dataset) can be computed as 
\begin{align}
    \frac{1}{S}\sum_{s = 1}^{S}\left(\frac{1}{F_{s}} \times I_{s}\right)
    \label{Eq:FakeRisk}
\end{align}
where $S$ is the number of records in the synthetic dataset, $F_{s}$ is the size of the equivalent class in the real dataset that shares the same combination of quasi-identifiers as a specific record $s$ in the synthetic sample, and $I_{s}$ is a binary indicator variable equal to one if at least one real record matches the synthetic records $s$. Interested readers can find more descriptions on El Emam \etal's metric in Appendix \ref{App:Security2}.

To assess the risk of information disclosure, we adopted the acceptable risk threshold value\footnote{In their work, El Emam \etal\cite{el2020evaluating} used $5\%$ instead. See page 9 under the section of Risk Assessment Parameters in the referenced work.} of $9\%$ proposed by the European Medicines Agency\cite{european2014european} and Health Canada\cite{canadian2019canadian} for the public release of clinical trial data. Some alternative disclosure risk metrics are discussed in Appendix \ref{App:Security3}.

\subsubsection*{Risk Assessment Outcomes}
\underline{Acute Hypotension}\\
As shown in Table \ref{Tab:Hypotension}, all variables in the synthetic hypotension dataset are associated with the patient's bio-physiological states and do not contain any quasi-identifiers or sensitive information. For this reason, for this dataset we tested the Euclidean distances but not the disclosure risk.

No records in the synthetic dataset completely matched any records in the real hypotension dataset. The smallest distance between any synthetic record and any real record was $49.06$ ($>0$). Therefore no record was leaked into the synthetic dataset.\\

\hspace{-5mm}\underline{Sepsis}\\
Through the Euclidean distance test, we found that no record in the synthetic sepsis dataset was identical to any record in the real sepsis dataset. The smallest distance between real and synthetic records was $328.78$ ($>0$), which was considerably larger than the smallest distance for the hypotension data ($49.06$). This is likely due to the larger number of variables in the sepsis dataset (44 vs 20 in the hypotension dataset, compare Table \ref{Tab:Hypotension} with Tables \ref{Tab:Sepsis_Numeric} and \ref{Tab:Sepsis.NonNumeric}). Furthermore, many sepsis variables are highly skewed (see Figure \ref{Fig:Sepsis_Validation001_p2}) hence exaggerating any value differences. Importantly, for both the synthetic hypotension dataset and the synthetic sepsis dataset, the minimal Euclidean distance is greater than zero. 

The sepsis variables include the quasi-identifiers age and gender. Therefore, age (rounded down to the closest year) and gender were combined to create different equivalence classes \eg all 21-year-old males and all 22-year-old makes were in separate equivalence classes. The risk of a successful synthetic-to-real attack was estimated to be $0.80\%$. This risk is much lower than the suggested threshold of $9\%$\cite{european2014european, canadian2019canadian}, indicating that there is minimal risk of sensitive information disclosure associated with the release of the synthetic sepsis dataset.\\

\hspace{-5mm}\underline{HIV}\\
The minimal Euclidean distance between any pair of real and synthetic HIV records was $0.11$ ($>0$); and hence no data leaked from the real dataset into the synthetic dataset. This value is relatively low for two reasons: 1) there are very few variables in the HIV dataset; 2) most variables are either binary or categorical. The reasons that inflate the Euclidean distance for sepsis are thus the same reasons that deflate the Euclidean distance for HIV.   

The HIV variables include the quasi-identifiers gender and ethnicity. These two variables were combined to create different equivalence classes (\eg male Asian and female Caucasian). The risk of a successful synthetic-to-real attack was estimated to be $0.041\%$. This risk is again much lower than the typical $9\%$ threshold, indicating that also the synthetic HIV dataset can be released with minimal risk of sensitive information disclosure. 

\section*{Code availability
}
The software codes related to the Health Gym project is publicly available at\\
\url{https://github.com/Nic5472K/ScientificData2021_HealthGym}.

%%%===%%%
\newpage
\section*{Figures \& Tables}
%%%===
\begin{table}[ht]
    \centering
    \begin{tabular}{|l||l|l|l|}
        \hline
        \textbf{Variable Name} & 
        \textbf{Data Type} & \textbf{Unit} &
        \textbf{Descriptive Statistics}\\
        
        \hline
        \hline
        Mean Arterial Pressure (MAP) & 
        \cellcolor{cyan!10}numeric & mmHg & \small{Median: $65.34$ \hspace{1.5mm}\quad(Q1: $59.30$, \hspace{1.625mm}Q3: $71.19$)}\\
        
        \hline
        Diastolic Blood Pressure (BP) & 
        \cellcolor{cyan!10}numeric & mmHg & \small{Median: $54.33$ \hspace{1.5mm}\quad(Q1: $48.37$, \hspace{1.5mm}Q3: $60.26$)}\\

        \hline
        Systolic BP & 
        \cellcolor{cyan!10}numeric & mmHg & \small{Median: $113.21$ \quad(Q1: $104.23$, Q3: $121.60$)}\\

        \hline
        Urine & 
        \cellcolor{cyan!10}numeric & mL & \small{Median: $106.21$ \quad(Q1: $68.92$, \hspace{1.5mm}Q3: $164.23$)}\\

        \hline
        Alanine Aminotransferase (ALT) & 
        \cellcolor{cyan!10}numeric & IU/L & \small{Median: $32.55$ \hspace{1.5mm}\quad(Q1: $24.59$, \hspace{1.5mm}Q3: $46.09$)}\\

        \hline
        Aspartate Aminotransferase (AST) & 
        \cellcolor{cyan!10}numeric & IU/L & \small{Median: $46.82$ \hspace{1.5mm}\quad(Q1: $35.81$, \hspace{1.5mm}Q3: $67.75$)}\\

        \hline
        Partial Pressure of Oxygen (PaO2) & 
        \cellcolor{cyan!10}numeric & mmHg & \small{Median: $103.02$ \quad(Q1: $91.34$, \hspace{1.5mm}Q3: $114.66$)}\\

        \hline
        Lactate & 
        \cellcolor{cyan!10}numeric & mmol/L & \small{Median: $1.50$ \hspace{3.125mm}\quad(Q1: $1.29$, \hspace{3.0625mm}Q3: $1.80$)}\\

        \hline
        Serum Creatinine &  
        \cellcolor{cyan!10}numeric & mg/dL & \small{Median: $1.11$ \hspace{3.125mm}\quad(Q1: $0.83$, \hspace{3.0625mm}Q3: $1.62$)}\\
        
        \hline
        \hline
        Fluid Boluses &  
        \cellcolor{magenta!10}categorical & mL & \small{4 Classes}\\
        & & & 
        \small{$[0, 250): 97.32\%$; \hspace{3mm}\quad$[250, 500): 0.28\%$}\\
        & & & 
        \small{$[500, 1000): 1.46\%$; \quad$\ge1000: 0.94\%$}\\
        
        \hline
        Vasopressors &  
        \cellcolor{magenta!10}categorical & mcg/kg/min & \small{4 Classes}\\
        & & & \small{$0: 84.14\%$; \hspace{1.75mm}\quad\quad\quad\quad$(0, 8.4): 8.34\%$}\\
        & & & \small{$[8.4, 20.28): 3.68\%$; \quad$\ge20.28: 3.83\%$}\\
        
        \hline
        Fraction of Inspired Oxygen (FiO2) &  
        \cellcolor{magenta!10}categorical & fraction & \small{10 Classes}\\
        & & & \small{$\le0.2: 0.00\%$; \hspace{4mm}\quad$0.2: 0.54\%$}\\
        & & & \small{$0.3: 2.84\%$; \hspace{7.125mm}\quad$0.4: 10.85\%$}\\
        & & & \small{$0.5: 63.30\%$; \hspace{5.625mm}\quad$0.6: 8.58\%$}\\
        & & & \small{$0.7: 1.32\%$; \hspace{7.125mm}\quad$0.8: 0.20\%$}\\
        & & & \small{$0.9: 2.63\%$; \hspace{7.125mm}\quad$1.0: 9.75\%$}\\
        
        \hline
        Glasgow Coma Scale Score (GCS) &  
        \cellcolor{magenta!10}categorical & point & \small{13 Classes}\\
        & & & \small{$3: 6.61\%$ \quad$4: 2.16\%$ \quad$5: 0.00\%$ \quad$6: 3.00\%$}\\
        & & & \small{$7: 4.77\%$ \quad$8: 0.00\%$ \quad$9: 2.22\%$ \quad$10: 4.32\%$}\\
        & & & \small{$11: 2.46\%$ \quad$12: 3.56\%$ \quad$13: 1.00\%$}\\
        & & & \small{$14: 9.80\%$ \quad$15: 60.09\%$}\\
        
        \hline
        \hline
        Urine Data Measured (Urine (M)) &  
        \cellcolor{brown!10}binary & - - & \small{False: $63.07\%$ \quad True: $36.93\%$}\\
        
        \hline
        ALT or AST Data Measured (ALT/AST (M)) &  
        \cellcolor{brown!10}binary & - - & \small{False: $98.50\%$ \quad True: $1.50\%$}\\
        
        \hline
        FiO2 (M) &  
        \cellcolor{brown!10}binary & - - & \small{False: $92.49\%$ \quad True: $7.51\%$}\\
        
        \hline
        GCS (M) &  
        \cellcolor{brown!10}binary & - - & \small{False: $81.49\%$ \quad True: $18.51\%$}\\
        
        \hline
        PaO2 (M) &  
        \cellcolor{brown!10}binary & - - & \small{False: $97.56\%$ \quad True: $2.44\%$}\\
        
        \hline
        Lactic Acid (M) &  
        \cellcolor{brown!10}binary & - - & \small{False: $96.98\%$ \quad True: $3.02\%$}\\
        
        \hline
        Serum Creatinine (M) &  
        \cellcolor{brown!10}binary & - - & \small{False: $95.26\%$ \quad True: $4.74\%$}\\

        \hline
    \end{tabular}
    
    \caption{\label{Tab:Hypotension}Variables in the Acute Hypotension Dataset\\
This table presents the variables shared by the real and synthetic datasets for the management of acute hypotension. The data types are colour-coded with cyan for numeric, magenta for categorical, and brown for binary. Those variables with suffix (M) indicate whether a data point has been measured (which is usually highly informative in medical time series). The descriptive statistics in this table are \textit{only} for the synthetic dataset$^\text{\ref{FN:StatisticsForRealDatasets}}$. For the numeric variables, we list the median as well as the first and third quantiles (Q1 and Q3). As for the categorical and binary variables, we report the share of each unique class in the synthetic dataset. The information in this table should be compared with the illustrations of Figure \ref{Fig:Hypotension.Validation001}.
}
\end{table}

%%%===
\newpage
\begin{table}[ht]
    \centering
    \begin{tabular}{|l||l|l|l|l|l|}
        \hline
        \textbf{Variable Name} & 
        \textbf{Data Type} & \textbf{Unit} &
        \multicolumn{3}{l|}{\textbf{Descriptive Statistics}}\\
        \hline
        & & & \textbf{Median} & \textbf{Q1} & \textbf{Q3}\\
        
        \hline
        \hline
        Age & 
        \cellcolor{cyan!10}numeric & year & $65.40$ & $58.29$ & $72.95$\\
        \hline
        
        Heart Rate (HR) & 
        \cellcolor{cyan!10}numeric & bpm & $89.09$ & $78.46$ & $99.82$\\
        \hline
        
        Systolic BP & 
        \cellcolor{cyan!10}numeric & mmHg & $123.67$ & $114.43$ & $133.03$\\
        \hline
        
        Mean BP & 
        \cellcolor{cyan!10}numeric & mmHg & $81.02$ & $75.18$ & $86.91$\\
        \hline
        
        Diastolic BP & 
        \cellcolor{cyan!10}numeric & mmHg & $58.90$ & $50.40$ & $66.95$\\
        \hline
        
        %%%===%%%
        Respiratory Rate (RR) & 
        \cellcolor{cyan!10}numeric & bpm & $21.46$ & $18.69$ & $24.28$\\
        \hline
        
        Potassium (K$^{+}$) & 
        \cellcolor{cyan!10}numeric & meq/L & $4.12$ & $3.78$ & $4.45$\\
        \hline
 
        Sodium (Na$^{+}$) & 
        \cellcolor{cyan!10}numeric & meq/L & $140.01$ & $136.59$ & $143.57$\\
        \hline
        
        Chloride (Cl$^{-}$) & 
        \cellcolor{cyan!10}numeric & meq/L & $105.23$ & $102.08$ & $108.03$\\
        \hline
        
        %%%===%%%
        Calcium (Ca$^{++}$) & 
        \cellcolor{cyan!10}numeric & mg/dL & $8.02$ & $7.37$ & $8.66$\\
        \hline
        
        Ionised Ca$^{++}$ & 
        \cellcolor{cyan!10}numeric & mg/dL & $1.11$ & $1.04$ & $1.18$\\
        \hline
        
        Carbon Dioxide (CO2) & 
        \cellcolor{cyan!10}numeric & meq/L & $25.27$ & $23.44$ & $27.29$\\
        \hline
        
        Albumin & 
        \cellcolor{cyan!10}numeric & g/dL & $3.01$ & $2.68$ & $3.32$\\
        \hline
        
        %%%===%%%
        Hemoglobin (Hb) & 
        \cellcolor{cyan!10}numeric & g/dL & $10.20$ & $9.17$ & $11.23$\\
        \hline
        
        Potential of Hydrogen (pH) & 
        \cellcolor{cyan!10}numeric & - - & $7.39$ & $7.34$ & $7.44$\\
        \hline
        
        Arterial Base Excess (BE) & 
        \cellcolor{cyan!10}numeric & meq/L & $0.16$ & $-2.04$ & $2.48$\\
        \hline
        
        Bicarbonate (HCO3) & 
        \cellcolor{cyan!10}numeric & meq/L & $24.38$ & $22.63$ & $26.13$\\
        \hline
        
        %%%===%%%
        FiO2 & 
        \cellcolor{cyan!10}numeric & fraction & $0.45$ & $0.38$ & $0.55$\\
        \hline
        
        Glucose & 
        \cellcolor{cyan!10}numeric & mg/dL & $134.11$ & $108.21$ & $167.06$\\
        \hline
        
        Blood Urea Nitrogen (BUN) & 
        \cellcolor{cyan!10}numeric & mg/dL & $25.38$ & $19.89$ & $31.92$\\
        \hline
        
        Creatinine & 
        \cellcolor{cyan!10}numeric & mg/dL & $1.13$ & $0.90$ & $1.44$\\
        \hline
        
        %%%===%%%
        Magnesium (Mg$^{++}$) & 
        \cellcolor{cyan!10}numeric & mg/dL & $2.04$ & $1.83$ & $2.29$\\
        \hline
        
        Serum Glutamic Oxaloacetic Transaminase (SGOT) & 
        \cellcolor{cyan!10}numeric & u/L & $50.78$ & $31.53$ & $88.97$\\
        \hline
        
        Serum Glutamic Pyruvic Transaminase (SGPT) &
        \cellcolor{cyan!10}numeric & u/L & $39.99$ & $26.20$ & $65.66$\\
        \hline
        
        Total Bilirubin (Total Bili) &
        \cellcolor{cyan!10}numeric & mg/dL & $1.19$ & $0.66$ & $2.32$\\
        \hline
        
        %%%===%%%
        White Blood Cell Count (WBC) &
        \cellcolor{cyan!10}numeric & E9/L & $10.60$ & $7.99$ & $13.92$\\
        \hline
        
        Platelets Count (Platelets) &
        \cellcolor{cyan!10}numeric & E9/L & $184.44$ & $141.97$ & $239.41$\\
        \hline
        
        PaO2 &
        \cellcolor{cyan!10}numeric & mmHg & $109.07$ & $84.22$ & $139.63$\\
        \hline
        
        Partial Pressure of CO2 (PaCO2) &
        \cellcolor{cyan!10}numeric & mmHg & $39.32$ & $34.92$ & $44.97$\\
        \hline
        
        %%%===%%%
        Lactate &
        \cellcolor{cyan!10}numeric & mmol/L & $1.82$ & $1.41$ & $2.40$\\
        \hline
        
        Total Volume of Intravenous Fluids (Input Total) &
        \cellcolor{cyan!10}numeric & mL & $4867.46$ & $1887.84$ & $11155.76$\\
        \hline
        
        Intravenous Fluids of Each 4-Hour Period (Input 4H) &
        \cellcolor{cyan!10}numeric & mL & $58.66$ & $13.83$ & $229.01$\\
        \hline
        
        Maximum Dose of Vasopressors in 4H (Max Vaso) &
        \cellcolor{cyan!10}numeric & mcg/kg/min & $0.0002$ & $0.0$ & $0.0017$\\
        \hline
        
        %%%===%%%
        Total Volume of Urine Output (Output Total) &
        \cellcolor{cyan!10}numeric & mL & $2505.54$ & $585.47$ & $6733.69$\\
        \hline
        
        Urine Output in 4H (Output 4H) &
        \cellcolor{cyan!10}numeric & mL & $159.33$ & $44.74$ & $361.69$\\
        \hline

    \end{tabular}
    
    \caption{\label{Tab:Sepsis_Numeric}Numeric Variables in the Sepsis Dataset\\The format of this table follows that of Table \ref{Tab:Hypotension}; and with more results in Table \ref{Tab:Sepsis.NonNumeric}. Only the first three columns are shared by both the real and synthetic sepsis datasets. The remaining columns show the descriptive statistics that are specific for the synthetic dataset. The content in this table should be compared with the illustrations in Figures \ref{Fig:Sepsis_Validation001_p1} and \ref{Fig:Sepsis_Validation001_p2}.}
\end{table}

\newpage
\begin{table}[ht]
    \centering
    \begin{tabular}{|l||l|l|l|}
        \hline
        \textbf{Variable Name} & 
        \textbf{Data Type} & \textbf{Unit} &
        \textbf{Descriptive Statistics}\\
        
        \hline
        \hline
        Gender &
        \cellcolor{brown!10}binary & - - & \small{Male: $73.41\%$ \quad True: $26.59\%$}\\
        \hline
        
        Readmission of Patient (Readmission) &
        \cellcolor{brown!10}binary & - - & \small{False: $60.20\%$ \quad True: $39.80\%$}\\
        \hline
        
        %%%===%%%
        Mechanical Ventilation (Mech) &
        \cellcolor{brown!10}binary & - - & \small{False: $56.89\%$ \quad True: $43.11\%$}\\
        \hline
        \hline
        
        GCS &
        \cellcolor{magenta!10}categorical & point & 
        \small{13 Classes}\\
        & & & \small{$3: 8.71\%$ \quad$4: 0.38\%$ \quad$5: 0.50\%$ \quad$6: 6.30\%$}\\
        & & & \small{$7: 0.74\%$ \quad$8: 2.27\%$ \quad$9: 1.52\%$ \quad$10: 9.31\%$}\\
        & & & \small{$11: 9.12\%$ \quad$12: 6.31\%$ \quad$13: 2.53\%$}\\
        & & & \small{$14: 15.45\%$ \quad$15: 36.85\%$}\\
        \hline
        
        Pulse Oximetry Saturation (SpO2) &
        \cellcolor{magenta!10}categorical & $\%$ &
        \small{10 Classes (C)}\\
        & & & \small{C1: $[0.00, 93.83): 13.38\%$; \hspace{1.5mm}C2: $[93.83, 95.14): 8.12\%$};\\
        & & & \small{C3: $[95.14, 96.00): 4.48\%$; \hspace{1.5mm}C4: $[96.00, 96.70): 10.64\%$};\\
        & & & \small{C5: $[96.70, 97.33): 12.61\%$; C6: $[97.33, 98.00): 11.36\%$};\\
        & & & \small{C7: $[98.00, 98.60): 11.52\%$; C8: $[98.60, 99.22): 11.84\%$};\\
        & & & \small{C9: $[99.22, 99.86): 8.39\%$; \hspace{1.5mm}C10: $[99.86, 100.0]: 7.66\%$};\\
        \hline
        
        Temperature (Temp) &
        \cellcolor{magenta!10}categorical & Celsius &
        \small{10 Classes (C)}\\
        & & & \small{C1: $[15.11, 35.95): 7.83\%$; \hspace{1.5mm}C2: $[35.95, 36.28): 6.55\%$};\\
        & & & \small{C3: $[36.28, 36.50): 12.87\%$; C4: $[36.50, 36.69): 16.56\%$};\\
        & & & \small{C5: $[36.69, 36.88): 4.21\%$; \hspace{1.5mm}C6: $[36.88, 37.06): 8.21\%$};\\
        & & & \small{C7: $[37.06, 37.28): 7.10\%$; \hspace{1.5mm}C8: $[37.28, 37.56): 9.37\%$};\\
        & & & \small{C9: $[37.56, 37.93): 10.96\%$; C10: $[37.93, 40.52]: 16.33\%$};\\
        \hline
        
        %%%===%%%
        Partial Thromboplastin Time (PTT) &
        \cellcolor{magenta!10}categorical & s & \small{10 Classes (C)}\\
        & & & \small{C1: $[17.80, 24.53): 7.69\%$; \hspace{1.5mm}C2: $[24.53, 26.63): 6.71\%$};\\
        & & & \small{C3: $[26.63, 28.20): 10.02\%$; C4: $[28.20, 29.60): 12.44\%$};\\
        & & & \small{C5: $[29.60, 31.45): 5.46\%$; \hspace{1.5mm}C6: $[31.45, 34.00): 9.27\%$};\\
        & & & \small{C7: $[34.00, 37.10): 9.99\%$; \hspace{1.5mm}C8: $[37.10, 42.80): 11.47\%$};\\
        & & & \small{C9: $[42.80, 57.90): 12.38\%$; C10: $[57.90, 150.00]: 14.58\%$};\\
        \hline
        
        Prothrombin Time (PT) &
        \cellcolor{magenta!10}categorical & s & \small{10 Classes (C)}\\
        & & & \small{C1: $[9.90, 12.20): 7.89\%$; \hspace{3mm}C2: $[12.20, 12.90): 8.2\%$};\\
        & & & \small{C3: $[12.90, 13.30): 11.02\%$; C4: $[13.30, 13.80): 9.84\%$};\\
        & & & \small{C5: $[13.80, 14.30): 9.45\%$; \hspace{1.5mm}C6: $[14.30, 14.90): 6.59\%$};\\
        & & & \small{C7: $[14.90, 15.90): 10.37\%$; C8: $[15.90, 17.51): 10.51\%$};\\
        & & & \small{C9: $[17.51, 22.00): 13.27\%$; C10: $[22.00, 146.70]: 12.85\%$};\\
        \hline
        
        International Normalised Ratio (INR) &
        \cellcolor{magenta!10}categorical & - - & \small{10 Classes (C)}\\
        & & & \small{C1: $[0.00, 1.00): 0.19\%$; \hspace{1.5mm}C2: $[1.00, 1.10): 8.88\%$};\\
        & & & \small{C3: $[1.10, 1.20): 23.35\%$}; C4: $[2.21, 17.60]: 0.09\%$\\
        & & & \small{C5: $[1.20, 1.30): 15.64\%$; C6: $[1.30, 1.31): 10.22\%$};\\
        & & & \small{C7: $[1.31, 1.50): 7.53\%$; \hspace{1.5mm}C8: $[1.50, 1.70): 9.71\%$};\\
        & & & \small{C9: $[1.70, 2.21): 10.67\%$; C10: $[2.21, 17.60]: 13.70\%$};\\
        \hline

    \end{tabular}
    
    \caption{\label{Tab:Sepsis.NonNumeric}Non-Numeric Variables in the Sepsis Dataset\\The format of this table follows that of Table \ref{Tab:Hypotension}; and it is a continuation of Table \ref{Tab:Sepsis_Numeric}.}
\end{table}

%%%===
\newpage
\begin{table}[ht]
    \centering
    \begin{tabular}{|l||l|l|l|}
        \hline
        \textbf{Variable Name} & 
        \textbf{Data Type} & \textbf{Unit} &
        \textbf{Descriptive Statistics}\\
        
        \hline
        \hline
        Viral Load (VL) & 
        \cellcolor{cyan!10}numeric & copies/mL & \small{Median: $54.77$ \hspace{0mm}\quad(Q1: $16.51$, \hspace{3mm}Q3: $209.03$)}\\
        
        \hline
        Absolute Count for CD4 (CD4) & 
        \cellcolor{cyan!10}numeric & cells/$\mu$L & \small{Median: $465.81$ \hspace{-1.625mm}\quad(Q1: $279.26$, \hspace{1.5mm}Q3: $840.34$)}\\

        \hline
        Relative Count for CD4 (Rel CD4) & 
        \cellcolor{cyan!10}numeric & cells/$\mu$L & \small{Median: $25.57$ \quad(Q1: $18.20$, \hspace{3mm}Q3: $35.72$)}\\
        
        \hline
        \hline
        Gender &  
        \cellcolor{brown!10}binary & - - & \small{Male: $93.42\%$ \quad\quad Female: $6.58\%$}\\
        
        \hline
        Ethnicity &  
        \cellcolor{magenta!10}categorical & - - & \small{4 Classes}\\
        & & & \small{Asian: $0.47\%$; 
        \quad\quad\quad\quad \hspace{1mm}African: $2.55\%$}\\
        & & & \small{Caucasian: $26.81\%$; 
        \quad\quad Other: $70.17\%$}\\
        
        \hline
        \hline
        Base Drug Combination &  
        \cellcolor{magenta!10}categorical & - - & \small{6 Classes}\\
        (Base Drug Combo) & & & 
        \small{FTC + TDF: $73.66\%$; 
        \quad\quad 3TC + ABC $14.08\%$}\\
        & & & 
        \small{FTC + TAF: $0.98\%$}\\
        & & & 
        \small{DRV + FTC + TDF: $5.50\%$}\\
        & & & 
        \small{FTC + RTVB + TDF: $2.30\%$}\\
        & & & 
        \small{Other: $3.47\%$}\\
        
        \hline
        Complementary INI &  
        \cellcolor{magenta!10}categorical & - - & \small{4 Classes}\\
        (Comp. INI) & & & 
        \small{DTG: $11.96\%$; 
        \quad RAL: $0.49\%$}\\
        & & & 
        \small{EVG: $4.69\%$; 
        \quad \hspace{1.5mm}Not Applied: $82.86\%$}\\
        
        \hline
        Complementary NNRTI &  
        \cellcolor{magenta!10}categorical & - - & \small{4 Classes}\\
        (Comp. NNRTI) & & & 
        \small{NVP: $0.19\%$; 
        \quad \hspace{1.75mm}EFV: $9.27\%$}\\
        & & & 
        \small{RPV: $43.76\%$; 
        \quad \hspace{0.5mm}Not Applied: $46.78\%$}\\
        
        \hline
        Extra PI &  
        \cellcolor{magenta!10}categorical & - - & \small{6 Classes}\\
        & & & 
        \small{DRV: $0.69\%$ 
        \quad\quad \hspace{-0.5mm}RTVB: $4.02\%$}\\
        & & & 
        \small{LPV: $1.08\%$ 
        \quad\quad RTV: $2.02\%$}\\
        & & & 
        \small{ATV: $4.26\%$ 
        \quad\quad Not Applied: $87.92\%$}\\
        
        \hline
        Extra pk Enhancer (Extra pk-En) &  
        \cellcolor{brown!10}binary & - - & 
        \small{False: $96.70\%$ 
        \quad True: $3.30\%$}\\
        
        \hline
        \hline
        VL Measured (VL (M)) &  
        \cellcolor{brown!10}binary & - - & 
        \small{False: $79.35\%$ 
        \quad True: $20.65\%$}\\
        
        \hline
        CD4 (M) &  
        \cellcolor{brown!10}binary & - - & 
        \small{False: $83.39\%$ 
        \quad True: $16.61\%$}\\
        
        \hline
        Drug Recorded (Drug (M)) &  
        \cellcolor{brown!10}binary & - - & 
        \small{False: $15.56\%$ 
        \quad True: $84.44\%$}\\

        \hline
    \end{tabular}
    
    \caption{\label{Tab:HIV}Variables in the HIV Dataset\\
This table presents the variables shared by the real and synthetic datasets for antiretroviral therapy in HIV. The format of this table follows that of Table \ref{Tab:Hypotension}. Only the first three columns are shared by both the real and synthetic HIV datasets. The last column shows the descriptive statistics that are specific to the synthetic dataset. The content in this table should be compared with the illustrations in Figure \ref{Fig:HIV_Validation001}.
}
\end{table}

%%%===
\newpage
\begin{figure}[ht]
    \begin{subfigure}[b]{\textwidth}
    \centering
       %%%===%%%
       \includegraphics[width=0.7\linewidth]{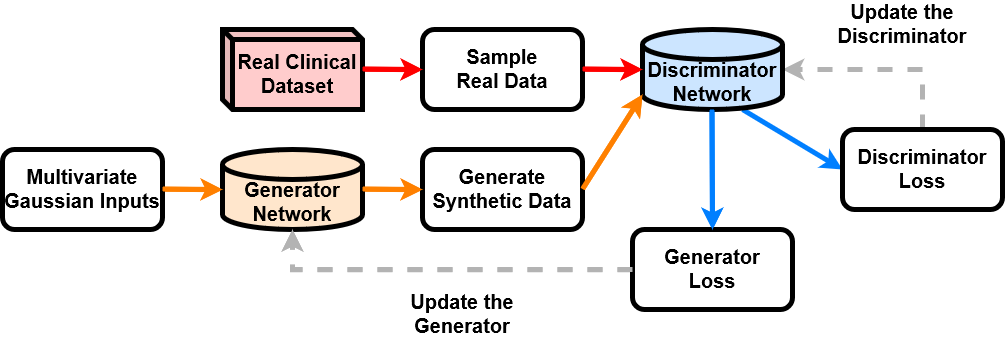}
       \caption{An Overview}
       
       %%%===%%%
       \vspace*{5mm}
       \includegraphics[width=0.7\linewidth]{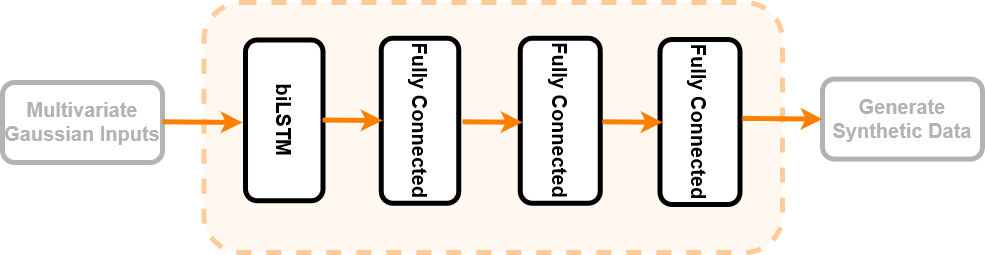}
       \caption{The Generator Network}
       
       %%%===%%%
       \vspace*{5mm}
       \includegraphics[width=0.8\linewidth]{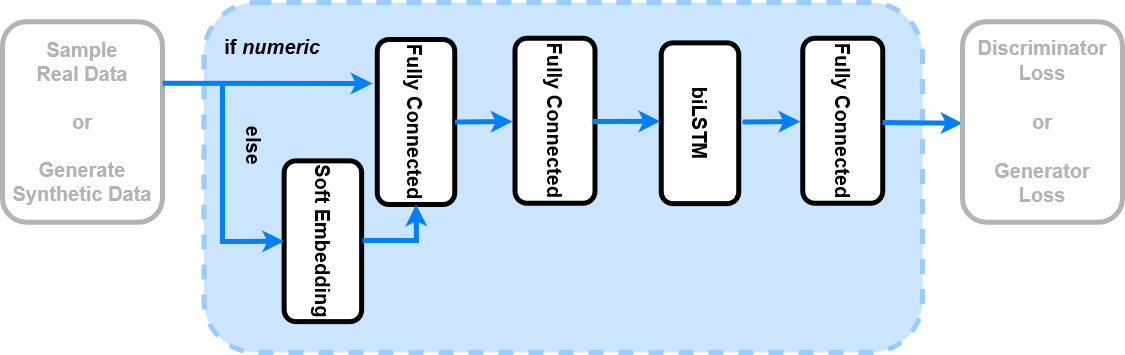}
       \caption{The Discriminator Network}
    \end{subfigure}

    \caption{The GAN Pipeline for the Health Gym Project\\
    The overview of the GAN pipeline is shown in (a). It conjointly trains a generator network which synthesises data, and a discriminator network which aims to classify the data as either real or fake. In (b), we show that the generator consists of one biLSTM layer followed by three fully conneted dense layers. Whereas in (c), the discriminator first embeds non-numeric data, then it passes all input to two fully connected layers, a biLSTM layer, then another fully connected layer.
    }
\label{Fig:GANs} 
\end{figure}

%%%===
\newpage

%%%===
\newpage
\begin{figure}[ht]

   \centering
   \includegraphics[width=0.8\linewidth]{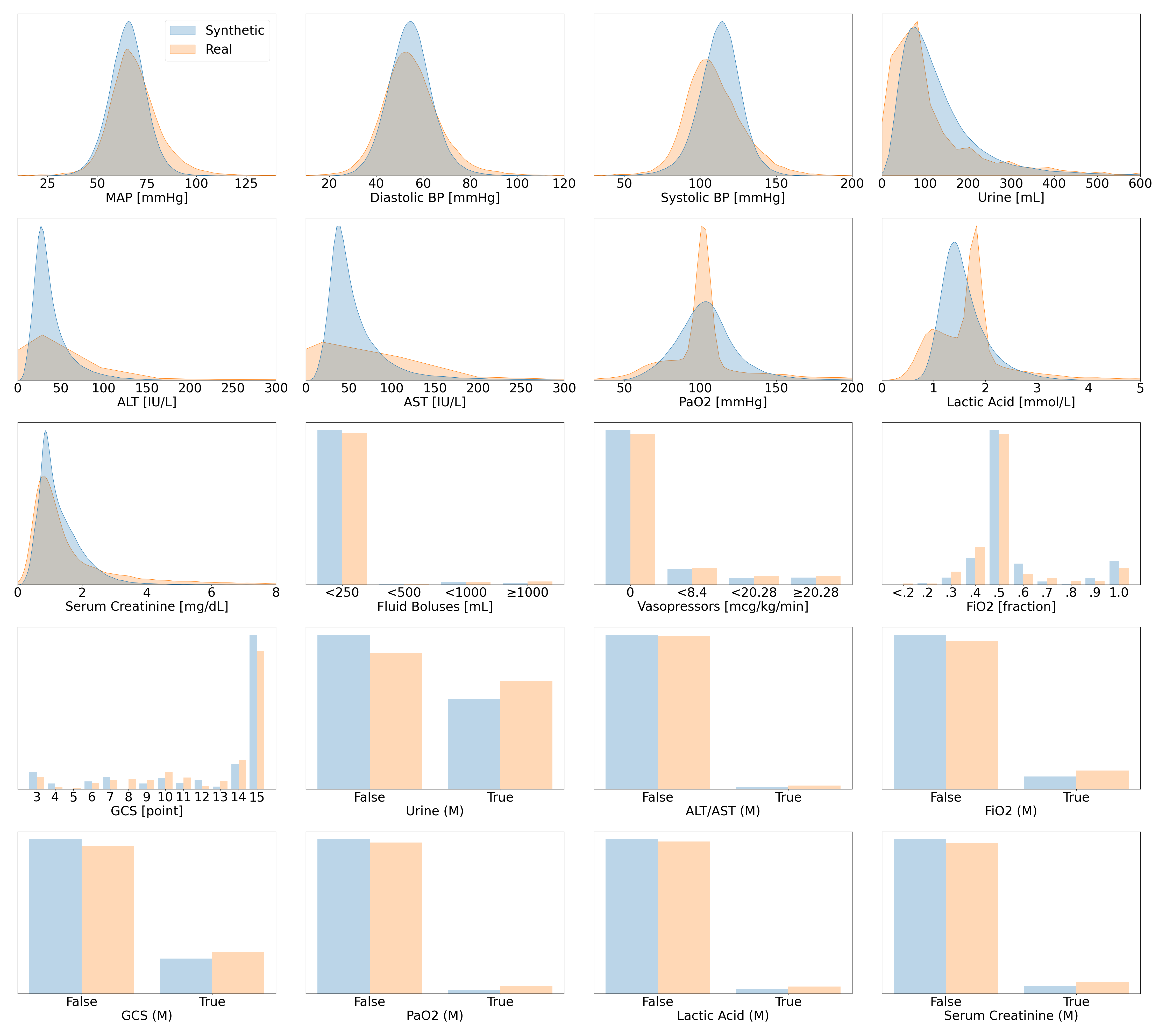}

\caption{Distribution Plots for Acute Hypotension\\
This figure presents visual comparisons between the distributions of variables in the real and synthetic datasets for the management of acute hypotension. The distributions of real variables are plotted in orange and their synthetic counterparts in blue.
}
\label{Fig:Hypotension.Validation001} 
\end{figure}

%%%===
\newpage
\begin{table}[ht]
    \centering
    \begin{tabular}{llll}
        \hline
        \textbf{Passed the KS Test} & \multicolumn{3}{l}{MAP, Diastolic BP, Systolic BP, Serum Creatinine, Fluid Boluses, Vasopressors,}\\
         & \multicolumn{3}{l}{FiO2, GCS, Urine, Lactic Acid, Urine (M), ALT/AST (M), FiO2 (M), GCS (M),}\\
         & \multicolumn{3}{l}{PaO2 (M), Lactic Acid (M), Serum Creatinine (M)}\\
        \hline
        \hline
        \textbf{Failed the KS Test} & \textbf{Variable Name} & \textbf{t-Test Status} & \textbf{F-Test Status}\\
        \hdashline
        & ALT & \cmark & \cmark \\
        & AST & \cmark & \cmark \\
        & PaO2 & \cmark & \cmark \\
        \hline
        \hline
        \textbf{The Three Sigma Rule Test} & \textit{passed}& \multicolumn{2}{l}{ALT, AST, PaO2}\\
        & \textit{failed}& \multicolumn{2}{l}{- -}\\
        \hline

    \end{tabular}
    
    \caption{\label{Tab:HypotensionStage2}The Stage Two Validation Results for Acute Hypotension\\
    This table summarises the results of the statistical tests. The tests were conducted in the order of the KS-test, then the t-test and F-test, and finally the three sigma rule test. Only those variables that failed the KS-test underwent the additional tests. $17$ of the $20$ variables in the synthetic hypotension dataset passed the KS-test and did not have different distributions from their real counterparts. The remaining $3$ variables passed all additional tests. Therefore, all variables of the synthetic hypotension dataset were realistic. This table should be compared with Figure \ref{Fig:Hypotension.Validation001} and Table \ref{Tab:Hypotension}.}
\end{table}

%%%===
\newpage

\newpage

%%%===
\newpage

\newpage

\footnotetext{Also see the related discussion on
Anon. Stack Exchange. (2014).\\Retrieved from \url{https://stats.stackexchange.com/questions/109549/negative-density-for-non-negative-variables}}

%%%===
\newpage
\begin{table}[ht]
    \centering
    \begin{tabular}{llll}
        \hline
        \textbf{Passed the KS Test} & \multicolumn{3}{l}{Age, HR, Systolic BP, Mean BP, Diastolic BP, RR, K, Na, Cl$^-$, Ca,}\\
         & \multicolumn{3}{l}{Ionised Ca, CO2, Albumin, HB, pH, BE, HCO3, FiO2, Glucose, BUN, Creatinine,}\\
         & \multicolumn{3}{l}{Mg, SGOT, SGPT, Total Bili, WBC, Platelets, PaO2,PaCO2, Lactate,}\\
         & \multicolumn{3}{l}{ Input Total, Input 4H, Output Total, Output 4H, Gender, Readmission, Mech, GCS,}\\
         & \multicolumn{3}{l}{SpO2, Temp, PTT, PT, INR}\\
        \hline
        \hline
        \textbf{Failed the KS Test} & \textbf{Variable Name} & \textbf{t-Test Status} & \textbf{F-Test Status}\\
        \hdashline
        & Max Vaso & \cmark & \xmark \\
        \hline
        \hline
        \textbf{The Three Sigma Rule Test} & \textit{passed}& \multicolumn{2}{l}{Max Vaso}\\
        & \textit{failed}& \multicolumn{2}{l}{- -}\\
        \hline

    \end{tabular}
    
    \caption{\label{Tab:SepsisStage2}The Stage Two Validation Results for Sepsis\\
This table presents the statistical results for the synthetic sepsis dataset. It follows the format of Table \ref{Tab:HypotensionStage2}; and should be compared with Figures \ref{Fig:Sepsis_Validation001_p1} and \ref{Fig:Sepsis_Validation001_p2} and Tables \ref{Tab:Sepsis_Numeric} and \ref{Tab:Sepsis.NonNumeric}.}
\end{table}

%%%===
\newpage

\newpage

%%%===
\newpage

%%%===
\newpage
\begin{table}[ht]
    \centering
    \begin{tabular}{llll}
        \hline
        \textbf{Passed the KS Test} & \multicolumn{3}{l}{CD4, Rel CD4, Gender, Ethnic, Base Drug Combo, Comp. INI, Comp. NNRTI, Extra PI}\\
         & \multicolumn{3}{l}{Extra pk-En, VL (M), CD4 (M), Drug (M)}\\
        \hline
        \hline
        \textbf{Failed the KS Test} & \textbf{Variable Name} & \textbf{t-Test Status} & \textbf{F-Test Status}\\
        \hdashline
        & VL & \cmark & \xmark \\
        \hline
        \hline
        \textbf{The Three Sigma Rule Test} & \textit{passed}& \multicolumn{2}{l}{VL}\\
        & \textit{failed}& \multicolumn{2}{l}{- -}\\
        \hline

    \end{tabular}
    
    \caption{\label{Tab:HIVStage2}The Stage Two Validation Results for HIV\\
This table presents the statistical results for the synthetic HIV dataset. It follows the format of Table \ref{Tab:HypotensionStage2}; and should be compared with Figure \ref{Fig:HIV_Validation001} and Table \ref{Tab:HIV}.}
\end{table}

%%%===
\newpage

\newpage

%%%===%%%
\newpage
\bibliography{A000_main}

%%%===%%%
\appendix
\newpage
\section*{Acknowledgements}
We would like to thank Sarah Garland for her contribution to accessing EuResist data and synthetic data validation. This study benefited from data provided by EuResist Network EIDB; and this project has been funded by a Wellcome Trust Open Research Fund (reference number 219691/Z/19/Z).

\section*{Author Contributions Statement}
\textbf{N.K.} and \textbf{S.B.} designed, implemented and validated the deep learning models used to generate the synthetic datasets. \textbf{L.J.} contributed to the design of the study and provided expertise regarding the risk of sensitive information disclosure. \textbf{M.P.} provided clinical expertise on antiretroviral therapy for HIV. \textbf{S.F.} provided clinical expertise on sepsis. \textbf{F.G.}, \textbf{A.S.}, \textbf{M.Z.}, and \textbf{M.B.} contributed patient data as part of the EuResist Integrated Database. Furthermore, \textbf{N.K.} wrote the manuscript and \textbf{S.B.} designed the study. All authors contributed to the interpretation of findings and manuscript revisions.

\section*{Competing Interests}
The authors declare no competing interests.

\newpage
\section*{The Appendices for the Health Gym Paper}

In this study, we presented a novel approach for creating synthetic healthcare datasets (related to acute hypotension\cite{gottesman2019guidelines}, sepsis\cite{komorowski2018artificial}, and HIV\cite{parbhoo2017combining, world2016consolidated}) using the machine learning algorithm of GAN\cite{goodfellow2014generative}. We carried out several statistical tests to verify the realisticness of the synthetic datasets and employed disclosure control techniques\cite{el2020evaluating} to ensure that the public release of these datasets is associated with very low risk of sensitive information disclosure. This appendix provides further technical details and discusses some of the concepts which were presented only briefly in the main body of the paper.\\

\hspace*{5mm} \textbf{Table of Contents}\\
\hspace*{20mm} \ref{Sec:BImpact}: \underline{Future Work and Broader Impact}\\
\hspace*{20mm} \ref{Sec:AppData}: \underline{Instructions on Deriving the Real Datasets}\\
\hspace*{30mm} \ref{App:HypotensionDetail}: \textit{Acute Hypotension Dataset}\\
\hspace*{30mm} \ref{App:Sepsis}: \textit{Sepsis Dataset}\\
\hspace*{30mm} \ref{App:HIV}: \textit{HIV Dataset}\\
\hspace*{20mm} \ref{App:GANsTraining}: \underline{The Training Details of GANs}\\
\hspace*{30mm} \ref{App:GAN_B_1}: \textit{The Dimensionality in the GAN Setup}\\
\hspace*{30mm} \ref{App:GAN_B_2}: \textit{Remarks on Training the GAN Model}\\
\hspace*{20mm} \ref{Sec:AppA}: \underline{The Statistical Tests of Stage 2}\\
\hspace*{30mm} \ref{Sec:A.1}: \textit{The Two Sample KS Test}\\
\hspace*{30mm} \ref{Sec:A.2}: \textit{The Two Independent Sample Student's t-Test}\\
\hspace*{30mm} \ref{Sec:A.3}: \textit{The Snedecor's F-Test}\\
\hspace*{30mm} \ref{Sec:A.4}: \textit{The Three Sigma Rule Test}\\
\hspace*{30mm} \ref{Sec:A.5}: \textit{Iteratively Executing the Statistical Tests}\\
\hspace*{30mm} \ref{App:StatsOutcome}: \textit{The Statistical Outcomes}\\
\hspace*{20mm} \ref{Sec:AppB}: \underline{The Correlations of Stage 3}\\
\hspace*{30mm} \ref{App:B1}: \textit{Kendall's Rank Correlation}\\
\hspace*{30mm} \ref{App:B2}: \textit{Correlation Between Variables}\\
\hspace*{30mm} \ref{App:B3}: \textit{Average Correlation in Trends and in Cycles}\\
\hspace*{30mm} \ref{App:B4}: \textit{Remarks on Kendall's Rank Correlation and Alternative Validation Metrics}\\
\hspace*{20mm} \ref{App:Z004}: \underline{Full Correlation Plots for Sepsis}\\
\hspace*{20mm} \ref{App:Security}: \underline{Assessment of Disclosure Risk}\\
\hspace*{30mm} \ref{App:Security2}: \textit{El Emam \etal's Disclosure Risk Metrics}\\
\hspace*{30mm} \ref{App:Security3}: \textit{Alternative Disclosure Risk Metrics}

\section{Future Work and Broader Impact}\label{Sec:BImpact}
Clinical data is highly confidential and therefore difficult to share between researchers, which has hampered the development of robust machine learning algorithms in health care. In the Health Gym project we have shown that generative models could be used to create highly-realistic, longitudinal, datasets with mixed numerical and categorical variables. The risk of sensitive information disclosure associated with the release of these datasets was estimated to be very low. We aim to add further health-related datasets to the Health Gym in the near future and hope that this project will foster the development of reproducible and generalisable machine learning, and in particular reinforcement learning, applications in health care.

We are aware of only one other GAN-based model which attempted to create both numeric and non-numeric variables at the same time. Li \etal\cite{li2021generating} proposed a twin-encoder approach to separately embed numeric and non-numeric data, which required adding a matching loss for training the generative model. In contrast, the GAN model proposed in this study requires only one encoder and one decoder, since binary and categorical variables are mapped to continuous vectors through the use of soft-embeddings. Nonetheless, the recurrent components of the Health Gym GAN could potentially benefit from existing work on network simplification\cite{kuo2020input}. Diffusion models are an alternative approach for generating synthetic data, and they have recently achieved results comparable to state-of-the-art GANs\cite{dhariwal2021diffusion}. In future work, we plan to explore the use of diffusion models for improving the sample diversity and robustness of the training process of our generative models. Furthermore, the generated data could be made more realistic, and the generative process more explainable, by incorporating causal layers\cite{yang2021causalvae}.

The authors of this manuscript would like to emphasise that the generated synthetic datasets should not be regarded as replacements for the real datasets. The synthetic datasets are realistic, but it is currently unknown whether training a machine learning model with synthetic data leads to the same performance using the real data. We intend to conduct more investigations in this direction as part of future work. 

\newpage
\section{Instructions on Deriving the Real Datasets}\label{Sec:AppData}
This appendix aims to clarify the details for preparing \textbf{The Real Datasets} described in the \textbf{Methods} section. Interested readers should also refer to the official website of MIMIC-III\cite{johnson2016mimic} at \url{https://physionet.org/content/mimiciii/1.4/}; and likewise consult the official website of EuResist\cite{zazzi2012predicting} at \url{https://www.euresist.org/}.\\

\hspace*{-5mm}Our readers can click \hyperlink{Sec:Methods-GroundTruths}{here} to resume to the \textbf{The Real Datasets} in the \textbf{Methods} section.

\subsection{Acute Hypotension Dataset}\label{App:HypotensionDetail}
We mainly followed the instructions provided in Appendix D on page 15 of Gottesman \etal\cite{gottesman2020interpretable}. Although the authors published their codes in an open repository (see \url{https://github.com/dtak/interpretable_ope_public}), they did not release the SQL\cite{kawasaki2003guide} code used to query the raw data from the MIMIC-III database. We derived a larger real hypotension dataset which consisted of a cohort of $3,910$ patients; there were $1,733$  distinct ICU admissions in Gottesman \etal's original cohort of patients.

\subsubsection{Inclusion/Exclusion Criteria}
Gottesman \etal \text{ }provided the selection criteria for their patient cohort in Appendix D.1 on page 15 of their paper. In brief, the authors included adult patients ($\ge18$ years old) in the MICU MIMIC-III sub-dataset with those with at least 24 hours of data collected from the MetaVision (MV) clinical information system. Then, they aggregated 48 hours of clinical variables from those patients with seven or more mean arterial pressure (MAP) values of $65$ mmHg or less, which indicated probable acute hypotension.

\subsubsection{Data Preparation}
Almost all of the variables in Table \ref{Tab:Hypotension}, except the Glasgow coma scale (GCS) score, are originally recorded as continuous numeric variables in the MIMIC-III database. GCS is a point-based system to reflect a patient's state of consciousness\cite{teasdale1974assessment}. Fluid boluses, vasopressors, and fraction of inspired oxygen (FiO2) were treated as categorical variables, for reasons described below.

As documented in Appendix D.2 on page 15 of Gottesman \etal, most variables were queried \textit{as-is} from the MIMIC-III database. However, total vasopressor measurements were derived from the administered fluids following the code provided by Komorowski \etal\cite{komorowski2018artificial} (refer to \textit{Vasopressors from MetaVision} in the script \texttt{AIClinician\_Data\_extract\_MIMIC3\_140219.ipynb} in \url{https://gitlab.doc.ic.ac.uk/AIClinician/AIClinician}).

After following the previous steps, we converted some variables from numeric to categorical as according to Appendix D.4 on page 16 of Gottesman \etal. In this section, Gottesman \etal \text{ }explained that they categorised fluid boluses and vasopressors for training their RL algorithm in a discretised action space. Fluid boluses were binned into one of the four categories of \textit{none, low, medium,} and \textit{high} reflecting the numeric ranges of $0,[250,500),[500,1000), \text{and} \ge1000$, in mL units. Whereas for vasopressor, they binned the variable according to its first and third quantiles (\ie Q1 and Q3) and categorised vasopressor into the four classes of $0,(0,8.1),[8.1,21.58), \text{and}\ge 21.58$ in the unit of total mcg of drug given each hour per
kg body weight. However, our cohort of patients was slightly different to theirs and hence we binned vasopressors according to our re-calculated Q1 and Q3 values. Our vasopressor variable was thus categorised according to the ranges of $0,(0,8.4),[8.4,20.28), \text{and}\ge 20.28$, respectively.

Another variable which we made categorical was FiO2. Since FiO2 was the \textit{fraction} of inspired oxygen, we binned the variable into $10$ classes according to the ranges of $[0, 0.1), [0.1, 0.2), \ldots, [0.8, 0.9), \text{and } [0.9, 1.0]$.

\subsubsection{Missing Values}\label{Sec:A.1.3}
As commented in Appendix D.5 on page 16, Gottesman \etal \text{ }noted that the MIMIC-III database contains missing values. However, these missing values do not occur at random and can be highly informative. For instance, updated laboratory test are often ordered when a patient is transferred from one ICU to another. To deal with the missing values, we followed Gottesman \etal \text{ }and we constructed indicator variables (the binary (M) variables of Table \ref{Tab:Hypotension}) that denoted whether
or not a value was measured. Missing values were then replaced with the last available recorded data. 

\subsection{Sepsis Dataset}\label{App:Sepsis}
We mainly followed the steps in Komorowski \etal\cite{komorowski2018artificial}'s repository \url{https://gitlab.doc.ic.ac.uk/AIClinician/AIClinician} for preparing the real sepsis dataset. They queried the MIMIC-III database with their SQL file\\ \texttt{AIClinician\_Data\_extract\_MIMIC3\_140219.ipynb} and processed the queired data with their codes in\\ \texttt{AIClinician\_sepsis3\_def\_160219.m}.

\subsubsection{Inclusion/Exclusion Criteria}
Komorowski \etal\textcolor{white}{.} did not explicitly state the inclusion/exclusion criteria for the sepsis cohort in their paper. However, the processing code indicates that they only included adult patients. Then for all adults, they sought for any suspicious infections based on the history of antibiotic administrations; and they included all clinical variables from \textit{at least} $44$ hours before the suspected infection and \textit{up to} $28$ hours after the infection. The use of such time period for defining a sepsis event is supported by a 2016 sepsis definition paper\cite{singer2016third} which states that one should ``\textit{consider[s] a period as great as 48 hours before and up to 24 hours after onset of infection...}''; Komorowski \etal\textcolor{white}{.} likely relaxed this condition with a -4/+4 window.

\subsubsection{Data Preparation}\label{App:Sepsis_DP}
The data preparation procedure for the real sepsis dataset was simple, thanks to the code provided by Komorowski \etal \textcolor{white}{.}in their repository. The file containing the SQL queries should be executed first, followed by the processing code. Further details are provided below.

We executed the entire file containing the SQL queries. The sepsis dataset contained a much larger subset of data from MIMIC-III than the hypotension dataset. Whereas the hypotension dataset only included patients from the MICU branch, the sepsis dataset included patients from the NICU, SICU, CSRU, CCU, MICU, and TSICU branches. Furthermore, while the hypotension dataset only considered those patients who had their information recorded by the MetaVision (MV) system, the sepsis dataset also included those patients whose data were collected by the CareVue (CV) system.

After executing the file containing the SQL queries, we executed lines 1 to 791 of the processing code. This was the part which we used to prepare all variables listed in Tables \ref{Tab:Sepsis_Numeric} and \ref{Tab:Sepsis.NonNumeric}. We did not execute the remaining lines of that code. This was because those lines were for deriving the secondary auxiliary items of P/F ratio, shock index, SOFA score, and SIRS criteria. For instance, the P/F ratio is computed by dividing PaO2 by FiO2, whereas the shock index is based on HR and systolic BP. All variables required to compute these secondary items are included in the dataset (see Tables \ref{Tab:Sepsis_Numeric} and \ref{Tab:Sepsis.NonNumeric}).  

\subsubsection{Missing Values}
Whereas Gottesman \etal \text{ } chose to fill the missing values in the hypotension dataset by last observation carried forward, Komorowski \etal \text{ }chose to impute the missing values. In line 739 of the procesing code, they mainly used Matlab's\cite{MATLAB:2010} built-in \texttt{knnimpute} function and their own code in \texttt{fixgaps.m}. They do not appear to have used the code contained in \texttt{fastknnsearch.m}.

\subsection{HIV Dataset}\label{App:HIV}
We aimed to create a real HIV dataset from the EuResist\cite{zazzi2012predicting} database with a similar purpose to the one that was presented in the work of Parbhoo \etal\cite{parbhoo2017combining}. However, unlike the work of Gottesman \etal\textcolor{white}{ } (for hypotension) and Komorowski \etal\textcolor{white}{ } (for sepsis), Parbhoo \etal\textcolor{white}{ } did not make their codes publicly available and thus we mainly consulted with HIV medical experts to create our own version of the real HIV dataset.

\subsubsection{WHO Guidelines}\label{Sec:A.3.1}

As previously mentioned in the \textbf{Methods} section, we based our real HIV dataset on the work of Parbhoo \etal\textcolor{white}{ } but we also incorporated a published WHO guideline\cite{world2016consolidated} which called for the standardisation of antiretroviral therapy for HIV. Chapter 4 ``Clinical Guidelines: Antiretroviral Therapy'', starting from page 71 in the referenced work, also accessible at \url{https://www.who.int/hiv/pub/arv/chapter4.pdf}.

The referenced WHO guideline contains an extensive discussion on the \textit{strategic timing of antiretroviral treatment}. Prior to the WHO guideline, there was no standardised approach for initiating antiretroviral therapy in HIV. Antiretroviral therapy (ART) medications could be initialised when a person's CD4 count dropped either below 500 cells/$\mu$L (\ie moderate) or below 350 cells/$\mu$L (\ie critical). However, some studies have observed that early ART initiation increases the overall survival rate of people with HIV. Nonetheless, close monitoring in the first 3 months following ART initiation is required because people can develop hypersensitivity towards these medications, with potentially severe consequences.

More evidence on the benefits of the early adoption of ART medications can be found in a randomised controlled trial by the INSIGHT START Study Group\cite{insight2015initiation}. In this longitudinal study over 4 years, the intervention group received immediate-initiation of ART, whereas the control group deferred-initiation. The median time until initiation in the control group was 3 years. The immediate-initiation group had either equivalent or better survival rates across a variety of AIDS-related and non-AIDS-related events (see Table 2 on page 7 in the referenced paper).

\subsubsection{Inclusion/Exclusion Criteria}\label{Sec:A.3.2} 
Parbhoo \etal\textcolor{white}{ } extracted their cohort of patients from the entire EuResist database. They included people who were treated with the 312 most common medication combinations (including 20 medications). They also included patient demographics; however, their code is not publicly available and it is unclear which individual medications and demographic information were included in their dataset.

In order to create a similar dataset while incorporating information from the 2015 WHO guidelines\cite{world2016consolidated}, we restricted our cohort of patients to people initiated ART after 2015. Among the included population, we further selected those who were treated with the 50 most common medication combinations (including 21 medications). This choice is discussed further in the \textbf{Data Preparation} subsection below.   

\subsubsection{Data Preparation}\label{App:A.3.3}

\underline{Class Imbalance}\\
There were a total of 11,252 people in the EuResist database who initiated ART after 2015. Among these people, there were 20 who identified as neither female nor male (\ie less than $0.02\%$ of the total population). During our experiments, we found that our GAN model, consistently over more than 50 random initialisations, was unable to create datasets containing patient information related to this minority group. We therefore decided to exclude this group of people from our cohort; however, we recognise this as a limitation of our model that will need to be addressed by future algorithmic improvements.

Ethnicities in the EuResist database were also considerably imbalanced. They were originally categorised as Asian, African, Caucasian, Hispanic, Other, and Unknown. Since there were a very low number of 179 Hispanic people, we decided to combine Hispanic, Other, and Unknown into \textit{Other}.\\ 

\hspace{-5mm}\underline{Persistent Low Standardisation in HIV ART}\\
In Sections \ref{Sec:A.3.1} and \ref{Sec:A.3.2}, we mentioned how our inclusion criteria were based on the publication of the WHO guidelines in 2015. Indeed, we observed several changes in medication regimens over time. Before 2015 and especially in the early 2000s, it was not uncommon to see a regimen composed of more than 10 medications. However, after 2015, almost all regimens contained less than 6 medications; with most regimens consisting of only 3 or 4 medications.

Despite the continuous changes, there is still a lack of standardisation in the practice of HIV ART therapy. For instance, there were a total of 1,179 different medication combinations prescribed post-2015; and this number greatly exceeded the 14 suggested first line treatment combinations for adults in the WHO guideline (see the first three rows of Table 4.1 on page 97 of the referenced work\cite{world2016consolidated}). Upon closer analysis, we found that most medication combinations were only prescribed once, likely to suit just one patient's need. Moreover, we also found that the 10 most common medication combinations, which included a small subset of only 14 different medications, made up 34.1$\%$ of all prescription records. Thus, our aim was to extract a real HIV dataset which preserved the diversity in medication combinations while limiting an over-representation of single-use medication combinations.

Ultimately, we decided to include only the 50 most common medication combinations. For the 11,252 people in EuResist who initiated ART post-2015, there were 22,622 records (\ie rows in the database) in total. Including only the 50 most common medication combinations resulted in 17,660 records (78.1$\%$ of the total). Furthermore, the 50 most common medication combinations included 21 individual medications and made up 51.2$\%$ of all medication combinations prescribed post-2015.\\

\hspace{-5mm}\underline{Handling Long Gaps}\\
Within our selected cohort, the shortest records were ten months long while the longest lasted for more than three years. However, a large proportion of people had long gaps in their records. This was most apparent in the EuResist regimen information table -- someone who had just finished one therapy treatment would usually not be immediately prescribed with another treatment. We found that over a third of the people had one gap of over six months in their regimen treatment regimen; and that more than half of the remaining people had multiple (\ie two or three) treatment gaps in their records.

The use of data imputation techniques (such as splines) to fill these long gaps was unlikely to yield meaningful results. Instead, we decided to split the original records into several shorter sub-records describing continuous period of ART. Furthermore, to facilitate training of our GAN model, we truncated the sub-records' lengths to the closest multiple of ten (\eg 32 months were truncated at 30 months, and 51 months were truncated at 50 months). As shown in Figure \ref{Fig:GANs}, both our generator and discriminator included RNNs; and it has been shown previously\cite{daniluk2019frustratingly} that RNNs are good for processing mid-length data but not necessarily well-suited for very long data.\\

\hspace{-5mm}\underline{Missing Values}\\
Any other missing data values were filled in using last observation carried forwards,as discussed for the hypotension dataset in Section \ref{Sec:A.1.3}.\\

\hspace{-5mm}\underline{Decomposing Patient Regimens}\\
After managing the gaps and missing values in the data, we investigated different ways to represent the ART treatment in a meaningful manner. As mentioned in the \underline{Persistent Low Standardisation in HIV ART} subsection, our cohort were treated with 50 medication combinations including 21 individual medications. The medications belonged to five different classes$^\text{\ref{FN:DClasses}}$: \textit{nucleoside reverse transcriptase inhibitors} (NRTIs), \textit{nucleotide reverse transcriptase inhibitors} (NtRTIs), \textit{non-nucleotide reverse transcriptase inhibitors} (NNRTIs), \textit{integrase inhibitor} (INI), and \textit{protease inhibitors} (PIs). In addition to the medications belonging to the previous classes, it was common among people in the EuResist database to be prescribed \textit{pharmacokinetic enhancers} (pk-En) to boost the effectiveness of the other medications. 

ART medications tend to be associated with strong positive inter-class and strong negative intra-class correlations. For instance, if the INI medication of dolutegravir (DTG) was selected for a person's therapy, it was highly unlikely that another INI such as raltegravir (RAL) was required for the same therapy regimen. Contrarily, it is unnecessary to select an INI whenever an NNRTI was already present in the regimen (compare Table 4.1 on page 97 in the WHO guideline\cite{world2016consolidated}). 

We found that if we tried to represent each medication as a binary (or categorical) variable, the combination of strong positive and negative correlations would lead our GAN model into creating under-diversified synthetic datasets. For instance, our generator sub-network could have heavily favoured the coexistence of emtricitabine (FTC), an NRTI, with DTG, an INI; and thus became unable to learn any association between any other pair of NRTIs and INIs.  

To address this problem, we decided to represent medication regimens using two categorical variables. The first variable includes \textcolor{red}{base drug combination}, while the second variable includes \textcolor{brown}{other auxiliary and secondary medications}. For instance, a person could be treated with \textcolor{red}{FTC + TDF} \textcolor{brown}{+ DTG + pk-En}. Since all NRTI and NtRTI medications were included in the base medication combination variable, it was unnecessary to create separate categorical variables representing these classes.

%%%===
\begin{CJK}{UTF8}{min}
\newpage
\section{The Training Details of GANs}\label{App:GANsTraining}

This appendix aims to provide more technical details on the training of \textbf{The Health Gym GAN} introduced in the \textbf{Methods} section. The codes for this paper are also made publicly available; see the \textbf{Data Records} section for more details.\\

\hspace*{-5mm}\textbf{Summary of the GAN dimensionality:} Input dimension $128$, hidden dimension $128$, output dimension is dependent on the dataset. The embedded dimension of numeric data is $1$; and it is $2$ for binary data and $4$ for categorical data.\\

\hspace*{-5mm}Our readers can click \hyperlink{Sec:Methods-GAN}{here} to resume to the \textbf{The Health Gym GAN} in the \textbf{Methods} section.

\subsection{The Dimensionality in the GAN Setup}\label{App:GAN_B_1}
As mentioned previously, the GAN setting included a generator $G$ and a discriminator $D$. The generator was trained to map vectors of multivariate Gaussian inputs to the synthetic data $G:z\rightarrow x_\text{syn}$; whereas the discriminator took a batch of either real data $x_\text{real}$ or synthetic data $x_\text{syn}$, and scored the realisticness of the data $D:x\rightarrow\mathbb{R}$. The important dimensionalities in this setup were thus associated with the variables of $z$, $x_\text{syn}$ and $x_\text{real}$, within the networks of $G$ and $D$, and the outputs of the two networks.

\subsubsection{The Input and Hidden Dimensions of the Generator}
The dimensionality of the multivariate latent vectors $z$ was $\mathbb{R}^\mathscr{I}$ per synthetic patient per instance. That is, in order to generate a synthetic patient record over $T$ units of time, we would need a matrix of $z^{(1:T)}\in\mathbb{R}^{\mathscr{I}\times T}$; and we referred to $\mathscr{I}$ as the input dimension. Both $\mathscr{I}$ and $T$ were hyper-parameters and hence were selected prior to the training phase of the GAN model. For this work, we chose $\mathscr{I}=128$ for all datasets. However, the values for $T$ varied across different datasets. This was because whereas the real hypotension data had 48 time points for all patients; the real sepsis data had non-uniform length for each patient -- the shortest record was 2 time points while the longest was 20 time points. Likewise, the shortest real HIV data had 10 time points while the longest was of length 100. To generate synthetic datasets that closely represented their real counterparts, we selected $T=48$ for hypotension, $T=20$ for sepsis, and $T=60$ for HIV.

Each instance of $z^{(1:T)}$ was forwarded to the generator network $G$ one at a time (recall that the first module in the generator is biLSTM). All modules in network $G$ had a dimensionality of $\mathscr{H}$. The mapping learnt by network $G$ was hence $\mathbb{R}^\mathscr{I}\rightarrow\mathbb{R}^\mathscr{H}$. We referred to $\mathscr{H}$ as the hidden dimension; it was a hyper-parameter and we selected it to be $\mathscr{H}=128$ across all datasets.

\subsubsection{The Output Dimension of the Generator and the Contents of $X_{\text{syn}}$ and $X_{\text{real}}$}\label{App:B_1_2}

The output of the generator was also the input of the discriminator. Hence the output dimension of the generator $\mathscr{O}$ not only described the amount of features in the synthetic data $X_\text{syn}$, but it was also identical to the input dimension of the discriminator network. Since the synthetic data mimicked the real data $X_\text{real}$, the dimensionality of $\mathscr{O}$ varied across different datasets because each real dataset comprised a different number of clinical variables. As reported in the \textbf{Methods} section, there were $20$ variables for acute hypotension, $44$ for sepsis, and $13$ for HIV. However, the output dimensions were not $20$, $44$, and $13$.

The reason for this was the presence of various data types in our datasets. There were continuous numeric variables (\eg systolic BP), binary variables (\eg gender), and also categorical variables (\eg ethnicity). Whereas the numeric variables could be forwarded directly as float values to the discriminator, binary and categorical variables required additional transformations to present them in a machine-readable format to the discriminator.\\

\hspace*{-5mm}\textbf{Pre-processing Variables in $X_{\text{real}}$}\\
For hypotension, we first transformed each real numeric variable according to their optimal \textit{Box-Cox}~\cite{box1964analysis} transformation using the \texttt{stats.boxcox} function in the \texttt{Python}~\cite{CS-R9526} package of \texttt{Scipy}~\cite{2020SciPy-NMeth}. Afterwards, we re-centred and re-scaled the variables within the range of $[0, 1]$ with the \texttt{preprocessing.MinMaxScaler} scaler of the \texttt{Python}~\cite{CS-R9526} package of \texttt{Scikit-learn}~\cite{scikit-learn}. 

For sepsis, we first analysed if the real numeric variables had extremely long tails. As discussed in the \textbf{Data Records} section, we decided to subdivide numeric variables with extremely long tails into deciles and process them as categorical variables. The remaining real numeric variables were log-transformed if necessary, before re-centring and re-scaling within the range of $[0, 1]$ with \texttt{preprocessing.MinMaxScaler}.

Numeric variable processing was slightly easier for the HIV dataset. This was because the only numeric variables were VL and CD4. The former required an optimal Box-Cox transformation followed with re-centring and re-scaling to $[0, 1]$; whereas the latter required log-transformation followed by re-centring and re-scaling to $[0, 1]$. 

While processing the numeric variables of the real datasets $X_{\text{real}}$, it was important to record the transformation procedure for every individual variable. Only by doing so, we would be able to back-transform (\ie revert the re-centring and re-scaling procedure) the synthetic data $X_{\text{syn}}$ generated by network $G$. The dimensionality of real continuous numeric variables (without an extremely long tail) was preserved after pre-processing. Binary and categorical variables were converted into one-hot-encoded vectors with total length equal to two (for binary variables) and equal to the number of unique classes (for categorical variables).

\hspace*{-5mm}\textbf{Example: Pre-processing the Real Acute Hypotension Dataset}\\
For the real hypotension dataset, there were $20$ variables as listed in Table \ref{Tab:Hypotension}. The dataset comprised $9$ numeric variables, $4$ categorical variables, and $7$ binary variables. After transforming the $9$ numeric variables, there were still $9$ features. The transformation of the $7$ binary variables resulted in $14$ features; and there were $4 + 4 + 10 + 13 = 31$ unique classes for the categorical variables of fluid boluses, vasopressors, FiO2, and GCS. Hence, the output dimension of the generator network $G$ was also $\mathscr{O} = 9 + 14 + 31 = 54$. Note that this was the feature dimension for both $X_{\text{real}}$ and $X_{\text{syn}}$.\\

\hspace*{-5mm}\textbf{Case Study: An Interpretation of the Output of the Generator for Hypotension}\\
As mentioned earlier in this appendix, the generator network for hypotension served as a mapping for $G:z\rightarrow x_{\text{syn}}$ from dimension $\mathbb{R}^{128} \rightarrow \mathbb{R}^{54}$. Dimensions $1-9$ corresponded to the clinical variables in the strict order of MAP, diastolic BP, systolic BP, urine, ALT, AST, PaO2, lactate, and serum creatinine, respectively. Furthermore, since all of the real numeric variables were readjusted to lie within the range of $[0, 1]$ (see \textbf{Pre-processing Variables in $X_\text{real}$}), we passed the first $9$ dimensions of the generator output to the \texttt{sigmoid} activation function to match the extreme values allowed for the numeric variables.

Dimensions $10 - 40$ corresponded to the categorical variables. It was $10 - 13$ for fluid boluses, $14 - 17$ for vasopressors, $18 - 27$ for FiO2, and $28 - 40$ for GCS. To represent these dimensions of the synthetic data as categorical variables, we passed the corresponding dimensions to a \texttt{softmax} activation function. For instance, after passing dimension $10 - 13$ through \texttt{softmax}, dimension $10$ represented class I of fluid boluses; and likewise, $11$ for class II, $12$ for class III, and $13$ for class IV for the fluid boluses categories, respectively. A similar procedure was repeated for all other categorical variables, and likewise for all the binary variables (dimensions 41-54).

\subsubsection{On Soft Embedding}\label{App:B_1_3}
Before discussing the dimensionalities of the discriminator network $D$, we need to address the soft-embedding function in Figure \ref{Fig:GANs}(c). The input to network $D$ (\ie the grey box on the left of \ref{Fig:GANs}(c)) was either $x_{\text{syn}}$ or $x_{\text{real}}$. As depicted in the figure, all continuous numeric variables were treated as-is from the data, but the binary and categorical variables were subject to a soft embedding transformation. We created a trainable embedding matrix $W_{\text{emb}}$ for every binary or categorical variable. The weights were of size $\mathbb{R}^{(\text{あ},\text{か})}$ where あ denotse the size of the unique classes of the binary or categorical variables and か denotes the pre-specified size of the projection (2 for binary variables and 4 for categorical variables). The soft embedding helped us to represent the binary and categorical variable features as vectors of high dimensional floating values $x^{\text{T}}W_{\text{emb}}$ for the discriminator.\\

\hspace*{-5mm}\textbf{Case Study: Soft Embedding for the Hypotension Data}\\
Let us denote $X$ as the hypotension data prior to soft embedding and refer to $U$ as the data after soft embedding. The first $9$ dimensions of the hypotension data $X^{(1:9)}$ represented continuous numeric variables and were therefore not subject to soft embedding. Then for each binary or categorical variable, we defined a unique matrix $W_{emb}$. If the variable was binary, we would set か$= 2$; and categorical variables had か$= 4$ instead. These were hyper-parameters that were specified before training the GANs model; and these settings were applied to all of the synthetic datasets.

Of the 20 variables in the hypotension data $X\in\mathbb{R}^{54}$, the one-hot-encoded binary variables for urine (M) was stored in $X^{(41:42)}$. Thus, the dimension of the soft embedding matrix for urine (M) was ${2\times2}$ and the soft embedding procedure was hence $U^{\text{urine (M)}} = X^{(41:42)^\text{T}}W_\text{emb}^{\text{urine (M)}}$ resulted in a continuous feature vector of size $2$.

As an additional example, the information related to GCS was stored in $X^{(28:40)}$. The dimension of the soft embedding matrix for GCS was $13$ (the number of unique classes of GCS) by $4$ (the projection size for categorical variables). The soft embedding of GCS was hence $U^{\text{GCS}} = X^{(28:40)^\text{T}}W_\text{emb}^{\text{GCS}}$ resulting in a continuous feature vector of size $4$.

After the binary and categorical variables were transformed through soft embedding, they were concatenated with the numeric variables such that $U = [X^{(1:9)} \bigoplus U^{\text{Fluid Boluses}} \bigoplus U^{\text{Vasopressors}} \bigoplus \ldots \bigoplus U^{\text{Serum Creatinine (M)}}]$. Hence, the total dimension of the post-embedding vector $U$ was $9 + 4\times4 + 7\times2 = 39$; that is, $9$ dimensions for numeric with feature of size $4$ for the $4$ categorical variables and feature vectors of size $2$ for the $7$ binary variables in the hypotension data. In conclusion, the input dimension of the discriminator network $D$ was $39$ for the hypotension dataset.

\subsubsection{On the Dimensionality of the Discriminator}

The discriminator $D$ had an input dimension that varied across task (see Section \ref{App:B_1_3}), but it had a fixed hidden dimension of size $128$, and a fixed output dimension of $1$. The output dimension of the discriminator was not a hyper-parameter; instead, it was fixed as $1$ due to the design of the GAN architecture\cite{goodfellow2014generative}. This was because the output of the discriminator correspond to a realisticness score of the data $X$ where $D(X)\in\mathbb{R}$. 

\subsection{Remarks on Training the GAN Model}\label{App:GAN_B_2}
The scheme for training the WGAN-GP was mostly identical to the setting in the original paper~\cite{gulrajani2017improved}. The Health Gym GAN received the training data in batches of size $32$ and was updated using the \textit{Adam} optimiser~\cite{kingma2015adam} with learning rate $1\times 10^{-3}$. The model was regularised with $\lambda_\text{GP}=10$ for the gradient penalty loss (see Equation (\ref{Eq:WGANGP.D})) and $\lambda_\text{corr}=10$ for the alignment loss (see Equation (\ref{Eq:OurGLoss})). To stabilise the learning curve of the model, we adopted a \textit{curriculum learning}~\cite{bengio2009curriculum, press2017language} strategy where we first trained the GAN to create short sequences and then progressively generated longer synthetic time series. Furthermore, the generator network was updated once for every $5$ updates of the discriminator network (see Algorithm 1 on page 4 of the referenced work~\cite{gulrajani2017improved}).

The Health Gym GAN was capable of generating relatively good data after the first $100$ epochs of training. However, we found that the first $100$ epochs of training could be highly unstable, causing the quality of the synthetic data to vary greatly (mostly from seed to seed). The quality of the generated data stabilised when we prolonged the training to $500$ epochs. This was when we terminated the training of the GAN model and used the generator network to synthesise our datasets.\\

\hspace*{-5mm}\textbf{The Final Transformation After Synthetic Data Generation}\\
As mentioned in Appendix \ref{App:B_1_2}, all of the synthetic data $X_\text{syn}$ were passed through either a \texttt{sigmoid} activation function (for the numeric variables) or a \texttt{softmax} (for the binary or categorical variables). To ensure that each synthetic variable could be meaningfully interpreted, it was necessary to execute a back-transformation.

For continuous numeric variables this involved reverting the centring and scaling steps; and also applying the inverse of any prior power transformations. For the binary and categorical variables we only had to seek the largest probability in the \texttt{softmax} activated values. For instance, since the hypotension GCS features were stored in $X_\text{syn}^{(28:40)}$, we would have to find which value among those dimensions was the largest. Say that it was dimension $29$, then it meant that the second class of the GCS features was the most probable -- and hence it was a GCS of score $4$ (remember that GCS scores starts from 3, so that $3 + 1 = 4$).

\end{CJK}

%%%===
\newpage
\section{The Statistical Tests of Stage 2}\label{Sec:AppA}

This appendix contains the definitions of the statistical test used in stage 2 of our realisticness validation procedure and their outcomes.\\

\hspace*{-5mm}Our readers can click \hyperlink{Sec:TV-DRC}{here} to resume to \textbf{Stage Two: Statistical Tests} in the \textbf{Realisticness Validation Procedure} section.

\subsection{The Two Sample KS Test}\label{Sec:A.1}
The null hypothesis of the two sample KS test assumes that there are 
\begin{center}
\textit{no significant differences between the distributions}    
\end{center}
of the synthetic data and the real data for a specific variable $X$. The KS test accepts both numeric and categorical data types. Given a sample of data $B_S(\mathbf{x})$ from the synthetic variable and a sample of data $B_R(\mathbf{x})$ from the real variable, the test computes the statistic 
\begin{align}
K_{B_S,B_R}=\text{sup}_\mathbf{x} \left|C_{B_S}(\mathbf{x}) - C_{B_R}(\mathbf{x})\right|
\end{align}
based on the supremum of two \textit{empirical cumulative distribution functions} (ECDFs) $C$; with $C_{B_S}$ and $C_{B_R}$ denoting the ECDFs of $B_S(\mathbf{x})$ and $B_R(\mathbf{x})$ respectively. The ECDFs are computed as
\begin{align}
C_B = \frac{1}{n_B}\sum^{n_B}_{j=1}\mathbf{1}_{(-\infty,\mathbf{x}]}(x_j) 
\end{align}
for $x_j\in B(\mathbf{x})$ of size $n_B$. We denote $\mathbf{1}_{(-\infty,\mathbf{x}]}(x_j)$ as the indicator function: $1$ if $x_j\le \mathbf{x}$ and $0$ otherwise. The null hypothesis is rejected if the statistic $K_{B_S,B_R}$ exceeds a critical value with a statistical significance level $\alpha_{KS}$ following the Kolmogorov distribution\cite{kolmogorov1933sulla, smirnov1948table}; we use $\alpha_{KS}=0.05$. In our work, we implemented the two sample KS test with the \texttt{stats.ks\_2samp} function of the \texttt{Python}~\cite{CS-R9526} package of \texttt{Scipy}~\cite{2020SciPy-NMeth}. 

\subsection{The Two Independent Sample Student's t-Test}\label{Sec:A.2}

The null hypothesis of the two independent sample Student's t-test assumes that there are
\begin{center}
\textit{no significant differences between the means}
\end{center}
of the synthetic data and the real data for a specific variable $X$. The t-test only accepts numeric data types. Following the notations given in Appendix \ref{Sec:A.1}, the test statistic is 
\begin{align}
t_{B_S,B_R}=\frac{\overline{B_S(\mathbf{x})} - \overline{B_R(\mathbf{x})}}{S}\sqrt{\frac{n_{B_S}n_{B_R}}{n_{B_S} + n_{B_R}}}
\end{align}
where $\overline{B_S(\mathbf{x})}$ and $\overline{B_R(\mathbf{x})}$ refers to the respective sample means of $B_S(\mathbf{x})$ and $B_R(\mathbf{x})$; and the pooled standard deviation $S$ is
\begin{align}
S = \sqrt{\frac{S_{B_S}^2 + S_{B_R}^2}{2}}
\end{align}
where $S_{B_S}^2$ and $S_{B_R}^2$ denote the respective unbiased estimators for the variances of $B_S(\mathbf{x})$ and $B_R(\mathbf{x})$. The null hypothesis is rejected if the statistic $t_{B_S,B_R}$ exceeds a critical value of the Student's t-distribution~\cite{student1908probable} with a pre-determined threshold level $\alpha_t$; we use $\alpha_t = 0.05$.  In our work, we implemented the two independent sample t-test with the \texttt{stats.ttest\_ind} function of the \texttt{Python}~\cite{CS-R9526} package of \texttt{Scipy}~\cite{2020SciPy-NMeth}.

\subsection{The Snedecor's F-Test}\label{Sec:A.3}
The null hypothesis of the Snedecor's F-test assumes that there are
\begin{center}
\textit{no significant differences between the variances}
\end{center}
of the synthetic data and the real data for a specific variable X. The F-test accepts both numeric and categorical data types. Following the notations given in Appendices \ref{Sec:A.1} and \ref{Sec:A.2}, the test statistic for the numeric case is
\begin{align}
F_{B_S,B_R} = \frac{S_{B_S}^2}{S_{B_R}^2}
\end{align}
and it measures the deviation of the two population variances. The more this value deviates from 1, the stronger the evidence for unequal variances. The null hypothesis is rejected if the statistics $F_{B_S,B_R}$ is greater than a pre-determined critical value from the F distribution~\cite{johnson1995continuous} with significance level $\alpha_F$; we use $\alpha_F=0.05$. We implemented the F-test with the \texttt{stats.f.cdf} function of the \texttt{Python}~\cite{CS-R9526} package of \texttt{Scipy}~\cite{2020SciPy-NMeth}. For the categorical case, the F-statistic measured the between-group variance against the within-group variance instead; and it was implemented with the \texttt{stats.f\_oneway} function of \texttt{Scipy}. 

\subsection{The Three Sigma Rule Test}\label{Sec:A.4}
The three sigma rule test, also known as the 68-95-99.7 rule test, is an empirical rule to determine whether a value (or a set of values) lies within an interval estimate of a normal distribution. A variable $X$ sampled from a normal distribution with mean $\mu$ and standard deviation $S$ has the following approximated probabilities of lying within one, two, or three standard deviations from the mean: 
\begin{align}
\mathbb{P}(\mu - S\le X \le \mu + S) &\approx 68\%,\\
\mathbb{P}(\mu - 2S\le X \le \mu + 2S) &\approx 95\%\text{, and}\\
\mathbb{P}(\mu - 3S\le X \le \mu + 3S) &\approx 99.7\%.
\end{align}

In this study, we used the three sigma rule test to access whether a synthetic variable covered a similar range as the real counterpart, if it failed the KS test. This test was added because both the t-test and the F-test could only reveal properties of the synthetic data distribution (and explain potential shortcomings) but they could not be used to determine whether the synthetic data was still acceptable for downstream machine learning applications. We chose to use the two standard deviation confidence interval (CI with $\pm2S$); see also the following Appendix \ref{Sec:A.5}.

\newpage
\subsection{Iteratively Executing the Statistical Tests}\label{Sec:A.5}
\begin{algorithm}[h!]
\caption{The Psuedo-Code for Executing Stage 2}\label{Alg:Stage2}

\begin{algorithmic}[1]
\State \textcolor{black}{\colorbox{cyan!50}{$\eta_{KS}$}} $ = $ \textcolor{gray}{\colorbox{cyan!50}{$ \eta_{t}$}} $ = $ \textcolor{orange}{\colorbox{cyan!50}{$\eta_{F}$}} $ = $ \textcolor{purple}{\colorbox{cyan!50}{$\eta_{3S}$}} $ = 0$ \Comment{Initiate the counters as 0.}

\State \textcolor{white}{.}
\For{$\xi = 1,\textcolor{blue}{\Xi}$}\Comment{Run the statistical tests for $\Xi$ iterations.}
\State \textcolor{purple}{$\{\}_{3S}$} $ = \emptyset$ \Comment{Instantiate an empty set for the three sigma rule test.}

\State \textcolor{white}{.}
\State $B^*_S(\textbf{x})\sim X_\text{syn} \text{ and } B^*_R(\textbf{x})\sim X_\text{real}$ \Comment{Sample raw data with size \textcolor{red}{$n_B$}.}
\State $B_S(\textbf{x}) = \texttt{transform}\left(B^*_S(\textbf{x})\right) \text{ and } B_R(\textbf{x}) = \texttt{transform}\left(B^*_R(\textbf{x})\right)$ \Comment{Optional \texttt{minmax} transformation.}

\State \textcolor{white}{.}
\If{$X_\text{real} \text{ is numeric }$}
\State $P_{KS} = \texttt{KS test}(B_S(\textbf{x}), B_R(\textbf{x}))$ \Comment{Acquire \textit{p}-values for the statistical hypothesis test.}
\State $P_{t} = \texttt{t-test}(B_S(\textbf{x}), B_R(\textbf{x}))$ 
\State $P_{F} = \texttt{F-test}(B_S(\textbf{x}), B_R(\textbf{x}))$

\State \textcolor{white}{.}
\State $\epsilon_\text{lower}, \epsilon_\text{upper} = \mu(B_R(\textbf{x})) \pm 2S(B_R(\textbf{x}))$\Comment{Compute real value CI with 2 standard deviations.}
\State \textcolor{purple}{$\{\}_{3S}$} $=$ \textcolor{purple}{$\{\}_{3S}$} $\cup_{i = 1}^{n_B}B_S(\textbf{x}_i) \text{ if } B_S(\textbf{x}_i)\in[\epsilon_\text{lower}, \epsilon_\text{upper}] $\Comment{Append the list with synthetic values within the CI.}

\State \textcolor{white}{.}
\State \textcolor{black}{\colorbox{cyan!50}{$\eta_{KS}$}} $\text{ }+\hspace{-1mm}=1 \text{ if } P_{KS} >$ \colorbox{pink!50}{$\alpha_{KS}$} \Comment{Increase the counter if the null hypothesis is kept.}
\State \textcolor{gray}{\colorbox{cyan!50}{$\eta_{t}$}} $\text{ }+\hspace{-1mm}=1 \text{ if } P_{t} >$ \textcolor{gray}{\colorbox{pink!50}{$\alpha_{t}$}}\Comment{(Note the ``$>$'' sign.)}
\State \textcolor{orange}{\colorbox{cyan!50}{$\eta_{F}$}} $\text{ }+\hspace{-1mm}=1 \text{ if } P_{F} > $ \textcolor{orange}{\colorbox{pink!50}{$\alpha_{F}$}}

\State \textcolor{white}{.}
\State \textcolor{purple}{\colorbox{cyan!50}{$\eta_{3S}$}} $\text{ }+\hspace{-1mm}=1 \text{ if } |$ \textcolor{purple}{$\{\}_{3S}$} $| > \tau_{3S}$ \Comment{Increase the counter if the majority of synthetic data is in the CI.}
\EndIf

\State \textcolor{white}{.}
\If{$X_\text{real} \text{ is categorical }$}
\State $P_{KS} = \texttt{KS test}(B_S(\textbf{x}), B_R(\textbf{x}))$
\State $P_{F} = \texttt{F-test}(B_S(\textbf{x}), B_R(\textbf{x}))$

\State \textcolor{white}{.}
\State \textcolor{black}{\colorbox{cyan!50}{$\eta_{KS}$}} $\text{ }+\hspace{-1mm}=1 \text{ if } P_{KS} > $ \colorbox{pink!50}{$\alpha_{KS}$}
\State \textcolor{orange}{\colorbox{cyan!50}{$\eta_{F}$}} $\text{ }+\hspace{-1mm}=1 \text{ if } P_{F} > $ \textcolor{orange}{\colorbox{pink!50}{$\alpha_{F}$}}
\EndIf
\EndFor

\end{algorithmic}
\end{algorithm}

\hspace*{-5mm}We summarised the execution procedure for the statistical tests in the pseudo-code overleaf. The purpose of the algorithm was to compare a synthetic variable $X_\text{syn}$ against its real counterpart $X_\text{real}$. Since the synthetic datasets of this paper were primarily designed for machine learning algorithms including RL, we tested the generated data using repeated mini-batch sampling, a scenario that resembles the iterative training phase\cite{rumelhart1986learning} of a neural network. The iterative process also helped us estimate the Type II error of the statistical tests. A detailed description of the pseudo-code is provided below.

In order to mimic the iterative training phase of a nueral network, we ran the statistical tests for a pre-specified \textcolor{blue}{maximum number of iterations $\Xi = 100$} (Line 3) with a \textcolor{red}{sampling size of $n_B = 32$} (Line 6). Prior to the tests, we defined \textcolor{black}{\colorbox{cyan!50}{$\eta_{KS}$}}, \textcolor{gray}{\colorbox{cyan!50}{$\eta_{t}$}}, \textcolor{orange}{\colorbox{cyan!50}{$\eta_{F}$}}, \textcolor{purple}{\colorbox{cyan!50}{$\eta_{3S}$}} as the counters for non-rejected KS tests, t-tests, F-tests, and three sigma rule tests, respectively (Line 1). Numeric variables were rescaled as in the pre-processing step for machine learning model building (Line 7). We increased the counters by 1 whenever we \textit{kept} a null hypotheses (Lines 17 - 19), \ie when the \textit{p}-values were greater than the statistical significance level of \colorbox{pink!50}{$\alpha=0.05$}. When the iterations ended, we used the counters as indicators of the realisticness of the synthetic variables. If the final counter values were $70\%$ or higher than the amount of total iterations (\colorbox{cyan!50}{$\eta$} $ > 0.7$ \textcolor{blue}{$\Xi$}), we concluded that the synthetic variable had accurately captured the distributional features of the real variable. Of note, only the KS test and F-test were suitable for categorical variables. The three sigma rule test is not a statistical hypothesis test, and is discussed separately below.

The $70\%$ threshold was chosen arbitrarily. However, it was because that it was a relatively strict threshold for preventing Type I error, since we only regarded synthetic variables to be realistic when $70\%$ of the trials were associated with less than $5\%$ risk (the $5\%$ significance level) of concluding that a difference existed when there was no true difference. For completeness, we reported the \textit{rates} of passing each test for all variables in the tables of Appendix \ref{App:StatsOutcome}.

To conduct the three sigma rule test, we instantiated an empty set \textcolor{purple}{$\{\}_{3S}$} at the beginning of each iteration (Line 4). Then, for this study, we chose the $2S$ interval of the real variable as the range for reliable values (Line 14). After defining the lower and upper bounds, we appended \textcolor{purple}{$\{\}_{3S}$} with synthetic variables that lied within the defined range (Line 15). We set a threshold $\tau_{3S} = 0.7$ \textcolor{red}{$n_B$}; that is, we considered a mini-batch of synthetic variable passing the three sigma rule test if more than $70\%$ of the synthetic data of that batch fell within the two standard deviation range (Line 21). Finally, if more than $70\%$ of all random mini-batches passed the aforementioned threshold (\textcolor{purple}{\colorbox{cyan!50}{$\eta_{3S}$}}$ > 0.7$ \textcolor{blue}{$\Xi$}), we considered the synthetic variable to be sufficiently accurate even if it had failed the KS test.

%\newpage
\subsection{The Statistical Outcomes}\label{App:StatsOutcome}

The content below supplements the \textbf{Validation Outcomes} section of the \textbf{Technical Validation}.

\subsubsection{Acute Hypotension}\label{App:HypotensionStatistics}
\begin{table}[ht]
    \centering
    \begin{tabular}{|l||c|c|c|c|}
        \hline
        \textbf{Variable Name} & \textbf{\textcolor{white}{.}\hspace{3mm} KS-Test \textcolor{white}{.}\hspace{3mm}} & 
        \textbf{\textcolor{white}{.}\hspace{3mm} t-Test \textcolor{white}{.}\hspace{3mm}} & 
        \textbf{\textcolor{white}{.}\hspace{3mm} F-Test \textcolor{white}{.}\hspace{3mm}} & 
        \textbf{Three Sigma Rule Test}\\
        \hline
        \hline
        MAP & \cellcolor{cyan!10}$96/100$ & 
              \cellcolor{cyan!10}$95/100$ & 
              \cellcolor{cyan!10}$86/100$ & 
              \cellcolor{cyan!10}$100/100$\\
        \hline
        Diastolic BP & \cellcolor{cyan!10}$92/100$ & 
              \cellcolor{cyan!10}$94/100$ & 
              \cellcolor{cyan!10}$93/100$ & 
              \cellcolor{cyan!10}$100/100$\\
        \hline
        Systolic BP & \cellcolor{cyan!10}$98/100$ & 
              \cellcolor{cyan!10}$98/100$ & 
              \cellcolor{cyan!10}$91/100$ & 
              \cellcolor{cyan!10}$97/100$\\
        \hline
        Urine & \cellcolor{cyan!10}$95/100$ & 
                \cellcolor{cyan!10}$92/100$ & 
                \cellcolor{cyan!10}$96/100$ &
                \cellcolor{cyan!10}$100/100$\\
        \hline
        ALT & \cellcolor{magenta!10}$62/100$ & 
                \cellcolor{cyan!10}$94/100$ & 
                \cellcolor{cyan!10}$90/100$ &
                \cellcolor{cyan!10}$94/100$\\
        \hline
        AST & \cellcolor{magenta!10}$59/100$ & 
                \cellcolor{cyan!10}$95/100$ & 
                \cellcolor{cyan!10}$91/100$ &
                \cellcolor{cyan!10}$97/100$\\
        \hline
        PaO2 & \cellcolor{magenta!10}$40/100$ & 
                \cellcolor{cyan!10}$94/100$ & 
                \cellcolor{cyan!10}$88/100$ &
                \cellcolor{cyan!10}$97/100$\\
        \hline
        Lactic Acid & \cellcolor{cyan!10}$74/100$ & 
                \cellcolor{cyan!10}$96/100$ & 
                \cellcolor{cyan!10}$83/100$ &
                \cellcolor{cyan!10}$98/100$\\
        \hline
        Serum Creatinine & \cellcolor{cyan!10}$90/100$ & 
              \cellcolor{cyan!10}$94/100$ & 
              \cellcolor{cyan!10}$94/100$ & 
              \cellcolor{cyan!10}$100/100$\\
        \hline
        \hline
        Fluid Boluses & \cellcolor{cyan!10}$100/100$ & 
              \cellcolor{gray!10}- - & 
              \cellcolor{cyan!10}$84/100$ & 
              \cellcolor{gray!10}- -\\
        \hline
        Vasopressors & \cellcolor{cyan!10}$100/100$ & 
              \cellcolor{gray!10}- - & 
              \cellcolor{cyan!10}$95/100$ & 
              \cellcolor{gray!10}- -\\
        \hline
        FiO2 & \cellcolor{cyan!10}$100/100$ & 
              \cellcolor{gray!10}- - & 
              \cellcolor{cyan!10}$93/100$ & 
              \cellcolor{gray!10}- -\\
        \hline
        GCS & \cellcolor{cyan!10}$97/100$ & 
              \cellcolor{gray!10}- - & 
              \cellcolor{cyan!10}$95/100$ & 
              \cellcolor{gray!10}- -\\
        \hline
        \hline
        Urine (M) & \cellcolor{cyan!10}$97/100$ & 
              \cellcolor{gray!10}- - & 
              \cellcolor{cyan!10}$91/100$ & 
              \cellcolor{gray!10}- -\\
        \hline
        ALT/AST (M) & \cellcolor{cyan!10}$100/100$ & 
              \cellcolor{gray!10}- - & 
              \cellcolor{cyan!10}$75/100$ & 
              \cellcolor{gray!10}- -\\
        \hline
        FiO2 (M) & \cellcolor{cyan!10}$100/100$ & 
              \cellcolor{gray!10}- - & 
              \cellcolor{cyan!10}$85/100$ & 
              \cellcolor{gray!10}- -\\
        \hline
        GCS (M) & \cellcolor{cyan!10}$100/100$ & 
              \cellcolor{gray!10}- - & 
              \cellcolor{cyan!10}$95/100$ & 
              \cellcolor{gray!10}- -\\
        \hline
        PaO2 (M) & \cellcolor{cyan!10}$100/100$ & 
              \cellcolor{gray!10}- - & 
              \cellcolor{cyan!10}$86/100$ & 
              \cellcolor{gray!10}- -\\
        \hline
        Lactic Acid (M) & \cellcolor{cyan!10}$100/100$ & 
              \cellcolor{gray!10}- - & 
              \cellcolor{cyan!10}$88/100$ & 
              \cellcolor{gray!10}- -\\
        \hline
        Serum Creatinine (M) & \cellcolor{cyan!10}$100/100$ & 
              \cellcolor{gray!10}- - & 
              \cellcolor{cyan!10}$92/100$ & 
              \cellcolor{gray!10}- -\\
        \hline

    \end{tabular}
    
    \caption{\label{Tab:HypotensionStats}The Stage Two Statistics for Acute Hypotension}
\end{table}

\hspace*{-5mm}An overview of the statistic tests for the synthetic hypotension dataset is presented in Table \ref{Tab:HypotensionStats}. As described in Appendix \ref{Sec:A.5}, the statistical tests were repeated for $\Xi = 100$ iterations and were considered ``passed'' if the null hypothesis was not rejected $70\%$ or more of the times (\colorbox{cyan!50}{$\eta$} $ > 0.7$ \textcolor{blue}{$\Xi$}). The denominators of the scores in the table refer to the maximum number of iterations, and the numerators correspond to the number of times that the null hypothesis was not rejected. For instance, MAP with a score of $96/100$ meant that $96$ out of $100$ times we kept the null hypothesis of the KS-test at a significance level of $0.05$. This passes the $70\%$ mark and is highlighted in blue. Failed tests are highlighted in red. 

All variables in the synthetic hypotension datasets are realistic. Although 3 variables (ALT, AST, and PaO2) fail the KS test, their means and variances are captured appropriately; and all variables are within plausible ranges of the real variables.

\subsubsection{Sepsis}\label{App:SepsisStatistics}
Of all the $44$ variables, only Max Vaso behaves undesirably. However, as we previously mentioned in the \textbf{Technical Validation} section, this was because Max Vaso was highly skewed. As described in the \textbf{Data Records} section, Max Vaso was not transformed into a categorical variable because its most related variables (\ie Input Total, Input 4H, Output Total, and Output 4H) were numeric. 

Nonetheless, the synthetic Max Vaso variable performed well regarding the three sigma rule test, indicating that the synthetic variable was within a plausible range of the real variable values. To conclude, the synthetic sepsis dataset is highly realistic.

\newpage
\begin{table}[ht]
    \centering
    \begin{tabular}{|l||c|c|c|c|}
        \hline
        \textbf{Variable Name} & \textbf{\textcolor{white}{.}\hspace{3mm} KS Test \textcolor{white}{.}\hspace{3mm}} & 
        \textbf{\textcolor{white}{.}\hspace{3mm} t-Test \textcolor{white}{.}\hspace{3mm}} & 
        \textbf{\textcolor{white}{.}\hspace{3mm} F-Test \textcolor{white}{.}\hspace{3mm}} & 
        \textbf{Three Sigma Rule Test}\\
        \hline
        \hline
        
        %%%===%%%
        Age & \cellcolor{cyan!10}$96/100$ & 
              \cellcolor{cyan!10}$96/100$ & 
              \cellcolor{cyan!10}$98/100$ & 
              \cellcolor{cyan!10}$100/100$\\
        \hline
        
        %%%===%%%
        HR & \cellcolor{cyan!10}$98/100$ & 
              \cellcolor{cyan!10}$95/100$ & 
              \cellcolor{cyan!10}$95/100$ & 
              \cellcolor{cyan!10}$100/100$\\
        \hline
        
        %%%===%%%
        Systolic BP & \cellcolor{cyan!10}$96/100$ & 
              \cellcolor{cyan!10}$93/100$ & 
              \cellcolor{cyan!10}$96/100$ & 
              \cellcolor{cyan!10}$100/100$\\
        \hline
        
        %%%===%%%
        Mean BP & \cellcolor{cyan!10}$94/100$ & 
              \cellcolor{cyan!10}$95/100$ & 
              \cellcolor{cyan!10}$92/100$ & 
              \cellcolor{cyan!10}$100/100$\\
        \hline
        
        %%%===%%%
        Diastolic BP & \cellcolor{cyan!10}$95/100$ & 
              \cellcolor{cyan!10}$94/100$ & 
              \cellcolor{cyan!10}$92/100$ & 
              \cellcolor{cyan!10}$100/100$\\
        \hline
        
        %%%===%%%
        RR & \cellcolor{cyan!10}$97/100$ & 
              \cellcolor{cyan!10}$95/100$ & 
              \cellcolor{cyan!10}$95/100$ & 
              \cellcolor{cyan!10}$98/100$\\
        \hline
        
        %%%===%%%
        K & \cellcolor{cyan!10}$94/100$ & 
              \cellcolor{cyan!10}$97/100$ & 
              \cellcolor{cyan!10}$91/100$ & 
              \cellcolor{cyan!10}$100/100$\\
        \hline
        
        %%%===%%%
        Na & \cellcolor{cyan!10}$97/100$ & 
              \cellcolor{cyan!10}$95/100$ & 
              \cellcolor{cyan!10}$91/100$ & 
              \cellcolor{cyan!10}$100/100$\\
        \hline
        
        %%%===%%%
         Cl$^-$ & \cellcolor{cyan!10}$96/100$ & 
              \cellcolor{cyan!10}$95/100$ & 
              \cellcolor{cyan!10}$97/100$ & 
              \cellcolor{cyan!10}$99/100$\\
        \hline
        
        %%%===%%%
         Ca & \cellcolor{cyan!10}$95/100$ & 
              \cellcolor{cyan!10}$97/100$ & 
              \cellcolor{cyan!10}$87/100$ & 
              \cellcolor{cyan!10}$100/100$\\
        \hline
        
        %%%===%%%
        Ionised Ca& \cellcolor{cyan!10}$93/100$ & 
              \cellcolor{cyan!10}$92/100$ & 
              \cellcolor{cyan!10}$85/100$ & 
              \cellcolor{cyan!10}$99/100$\\
        \hline
        
        %%%===%%%
        CO2 & \cellcolor{cyan!10}$92/100$ & 
              \cellcolor{cyan!10}$93/100$ & 
              \cellcolor{cyan!10}$93/100$ & 
              \cellcolor{cyan!10}$99/100$\\
        \hline
        
        %%%===%%%
        Albumin & \cellcolor{cyan!10}$96/100$ & 
              \cellcolor{cyan!10}$95/100$ & 
              \cellcolor{cyan!10}$96/100$ & 
              \cellcolor{cyan!10}$99/100$\\
        \hline
        
        %%%===%%%
        Hb & \cellcolor{cyan!10}$96/100$ & 
              \cellcolor{cyan!10}$95/100$ & 
              \cellcolor{cyan!10}$96/100$ & 
              \cellcolor{cyan!10}$100/100$\\
        \hline
        
        %%%===%%%
        pH & \cellcolor{cyan!10}$98/100$ & 
              \cellcolor{cyan!10}$98/100$ & 
              \cellcolor{cyan!10}$88/100$ & 
              \cellcolor{cyan!10}$100/100$\\
        \hline
        
        %%%===%%%
        BE & \cellcolor{cyan!10}$90/100$ & 
              \cellcolor{cyan!10}$94/100$ & 
              \cellcolor{cyan!10}$90/100$ & 
              \cellcolor{cyan!10}$98/100$\\
        \hline
        
        %%%===%%%
        HCO3 & \cellcolor{cyan!10}$95/100$ & 
              \cellcolor{cyan!10}$95/100$ & 
              \cellcolor{cyan!10}$94/100$ & 
              \cellcolor{cyan!10}$100/100$\\
        \hline
        
        %%%===%%%
        FiO2 & \cellcolor{cyan!10}$91/100$ & 
              \cellcolor{cyan!10}$99/100$ & 
              \cellcolor{cyan!10}$94/100$ & 
              \cellcolor{cyan!10}$99/100$\\
        \hline
        
        %%%===%%%
        Glucose & \cellcolor{cyan!10}$96/100$ & 
              \cellcolor{cyan!10}$96/100$ & 
              \cellcolor{cyan!10}$89/100$ & 
              \cellcolor{cyan!10}$99/100$\\
        \hline
        
        %%%===%%%
        BUN & \cellcolor{cyan!10}$94/100$ & 
              \cellcolor{cyan!10}$94/100$ & 
              \cellcolor{cyan!10}$93/100$ & 
              \cellcolor{cyan!10}$100/100$\\
        \hline
        
        %%%===%%%
        Creatinine & \cellcolor{cyan!10}$94/100$ & 
              \cellcolor{cyan!10}$94/100$ & 
              \cellcolor{cyan!10}$90/100$ & 
              \cellcolor{cyan!10}$100/100$\\
        \hline
        
        %%%===%%%
        Mg & \cellcolor{cyan!10}$90/100$ & 
              \cellcolor{cyan!10}$93/100$ & 
              \cellcolor{cyan!10}$90/100$ & 
              \cellcolor{cyan!10}$98/100$\\
        \hline
        
        %%%===%%%
        SGOT & \cellcolor{cyan!10}$92/100$ & 
              \cellcolor{cyan!10}$96/100$ & 
              \cellcolor{cyan!10}$84/100$ & 
              \cellcolor{cyan!10}$98/100$\\
        \hline
        
        %%%===%%%
        SGPT & \cellcolor{cyan!10}$90/100$ & 
              \cellcolor{cyan!10}$89/100$ & 
              \cellcolor{cyan!10}$93/100$ & 
              \cellcolor{cyan!10}$100/100$\\
        \hline
        
        %%%===%%%
        Total Bili & \cellcolor{cyan!10}$91/100$ & 
              \cellcolor{cyan!10}$91/100$ & 
              \cellcolor{cyan!10}$87/100$ & 
              \cellcolor{cyan!10}$100/100$\\
        \hline
        
        %%%===%%%
        WBC & \cellcolor{cyan!10}$97/100$ & 
              \cellcolor{cyan!10}$99/100$ & 
              \cellcolor{cyan!10}$93/100$ & 
              \cellcolor{cyan!10}$100/100$\\
        \hline
        
        %%%===%%%
        Platelets & \cellcolor{cyan!10}$94/100$ & 
              \cellcolor{cyan!10}$96/100$ & 
              \cellcolor{cyan!10}$95/100$ & 
              \cellcolor{cyan!10}$99/100$\\
        \hline
        
        %%%===%%%
        PaO2 & \cellcolor{cyan!10}$97/100$ & 
              \cellcolor{cyan!10}$98/100$ & 
              \cellcolor{cyan!10}$95/100$ & 
              \cellcolor{cyan!10}$100/100$\\
        \hline
        
        %%%===%%%
        PaCO2 & \cellcolor{cyan!10}$95/100$ & 
              \cellcolor{cyan!10}$95/100$ & 
              \cellcolor{cyan!10}$95/100$ & 
              \cellcolor{cyan!10}$99/100$\\
        \hline
        
        %%%===%%%
        Lactate & \cellcolor{cyan!10}$97/100$ & 
              \cellcolor{cyan!10}$97/100$ & 
              \cellcolor{cyan!10}$84/100$ & 
              \cellcolor{cyan!10}$100/100$\\
        \hline
        
        %%%===%%%
        Input Total & \cellcolor{cyan!10}$95/100$ & 
              \cellcolor{cyan!10}$95/100$ & 
              \cellcolor{cyan!10}$84/100$ & 
              \cellcolor{cyan!10}$99/100$\\
        \hline
        
        %%%===%%%
        Input 4H & \cellcolor{cyan!10}$83/100$ & 
              \cellcolor{cyan!10}$94/100$ & 
              \cellcolor{cyan!10}$94/100$ & 
              \cellcolor{cyan!10}$100/100$\\
        \hline
        
        %%%===%%%
        Max Vaso & \cellcolor{magenta!10}$0/100$ & 
              \cellcolor{cyan!10}$96/100$ & 
              \cellcolor{magenta!10}$30/100$ & 
              \cellcolor{cyan!10}$97/100$\\
        \hline
        
        %%%===%%%
        Output Total & \cellcolor{cyan!10}$92/100$ & 
              \cellcolor{cyan!10}$95/100$ & 
              \cellcolor{cyan!10}$89/100$ & 
              \cellcolor{cyan!10}$99/100$\\
        \hline
        
        %%%===%%%
        Output 4H & \cellcolor{cyan!10}$90/100$ & 
              \cellcolor{cyan!10}$96/100$ & 
              \cellcolor{cyan!10}$90/100$ & 
              \cellcolor{cyan!10}$100/100$\\
        \hline

    \end{tabular}
    
    \caption{\label{Tab:SepsisStatsp1}A Subset of Stage Two Statistics for Sepsis}
\end{table}

\begin{table}[h!]
    \centering
    \begin{tabular}{|l||c|c|c|c|}
        \hline
        \textbf{Variable Name} & \textbf{\textcolor{white}{.}\hspace{3mm} KS-Test \textcolor{white}{.}\hspace{3mm}} & 
        \textbf{\textcolor{white}{.}\hspace{3mm} t-Test \textcolor{white}{.}\hspace{3mm}} & 
        \textbf{\textcolor{white}{.}\hspace{3mm} F-Test \textcolor{white}{.}\hspace{3mm}} & 
        \textbf{Three Sigma Rule Test}\\
        \hline
        \hline
        
        %%%===%%%
        Gender & \cellcolor{cyan!10}$88/100$ & 
              \cellcolor{gray!10}- - & 
              \cellcolor{cyan!10}$64/100$ & 
              \cellcolor{gray!10}- -\\
        \hline
        
        %%%===%%%
        Readmission & \cellcolor{cyan!10}$97/100$ & 
              \cellcolor{gray!10}- - & 
              \cellcolor{cyan!10}$88/100$ & 
              \cellcolor{gray!10}- -\\
        \hline
        
        %%%===%%%
        Mech & \cellcolor{cyan!10}$100/100$ & 
              \cellcolor{gray!10}- - & 
              \cellcolor{cyan!10}$96/100$ & 
              \cellcolor{gray!10}- -\\
        \hline
        \hline
        
        %%%===%%%
        GCS & \cellcolor{cyan!10}$93/100$ & 
              \cellcolor{gray!10}- - & 
              \cellcolor{cyan!10}$86/100$ & 
              \cellcolor{gray!10}- -\\
        \hline
        
        %%%===%%%
        SpO2 & \cellcolor{cyan!10}$97/100$ & 
              \cellcolor{gray!10}- - & 
              \cellcolor{cyan!10}$96/100$ & 
              \cellcolor{gray!10}- -\\
        \hline
        
        %%%===%%%
        Temp & \cellcolor{cyan!10}$95/100$ & 
              \cellcolor{gray!10}- - & 
              \cellcolor{cyan!10}$91/100$ & 
              \cellcolor{gray!10}- -\\
        \hline
        
        %%%===%%%
        PTT & \cellcolor{cyan!10}$94/100$ & 
              \cellcolor{gray!10}- - & 
              \cellcolor{cyan!10}$87/100$ & 
              \cellcolor{gray!10}- -\\
        \hline
        
        %%%===%%%
        PT & \cellcolor{cyan!10}$97/100$ & 
              \cellcolor{gray!10}- - & 
              \cellcolor{cyan!10}$96/100$ & 
              \cellcolor{gray!10}- -\\
        \hline
        
        %%%===%%%
        INR & \cellcolor{cyan!10}$99/100$ & 
              \cellcolor{gray!10}- - & 
              \cellcolor{cyan!10}$97/100$ & 
              \cellcolor{gray!10}- -\\
        \hline

    \end{tabular}
    
    \caption{\label{Tab:SepsisStatsp2}The Remaining Stage Two Statistics for Sepsis}
\end{table}

\newpage
\subsubsection{HIV}\label{App:HIVStatistics}
\begin{table}[ht]
    \centering
    \begin{tabular}{|l||c|c|c|c|}
        \hline
        \textbf{Variable Name} & \textbf{\textcolor{white}{.}\hspace{3mm} KS-Test \textcolor{white}{.}\hspace{3mm}} & 
        \textbf{\textcolor{white}{.}\hspace{3mm} t-Test \textcolor{white}{.}\hspace{3mm}} & 
        \textbf{\textcolor{white}{.}\hspace{3mm} F-Test \textcolor{white}{.}\hspace{3mm}} & 
        \textbf{Three Sigma Rule Test}\\
        \hline
        \hline
        VL & 
        \cellcolor{magenta!10}$48/100$ & 
        \cellcolor{cyan!10}$90/100$ & 
        \cellcolor{magenta!10}$15/100$ & 
        \cellcolor{cyan!10}$100/100$\\
        \hline
        CD4 & 
        \cellcolor{cyan!10}$83/100$ & 
        \cellcolor{cyan!10}$97/100$ & 
        \cellcolor{cyan!10}$91/100$ & 
        \cellcolor{cyan!10}$91/100$\\
        \hline
        Rel CD4 & \cellcolor{cyan!10}$85/100$ & 
        \cellcolor{cyan!10}$92/100$ & 
        \cellcolor{cyan!10}$100/100$ & 
        \cellcolor{cyan!10}$99/100$\\
        \hline
        \hline
        Gender & 
        \cellcolor{cyan!10}$97/100$ & 
        \cellcolor{gray!10}- - & 
        \cellcolor{magenta!10}$49/100$ &
        \cellcolor{gray!10}- -\\
        \hline
        Ethnic & 
        \cellcolor{cyan!10}$81/100$ & 
        \cellcolor{gray!10}- - & 
        \cellcolor{magenta!10}$38/100$ &
        \cellcolor{gray!10}- -\\
        \hline
        \hline
        Base Drug Combo & \cellcolor{cyan!10}$73/100$ & 
        \cellcolor{gray!10}- - & 
        \cellcolor{cyan!10}$71/100$ &
        \cellcolor{gray!10}- -\\
        \hline
        Comp. INI & \cellcolor{cyan!10}$96/100$ & 
        \cellcolor{gray!10}- - & 
        \cellcolor{cyan!10}$73/100$ &
        \cellcolor{gray!10}- -\\
        \hline
        Comp. NNRTI & \cellcolor{cyan!10}$92/100$ & 
        \cellcolor{gray!10}- - & 
        \cellcolor{cyan!10}$95/100$ &
        \cellcolor{gray!10}- -\\
        \hline
        Extra PI & \cellcolor{cyan!10}$100/100$ & 
        \cellcolor{gray!10}- - & 
        \cellcolor{cyan!10}$73/100$ & 
        \cellcolor{gray!10}- -\\
        \hline
        Extra pk-En & \cellcolor{cyan!10}$100/100$ & 
        \cellcolor{gray!10}- - & 
        \cellcolor{cyan!10}$95/100$ & 
        \cellcolor{gray!10}- -\\
        \hline
        \hline
        VL (M) & 
        \cellcolor{cyan!10}$100/100$ & 
        \cellcolor{gray!10}- - & 
        \cellcolor{cyan!10}$72/100$ & 
        \cellcolor{gray!10}- -\\
        \hline
        CD4 (M) & \cellcolor{cyan!10}$99/100$ & 
        \cellcolor{gray!10}- - & 
        \cellcolor{cyan!10}$94/100$ & 
        \cellcolor{gray!10}- -\\
        \hline
        Drug (M) & \cellcolor{cyan!10}$100/100$ & 
        \cellcolor{gray!10}- - & 
        \cellcolor{cyan!10}$88/100$ & 
        \cellcolor{gray!10}- -\\
        \hline
        
    \end{tabular}
    
    \caption{\label{Tab:HIVStats}The Stage Two Statistics for HIV}
\end{table}

\hspace*{-5mm}Of all the $13$ variables, only VL failed the KS-test. Similar to Max Vaso in sepsis, VL was highly skewed and also failed the F-test. This shows the difficulty in appropriately capturing the variability of highly skewed distributions. Nonetheless, we found that VL was still realistic as it passed the three sigma rule test with very good scores.

Both gender and ethnicity required extra attention. Even though both variables passed the KS-test, they did not perform well on the ANOVA F-test. In fact, several binary and categorical variables in the synthetic HIV dataset only achieved acceptable (70$\%$+), instead of ideal (preferrably 85$\%$+), scores for their F-tests. While we were unable to explain the exact reason for this, we speculated that this was due to the combination of variables in the dataset. Unlike the hypotension and sepsis datasets, the HIV dataset contained very few numeric variables. Therefore, it is possible that many categorical variables were more difficult to synthesise.

Overall, we believe that the synthetic HIV dataset is still realistic. All but one variable passed the KS-test, and the numeric variables performed very well on the three sigma rule test.

%%%===
\newpage
\section{The Correlations of Stage 3}\label{Sec:AppB}

This appendix contains details on the correlation assessments that we used in stage 3 of the \textbf{Realisticness Validation Procedure} in the \textbf{Technical Validation} section. The definitions of each test and potential alternatives are discussed below.\\

\hspace*{-5mm}Our readers can click \hyperlink{Sec:TV-CV}{here} to resume to \textbf{Stage Three: Correlations} in the \textbf{Realisticness Validation Procedure} section.

\subsection{Kendall's Rank Correlation}\label{App:B1}
The non-parametric Kendall rank correlation, otherwise known as Kendall's $\tau$ correlation, is a statistic that measures the strength of association between two variables. In this study, we used the $\tau$-b variant and we refer to the original paper\cite{kendall1945treatment} for its precise definition. 

The magnitude of the correlation between a pair of variables, $X^{(i)}$ and $X^{(j)}$ for $i\neq j$, was measured by the score $\tau^{(i, j)} \in[-1, 1]$. A score of 1 or -1 indicated perfect positive or negative alignment between $X^{(i)}$ and $X^{(j)}$ respectively; whereas 0 indicated no correlation. We calculated the Kendall's $\tau$ with the \texttt{.corr()} function of the \texttt{Python}~\cite{CS-R9526} package of \texttt{Pandas}~\cite{mckinney2010data}.   

\subsection{Correlation Between Variables}\label{App:B2}

Given the synthetic dataset $D_\text{syn}$ and the real dataset $D_\text{real}$, the Kendall rank correlation scores were used to determine whether correlations between any pair of variables (based on data from all patients and timepoints) were captured in the generated data.

\subsection{Average Correlation in Trends and in Cycles}\label{App:B3}

The correlations in this section are used to investigate the behaviour of the generated data over time. Our datasets comprise many patients, each of whom is associated with a specific trajectory over time. Thus, correlations over time were computed individually for every patient and pair of variables. Afterwards, these correlations were averaged across patients.

More in detail, for a specific patient $p_k$ we selected a pair of variables $_{p_k}X^{(i)}$ and $_{p_k}X^{(j)}$. We assumed that they could be linearly decomposed into trends an cycles:
\begin{align}
    _{p_k}X^{(i)} &= \text{Trend}(_{p_k}X^{(i)}) + \text{Cycle}(_{p_k}X^{(i)}) \text{ and}\\
    _{p_k}X^{(j)} &= \text{Trend}(_{p_k}X^{(j)}) + \text{Cycle}(_{p_k}X^{(j)}).
\end{align}
We implemented Cycle() with the \texttt{signal.detrend()} function of the \texttt{Python}~\cite{CS-R9526} package of \texttt{Scipy}~\cite{2020SciPy-NMeth}; whereas Trend() is simply the remainder of $X - \text{Cycle}(X)$. By using the \texttt{signal.detrend()} function, we assumed that each time series has a linear trend. Alternatively, non-linear trends over time could be determined using moving average algorithms. 

We repeat this process for every patient $p_k$, $k = 1,\ldots,n$, and average the correlation scores for trends and cycles (the number of patients in the synthetic dataset might be different from the one in the real dataset, since the GAN can be used to generate any amount of data). Pseudo-code for computing the average correlations in trends and cycles is provided in Algorithm \ref{Alg:Corr2n3} for additional clarity.

\subsection{Remarks on Kendall's Rank Correlation and Alternative Validation Metrics}\label{App:B4}
In this study, we conducted the technical validation using Kendall's rank correlation instead of the more common \textit{Pearson's r correlation}~\cite{mukaka2012guide}. This was because of two reasons. First, Pearson's correlation requires all test variables to be normally distributed~\cite{kowalski1972effects}. As shown in Figure \ref{Fig:Hypotension.Validation001}, there were several variables that had long tail distributions and could not be power-transformed appropriately. Second and unlike Pearson's correlation, Kendall's rank correlation could be used for both binary and categorical variables. Furthermore, Kendall's rank correlation was even applicable in scenarios where we needed to test the relations between a multi-class variable and a binary-class variable.

Besides correlations, we also considered other metrics for validating the quality of the generated synthetic data. For example, \textit{auto-correlations} and \textit{cross-correlations}~\cite{bracewell1986fourier} would be useful to better understand the properties of the generated time series. However, they do not appear suitable to assess \textit{the similarity between patterns in two time series from two datasets}. Similarly, the \textit{Wilcoxon-Mann-Whitney}~\cite{mann1947test} test would be helpful to determine if two variables are independent; however, our focus was on \textit{the similarity of relations between variables from two datasets}.

Our validation setup also differs from the work of Goncalves \textit{et al.}\cite{goncalves2020generation}. In their study, Goncalves \textit{et al.} employed log-cluster\cite{woo2009global} and the Kullback–Leibler (KL) divergence\cite{kullback1951information} to compare the relative similarity of synthetic datasets generated by different models against their real ``ground truth'' dataset. In our study, we compared synthetic datasets generated using one GAN model to the real datasets.

\newpage
\begin{algorithm}[ht]
\caption{The Psuedo-Code for the Average Correlations}\label{Alg:Corr2n3}

\begin{algorithmic}[1]
\State $\{\}_\text{trend:syn}, \{\}_\text{cycle:syn}, \{\}_\text{trend:real}, \{\}_\text{cycle:real} = \emptyset$ \Comment{Create storage to record correlation scores.}

\State \textcolor{white}{.}
\For{$k = 1, n$}\Comment{Loop over each patient.}

    \State \textcolor{white}{.}
    \For{$i = 1, N$}\Comment{Select a variable $i$.}
        \State $\{\}_\text{trend:syn}^\text{k, i}, \{\}_\text{cycle:syn}^\text{k, i}, \{\}_\text{trend:real}^\text{k, i}, \{\}_\text{cycle:real}^\text{k, i} = \emptyset$ \Comment{Create a separate storage for patient-variables pair.}
        
        \State \textcolor{white}{.}
        \For{$j = 1, (i-1)$}\Comment{And select another variable $j$.}
            
            \State \textcolor{white}{.}
            \State $_{p_k}X^{(i)}_\text{syn} = \text{Trend}(_{p_k}X^{(i)}_\text{syn}) + \text{Cycle}(_{p_k}X^{(i)}_\text{syn})$ \Comment{Decompose the selected time series}
            \State $_{p_k}X^{(j)}_\text{syn} = \text{Trend}(_{p_k}X^{(j)}_\text{syn}) + \text{Cycle}(_{p_k}X^{(j)}_\text{syn})$
            \State $_{p_k}X^{(i)}_\text{real} = \text{Trend}(_{p_k}X^{(i)}_\text{real}) + \text{Cycle}(_{p_k}X^{(i)}_\text{real})$
            \State $_{p_k}X^{(j)}_\text{real} = \text{Trend}(_{p_k}X^{(j)}_\text{real}) + \text{Cycle}(_{p_k}X^{(j)}_\text{real})$
            
            \State \textcolor{white}{.}
            \State $\tau^{(i,j)}_\text{trend:syn} = \tau\left(\text{Trend}(X^{(i)}_\text{syn}), \text{Trend}(X^{(j)}_\text{syn})\right)$ \Comment{Compute the respective correlation scores in trend.}
            \State $\tau^{(i,j)}_\text{trend:real} = \tau\left(\text{Trend}(X^{(i)}_\text{real}), \text{Trend}(X^{(j)}_\text{real})\right)$
            
            \State $\tau^{(i,j)}_\text{cycle:syn} = \tau\left(\text{Cycle}(X^{(i)}_\text{syn}), \text{Cycle}(X^{(j)}_\text{syn})\right)$ \Comment{And compute the respective correlation scores in cycle.}
            \State $\tau^{(i,j)}_\text{cycle:real} = \tau\left(\text{Cycle}(X^{(i)}_\text{real}), \text{Cycle}(X^{(j)}_\text{real})\right)$
            
            \State \textcolor{white}{.}
            \State $\{\}_\text{trend:syn}^\text{k, i} = \{\}_\text{trend:syn}^\text{k, i}\cup \tau^{(i,j)}_\text{trend:syn}$ \Comment{Record the scores for each patient.}
            \State $\{\}_\text{cycle:syn}^\text{k, i} = \{\}_\text{cycle:syn}^\text{k, i}\cup \tau^{(i,j)}_\text{cycle:syn}$
            \State $\{\}_\text{trend:real}^\text{k, i} = \{\}_\text{trend:real}^\text{k, i}\cup \tau^{(i,j)}_\text{trend:real}$
            \State $\{\}_\text{cycle:real}^\text{k, i} = \{\}_\text{cycle:real}^\text{k, i}\cup \tau^{(i,j)}_\text{cycle:real}$            
        
        \EndFor
    \EndFor
\EndFor

\State \textcolor{white}{.}
\For{$i = 1, N$}\Comment{For averaging across each patient.}
    \State $\{\}_\text{trend:syn} = \{\}_\text{trend:syn}\cup \mathbb{E}_k(\{\}_\text{trend:syn}^\text{k, i})$\Comment{Acquire the average correlation in trends.}  
    \State $\{\}_\text{trend:real} = \{\}_\text{trend:real}\cup \mathbb{E}_k(\{\}_\text{trend:real}^\text{k, i})$
    
    \State $\{\}_\text{cycle:syn} = \{\}_\text{cycle:syn}\cup \mathbb{E}_k(\{\}_\text{cycle:syn}^\text{k, i})$\Comment{Acquire the average correlation in cycles.}  
    \State $\{\}_\text{cycle:real} = \{\}_\text{cycle:real}\cup \mathbb{E}_k(\{\}_\text{cycle:real}^\text{k, i})$
\EndFor
\end{algorithmic}
\end{algorithm}

%%%===
\newpage
\section{Full Correlation Plots for Sepsis}\label{App:Z004}

\begin{figure}[ht]
   \centering
   \includegraphics[width=0.6\linewidth]{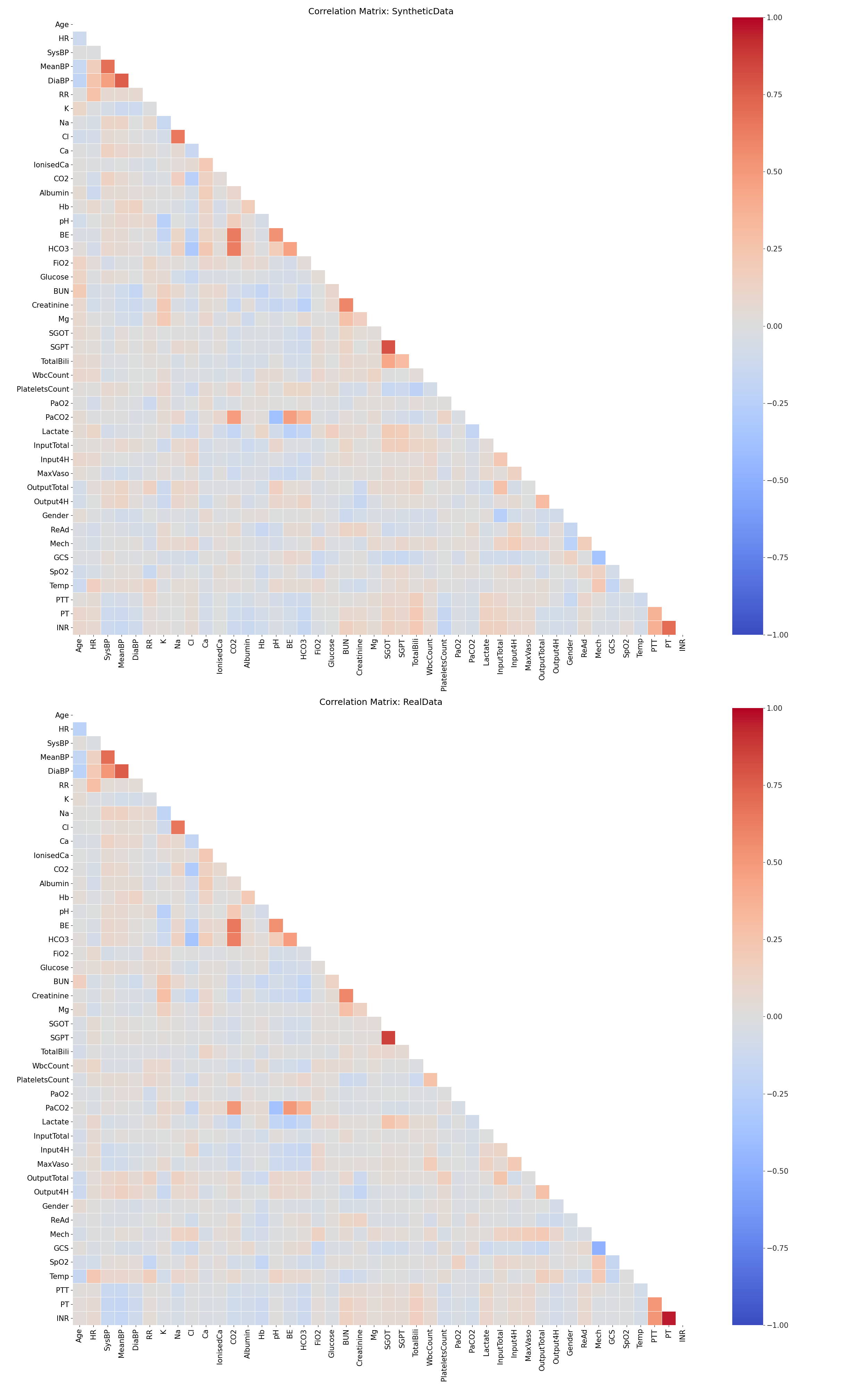}

\caption{The Complete Static Correlation Plots for Sepsis
}
\label{Fig:Sepsis_Complete_Static} 
\end{figure}

\hspace*{-5mm}Our readers can click \hyperlink{Sec:TV-Sepsis}{here} to resume to \textbf{Validation Outcome: Sepsis} in the \textbf{Technical Validation} section.

\newpage
\begin{figure}[ht]
   \centering
   \includegraphics[width=0.6\linewidth]{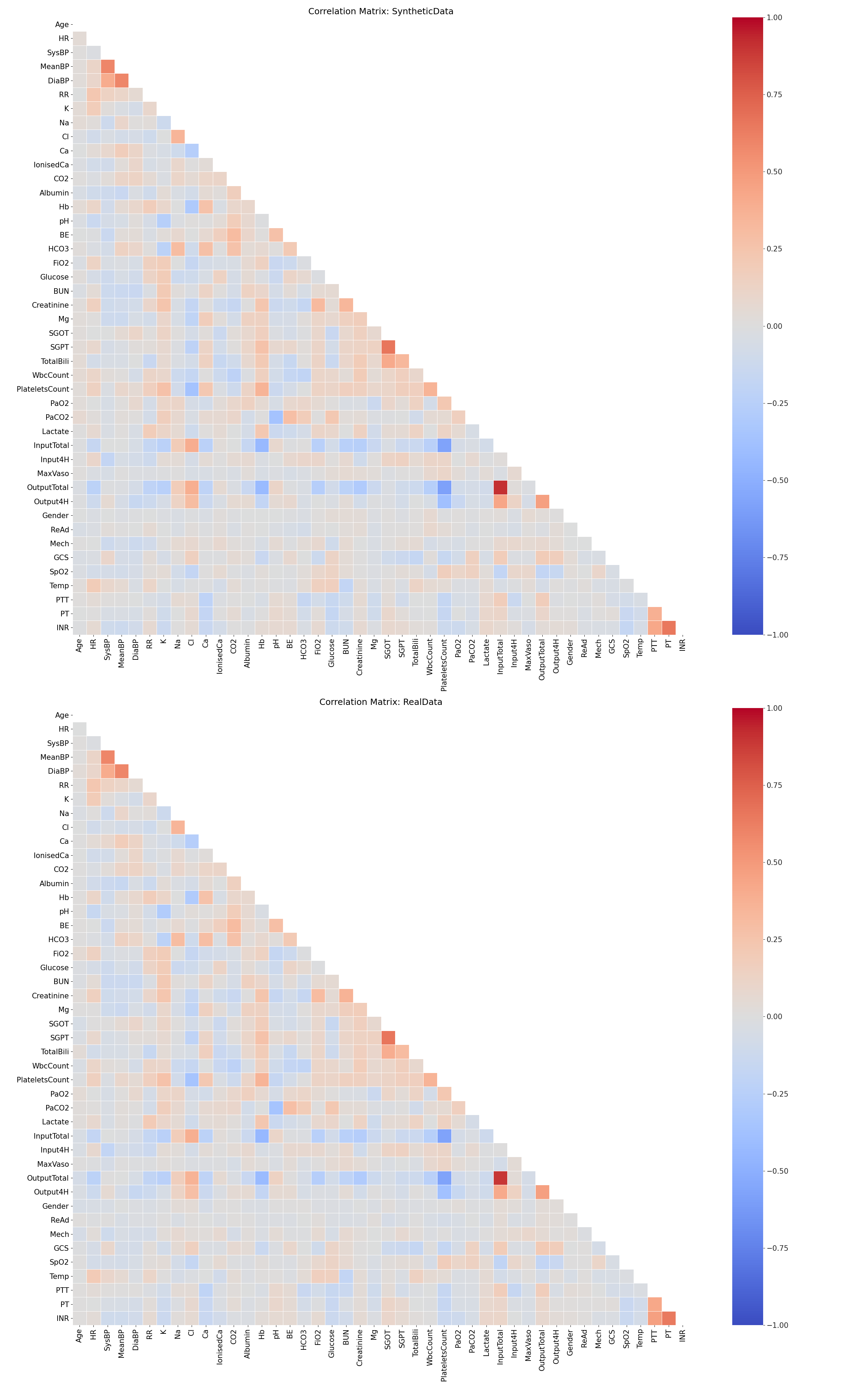}

\caption{The Complete Dynamic Correlations in Trends for Sepsis
}
\label{Fig:Sepsis_Complete_Trends} 
\end{figure}

\newpage
\begin{figure}[ht]
   \centering
   \includegraphics[width=0.6\linewidth]{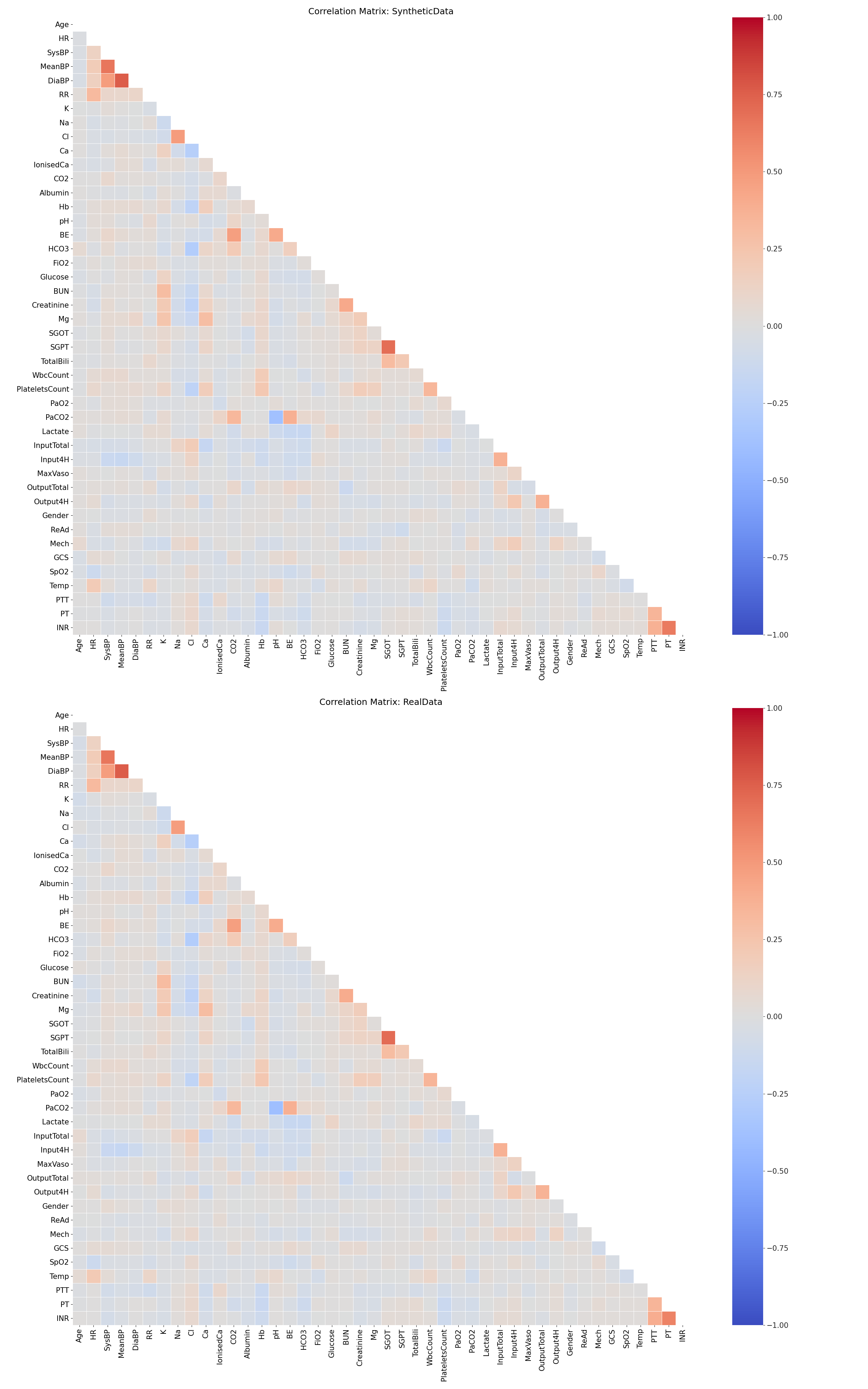}

\caption{The Complete Dynamic Correlations in Cycles for Sepsis
}
\label{Fig:Sepsis_Complete_Cycles} 
\end{figure}

%%%===
\newpage
\section{Assessment of Disclosure Risk}\label{App:Security}

This appendix aims to provide more details for the  assessment of disclosure risk and covers some potential alternative metrics.\\

\hspace*{-5mm}Our readers can click \hyperlink{Sec:TV-SCR}{here} to resume to the \textbf{Security Confirmation Report} in the \textbf{Technical Validation} section.

\subsection{El Emam \etal's Disclosure Risk Metrics}\label{App:Security2}

This appendix provides more intuitions to El Emam \etal\cite{el2020evaluating}'s risk evaluation metrics. Assuming that an adversary was able to recover partial information (\ie the quasi-identifiers) of an individual (\ie the acquaintance), El Emam \etal\cite{el2014concepts} formulated the probability of the successful re-identification (re-id) of the information of an acquaintance as
\begin{align}
    \mathbb{P}(\text{re-id}, \text{acquaintance}) = \mathbb{P}(\text{acquaintance}) \times \mathbb{P}(\text{re-id}\vert \text{acquaintance})\label{Eq:ReIdAc}
\end{align}
using the Bayes rule.

Based on Equation (\ref{Eq:ReIdAc}), their population-to-sample re-identification risk for the real dataset of
\begin{align*}
    \underbrace{\frac{1}{P}}_{\circled{1}}\underbrace{\sum_{s = 1}^{S}}_{\circled{2}}\underbrace{\left(\frac{1}{f_{s}}\right)}_{\circled{3}}
\end{align*}
considers that \circled{2} for all individuals in the real sample dataset, calculate the chance of \circled{1} \textit{any} randomly selected record from the population \circled{3} being matched to a specific record in the real dataset with its equivalent class of quasi-identifiers known (\eg male and 21 years old). Note that component \circled{1} corresponds to $\mathbb{P}(\text{acquaintance})$ -- the selection of a random record from a population of size $P$. In addition, components \circled{2} and \circled{3} are affiliated to $\mathbb{P}(\text{re-id}\vert \text{acquaintance})$ because all patients with the same equivalent class of size $f_s$ (\eg $3$ if there are 3 males that are 21 years old) have an equal chance of being pinpointed with accuracy $\frac{1}{f_s}$ (\ie $\frac{1}{3}$ chance for each 21 years old male to be identified). 

Following the same line of thoughts, their sample-to-population re-identification risk for the real dataset of
\begin{align*}
    \underbrace{\frac{1}{S}}_{\circled{4}}\underbrace{\sum_{s = 1}^{S}}_{\circled{2}}\underbrace{\left(\frac{1}{F_{s}}\right)}_{\circled{5}}
\end{align*}
describes the scenario that \circled{2} for all individuals in the real sample dataset, calculate the chance of \circled{4} \textit{any} randomly selected record from the real dataset \circled{5} that can be traced back to a specific record in the population sharing the same equivalent class of quasi-identifiers. This time, $\mathbb{P}(\text{acquaintance})$ corresponds to component \circled{4} with $\frac{1}{S}$ because the real dataset has a total of $S$ records; and that $\mathbb{P}(\text{re-id}\vert \text{acquaintance})$ is now affiliated with \circled{5} with $\frac{1}{F_s}$ because the total amount of records and their equivalent classes are now sourced from the population.

Our readers should note that Equation (\ref{Eq:FakeRisk}) was actually not the full formulation of El Eman \etal's risk definitions for the synthetic datasets. Instead, their full formulations were expressed as  
\begin{align}
    &\text{the population-to-sample risk}: \hspace{3mm} \frac{1}{P}\sum_{s = 1}^{S}\left(\frac{1}{f_{s}} \times \lambda_s \times R_s \right) \hspace{3mm} \text{and}\\ &\text{the sample-to-population risk}: \hspace{3mm} \frac{1}{S}\sum_{s = 1}^{S}\left(\frac{1}{F_{s}}\times \lambda_s \times R_s\right).
    \label{Eq:FullRisk}
\end{align}
The full expressions included two extra terms. Under the section of \textbf{Adjusting for Incorect Matches} in their paper, the authors noted that $\lambda$ was a constant term $\in(0, 1]$ that scaled down the original re-identifiability for the records in the dataset. This term existed because that health data, and generally any type of data from a data broker, were known to have some error rates\cite{elliot1999scenarios}; and $\lambda$ was introduced to correct the overestimation of meaningful adversarial re-identifications. Furthermore and under the section of \textbf{Learning Something New}, the authors noted that $R$ was a binary term and that $R = 1$ only when any matched records contributed towards learning non-duplicated re-identified information. These two extra terms thus scaled down the original population-to-sample risk and sample-to-population risk. In our own work, we left these two terms out because that the unscaled risks were already very low. 

\subsection{Alternative Disclosure Risk Metrics}\label{App:Security3}

In this study, we employed the Euclidean distance and the synthetic-to-real disclosure risk to investigate the risks associated with the public release of synthetic datasets. However, that there are many alternative options.

Instead of the Euclidean distance, the Mahalanobis distance\cite{de2000mahalanobis} may be well suited to the high dimensionality of electronic health records. Nonetheless, the Euclidean distance was perfectly suitable to show that there were no perfect matches between real and synthetic records.

Furthermore, El Emam \etal's metrics are just two of the many available metrics in the risk disclosure literature. Both the population-to-sample risk and the sample-to-population risk can be seen as extended applications of \textit{k-anonymity}\cite{samarati2001protecting} -- a desirable property for a dataset implying that multiple records share the same equivalence class of quasi-identifiers, thus lowering the chance of identity exposure (\eg it is better if a released dataset contains multiple 21-year-old makes instead of just one). However, it is well-known that attributes of an anonymised dataset could still be inferred even if k-anonymity is satisfied. Attribute disclosure is especially concerning if homogeneity in patient data is high. For instance, if all 21-year-old males suffer from stomach problems and a specific 21-year-old male is the target of an adversary, then the adversary can acquire knowledge automatically without the need to match any records. Attribute disclosure can be mitigated by achieving \textit{l-diversity}\cite{machanavajjhala2007diversity} and/or \textit{t-closeness}\cite{li2007t} in addition to k-anonymity. The former aims to increase the number of sensitive attributes within each equivalence class; whereas the latter uses information theoretic measures to verify if the records satisfy a minimal level of diversification.

Aside from Euclidean distances and disclosure risks, a practitioner could also test the security of their synthetic datasets by building predictive models. For instance, Choi \etal\cite{choi2017generating} used k nearest neighbours to predict a varying number of known attributes in the datasets. This could be helpful to determine whether any known sensitive variables could be easily reconstructed through the quasi-identifiers.

%%%===%%%
\end{document}